\pdfoutput=1
\documentclass{article}

\PassOptionsToPackage{numbers, compress}{natbib}
\usepackage[preprint]{neurips_data_2024}

\usepackage[utf8]{inputenc} % allow utf-8 input
\usepackage[T1]{fontenc}    % use 8-bit T1 fonts
\usepackage{hyperref}
\usepackage{booktabs}       % professional-quality tables
\usepackage{amsfonts}       % blackboard math symbols
\usepackage{nicefrac}       % compact symbols for 1/2, etc.
\usepackage{microtype}      % microtypography
\usepackage{times}
\usepackage{soul}
\usepackage{graphicx}
\usepackage{amsmath}
\usepackage{amsthm}
\usepackage{algorithm}
\usepackage{algorithmic}
\usepackage[switch]{lineno}
\usepackage[table]{xcolor}
\usepackage{caption}
\usepackage{natbib}
\usepackage{enumitem}
\usepackage{overpic}
\usepackage{multirow}
\usepackage{subcaption}
\usepackage{array}
\usepackage{tabularx}
\usepackage{bbm}
\usepackage{makecell}
\usepackage{adjustbox}
\usepackage{xparse}
\usepackage{pifont}% http://ctan.org/pkg/pifont
\usepackage{url}

\usepackage[toc,page]{appendix}
\usepackage{minitoc}
\usepackage{tocloft}

\newcommand{\toccolor}{\hypersetup{linkcolor=black}}
\newcommand{\defaultcolor}{\hypersetup{linkcolor=red}}

\definecolor{lightgrey}{rgb}{0.92,0.92,0.92}
\definecolor{darkgreen}{rgb}{0.502, 0.643, 0.502}
\definecolor{tablered}{rgb}{0.95, 0.9, 1}
\definecolor{ao(english)}{rgb}{0.0, 0.5, 0.0}
\definecolor{CRcolor}{rgb}{0.85, 0.93, 0.75}
\definecolor{VEcolor}{rgb}{1, 0.725, 0.59}
\definecolor{VCcolor}{rgb}{1, 0.812, 0.51}
\definecolor{GCcolor}{rgb}{0.99, 1, 0.67}

\newcommand{\red}[1]{{\color{red}#1}}

\newcommand{\itemnote}[1]{\textit{\textbf{#1}}}
\newcommand{\good}[1]{\textcolor{ao(english)}{{\scriptsize #1}}}
\newcommand{\bad}[1]{\textcolor{red}{\scriptsize #1}}
\newcommand{\myparagraph}[1]{\vspace{1mm}\noindent{\textbf{#1}}}

\title{ChartBench: A Benchmark for Complex Visual Reasoning in Charts}

\author{
Zhengzhuo Xu$^{1,2}$\thanks{Equal Contribution}~~, \quad \quad 
Sinan Du$^{2*}$~~,\quad \quad 
Yiyan Qi$^{1}$ \\
\textbf{Chengjin Xu$^{1}$},\quad \quad 
\textbf{Chun Yuan$^{2}$},\quad \quad 
\textbf{Jian Guo$^{1,3}$} \\
\textsuperscript{1}International Digital Economy Academy (IDEA) \quad \\
\textsuperscript{2}Shenzhen International Graduate School, Tsinghua University \quad \\
\textsuperscript{3}The Hong Kong University of Science and Technology, Guangzhou
}

\begin{document}

\maketitle

\begin{abstract}
Multimodal Large Language Models (MLLMs) have shown impressive capabilities in image understanding and generation. However, current benchmarks fail to accurately evaluate the chart comprehension of MLLMs due to limited chart types and inappropriate metrics. To address this, we propose ChartBench, a comprehensive benchmark designed to assess chart comprehension and data reliability through complex visual reasoning. ChartBench includes 42 categories, 66.6k charts, and 600k question-answer pairs. Notably, many charts lack data point annotations, which requires MLLMs to derive values similar to human understanding by leveraging inherent chart elements such as color, legends, and coordinate systems. We also design an enhanced evaluation metric, \textit{Acc+}, to evaluate MLLMs without extensive manual or costly LLM-based evaluations. Furthermore, we propose two baselines based on the chain of thought and supervised fine-tuning to improve model performance on unannotated charts. Extensive experimental evaluations of 18 open-sourced and 3 proprietary MLLMs reveal their limitations in chart comprehension and offer valuable insights for further research. Code and dataset are publicly available at {\url{https://chartbench.github.io}}.
\end{abstract}

\addtocontents{toc}{\protect\setcounter{tocdepth}{-1}}

\section{Introduction}
\label{sec:intro}
\vspace{-5pt}

Given the groundbreaking advancements in Large Language Models (LLMs)~\cite{clip,GPT-3,PaLM,llama}, Multimodal Large Language Models (MLLMs)~\cite{BLIP2,llava,Minigptv1} have become the leading approach in multimodal learning, which exhibit excellent visual semantics understanding performance~\cite{GPT4,cogvlm}. However, existing MLLMs face challenges in effectively reading, comprehending, and summarizing articles that contain embedded charts~\cite{ChartQA,Chartllama,sciGraphQA}. Unlike natural images, which are typically interpreted based on discernible objects, relative positions, or interactions, charts convey nuanced semantic meanings through \textit{visual logic}, such as trend lines or color-coded legends. They present detailed and intricate data narratives in visual formats, making it essential to evaluate MLLMs' chart comprehension ability and data reliability in understanding these visual representations.

Previous works~\cite{ChartQA,PlotQA,OpenCQA,ChartX,OneChart} have attempted to address this issue but have encountered some limitations. 1) They primarily focus on 3 regular chart types (i.e., line, bar, and pie charts), neglecting more intricate formats such as scatter or combination charts which are equally prevalent in real-world scenarios. Robust MLLMs should be able to adeptly handle a diverse range of chart types. 2) They heavily rely on \textit{datapoint annotation} on charts or \textit{meta table data} as textual prompts~\cite{ChartQA,Chartllama,OneChart} to generate content, allowing models to easily obtain candidate answers while ignoring the charts' \textit{visual logic}. This will cause MLLMs to struggle with unannotated charts in real-world applications. 3) Current evaluation metrics cannot avoid lucky guesses and thus result in overestimated baseline performance, which requires refinement to enhance assessment objectivity and precision.

\begin{table}[t]
\setlength\tabcolsep{8pt}
\caption{Comparative analysis with the existing benchmarks for chart-related evaluations. 
%ChartBench is proposed to comprehensively evaluate off-the-shelf MLLMs with more chart types, particularly in scenarios where chart data points are unannotated. 
\textit{Aggregated} charts are derived from the consolidation of existing datasets. \# refers to corresponding quantity. \textit{Visual} refers to the inclusion of assessments for unannotated charts, where the models are expected to interpret the chart logical structure without relying on OCR to answer queries.}
\resizebox{\linewidth}{!}{
\begin{tabular}{@{}lccccccccc@{}}
\toprule[1.5pt]
\multicolumn{1}{c}{\multirow{2}{*}{Benchmark}} & \multirow{2}{*}{Image Source} & \multicolumn{2}{c}{Type} & \multicolumn{2}{c}{Train Set} & \multicolumn{2}{c}{Test Set} & \multirow{2}{*}{\begin{tabular}[c]{@{}c@{}}Multi-task\\ Evaluation\end{tabular}} & \multirow{2}{*}{\textit{Visual}} \\ \cmidrule(lr){3-8}
\multicolumn{1}{c}{} &  & \#Chart & \#Task & \#Chart & \#QA & \#Chart & \#QA &  &  \\ \midrule
\multicolumn{1}{l|}{ChartQA~\cite{ChartQA}} & \multicolumn{1}{c|}{\textit{Original}} & 3 & \multicolumn{1}{c|}{1} & 21.9K & \multicolumn{1}{c|}{32.7K} & 1.5K & \multicolumn{1}{c|}{1.5K} & \textcolor{black}{\ding{56}} & \textcolor{black}{\ding{56}} \\
\multicolumn{1}{l|}{PlotQA~\cite{PlotQA}} & \multicolumn{1}{c|}{\textit{Original}} & 3 & \multicolumn{1}{c|}{1} & 224K & \multicolumn{1}{c|}{28M} & 33.7K & \multicolumn{1}{c|}{33.7K} & \textcolor{black}{\ding{56}} & \textcolor{black}{\ding{56}} \\
\multicolumn{1}{l|}{Chart-to-text~\cite{Chart2Text}} & \multicolumn{1}{c|}{\textit{Original}} & 6 & \multicolumn{1}{c|}{1} & 44K & \multicolumn{1}{c|}{44K} & 6.6K & \multicolumn{1}{c|}{6.6K} & \textcolor{black}{\ding{56}} & \textcolor{black}{\ding{56}} \\
\multicolumn{1}{l|}{OpenCQA~\cite{OpenCQA}} & \multicolumn{1}{c|}{\textit{Original}} & 5 & \multicolumn{1}{c|}{1} & - & \multicolumn{1}{c|}{-} & 1.2K & \multicolumn{1}{c|}{1.2K} & \textcolor{black}{\ding{56}} & \textcolor{black}{\ding{56}} \\
\multicolumn{1}{l|}{UniChart~\cite{unichart}} & \multicolumn{1}{c|}{\textit{Aggregated}} & 3 & \multicolumn{1}{c|}{3} & 627K & \multicolumn{1}{c|}{7M} & - & \multicolumn{1}{c|}{-} & \textcolor{black}{\ding{52}} & \textcolor{black}{\ding{56}} \\
\multicolumn{1}{l|}{ChartLlama~\cite{Chartllama}} & \multicolumn{1}{c|}{\textit{Original}} & 10 & \multicolumn{1}{c|}{7} & 11K & \multicolumn{1}{c|}{160K} & 2.1K & \multicolumn{1}{c|}{3.5K} & \textcolor{black}{\ding{52}} & \textcolor{black}{\ding{56}} \\
\multicolumn{1}{l|}{MMC~\cite{MMC}} & \multicolumn{1}{c|}{\textit{Aggregated}} & 6 & \multicolumn{1}{c|}{9} & 600K & \multicolumn{1}{c|}{600K} & 2K & \multicolumn{1}{c|}{2K} & \textcolor{black}{\ding{52}} & \textcolor{black}{\ding{56}} \\
\multicolumn{1}{l|}{ChartX~\cite{ChartX}} & \multicolumn{1}{c|}{\textit{Original}} & 18 & \multicolumn{1}{c|}{7} & - & \multicolumn{1}{c|}{-} & 6K & \multicolumn{1}{c|}{6K} & \textcolor{black}{\ding{52}} & \textcolor{black}{\ding{56}} \\ \midrule
\multicolumn{1}{l|}{ChartBench(ours)} & \multicolumn{1}{c|}{\textit{Original}} & 9\ /\ 42 & \multicolumn{1}{c|}{5} & 66.6K & \multicolumn{1}{c|}{599.6K} & 2.1K & \multicolumn{1}{c|}{18.9K} & \textcolor{black}{\ding{52}} & \textcolor{black}{\ding{52}} \\ 
\bottomrule[1.5pt]
\end{tabular}
}
\vspace{-20pt}
\label{tab_bench_info}
\end{table}

\begin{table}[t]
\centering
\caption{ChartBench comprises 3 regular charts and expands to include 6 additional types. ChartBench emphasizes charts that lack data point annotations, requiring the MLLMs to infer the correct answers by considering elements such as color, legends, and coordinate systems like humans.}
\resizebox{\linewidth}{!}{
\setlength{\tabcolsep}{10pt}
\begin{tabular}{@{}c|cc|ccccccccc@{}}
\toprule[1.5pt]
\multirow{2}{*}{Data Split} & \multicolumn{2}{c|}{Annotation Distribution} & \multicolumn{9}{c}{Chart Type Distribution} \\ \cmidrule(l){2-12} 
 & \textit{w/i} & \textit{w/o} & Line & Bar & \multicolumn{1}{c|}{Pie} & Area & Box & Radar & Scatter & Node & Combin. \\ \midrule
Train Set & 15.04\% & 84.96\% & 11.75\% & 36.89\% & \multicolumn{1}{c|}{12.72\%} & 8.42\% & 6.11\% & 4.59\% & 3.07\% & 5.97\% & 10.47\% \\
Test Set & 23.80\% & 76.20\% & 11.90\% & 31.00\% & \multicolumn{1}{c|}{11.90\%} & 7.10\% & 7.10\% & 9.50\% & 7.10\% & 4.80\% & 11.90\% \\ 
\bottomrule[1.5pt]
\end{tabular}
}
\vspace{-15pt}
\label{tab_chartbench_distribution}
\end{table}

To address these limitations, we introduce ChartBench, which comprehensively evaluates the performance of MLLMs on a wider variety of chart types, including both annotated and unannotated charts. As summarized in Tab.~\ref{tab_bench_info}, ChartBench includes over 68k charts and more than 600k high-quality instruction data, covering 9 major categories and 42 subcategories of charts. Additionally, ChartBench has 5 chart question-answering tasks to assess the models' cognitive and perceptual abilities.
To assess MLLMs' abilities on unannotated charts, ChartBench includes unannotated charts across all 42 categories. Experimental results show a significant performance gap between charts with and without datapoint annotations (Tab.~\ref{tab_data_anno}). To enhance model capabilities on unannotated charts, over 80\% of the training set in ChartBench are unannotated charts (Tab.~\ref{tab_chartbench_distribution}) and our supervised fine-tuning baselines demonstrate significant improvement. 
We further introduce the \textit{Acc+} metric inspired by MME~\cite{MME} to ensure rigorous evaluations. This metric requires MLLMs to accurately judge both positive and negative assertions. The negative one differs from the positive only in the ground truth value, derived from other data within the same chart, ensuring realism. This approach minimizes lucky guesses, as MLLMs may produce identical responses for both query types if they fail to understand the chart. To prevent excessive negative samples in value extraction tasks, we also improve the metric in ChartQA~\cite{ChartQA} from the query format and answer extraction.

The evaluation of 18 mainstream open-source and 3 closed-source models shows that current MLLMs cannot effectively understand complex charts, especially those without data annotations, raising concerns about the reliability of their data interpretation. Detailed examinations on ChartBench reveal the reasons behind the suboptimal performance of MLLMs on charts, highlighting ChartBench's meticulous curation to explore the nuances of chart reasoning. We introduce two simple yet effective baselines based on the chain of thought (CoT, Fig.~\ref{fig_CoT}) and supervised fine-tuning (SFT) to improve MLLMs' performance on ChartBench, aiming to inspire more innovative proposals in the future.

Our contributions can be summarized as follows:
\setlength{\leftmargini}{25pt}
\begin{enumerate}[itemsep=-2pt, topsep=-5pt, label=\alph*)]
    \item We introduce ChartBench, a large-scale dataset with over 42 types of charts, 66k charts, and 600k instructions. It primarily includes charts without data point annotations, assessing MLLMs' ability to reason through visual elements instead of OCR.
    \item We refine the \textit{Acc+} metric and value matching criteria to effectively reduce random guesses and provide more robust evaluation results of 18 open-sourced and 3 closed-sourced MLLMs.
    \item We propose two efficient baselines based on the chain of thought and supervised fine-tuning, inspiring more methods to enhance MLLMs' understanding of unannotated charts.
    \item Extensive experiments reveal existing MLLMs' inadequacies in chart comprehension, highlighting potential directions for future optimization.
\end{enumerate}

\section{Related Works}
\label{sec_related_work}
\vspace{-5pt}

\subsection{Multimodal LLMs}
Current LLMs~\citep{Transformer, GPT-1,GPT-3,OPT,PaLM,llama,llama-2, internlmfamily} successfully bridge the multimodal areas via instruction tuning ~\citep{InstructGPT,otter,self-instruct}.  The connectors are proposed to align visual and text modality to train MLLMs~\cite{VisualGPT, Flamingo}, e.g., Q-Former~\citep{BLIP2} or MLP~\cite{qwen-vl}. Mini-GPT4~\citep{Minigptv1,Minigptv2}, mPLUG-Owl~\citep{Mplug}, and InstructBLIP~\citep{InstructBLIP} extend language-only instruction tuning to multimodal tasks using Q-Former. LLaVA~\citep{llava,Improvedllava} maps visual features into the LLaMA~\citep{llama} embedding space by a linear layer, while concurrently fine-tuning with LLaMA. The closed-source Baidu ERNIE~\cite{wenxin} and GPT-4~\citep{GPT4} further show satisfactory image understanding capabilities. Despite the impressive achievements of existing MLLMs~\citep{cogview,glm,internlm-xcomposer,qwen-vl,shikra,sphinx} in common multimodal tasks like VQA~\citep{VQA} and image captioning~\citep{ImageCaption}, their focus tends to be on general image understanding, neglecting the specialized task of comprehending chart data in domain-specific contexts~\citep{ChartQA,sciGraphQA,Chartllama, MMC, Structchart}. Existing research can be divided into two categories. 1) two-stage methods mainly transform multimodal queries into text QAs by extracting table information as prompt~\cite{Pix2Str, MatCha, DePlot, ChartX}. 2) end-to-end approaches adopt chart-question pair data to align and supervised fine-tune the MLLMs~\cite{Chartllama, ChartPaLI, ChartAst, MMC, Ureader, ChartThinker, DOMINO, LAMENDA, ChartReformer, OneChart, TinyChart}. Although these efforts have improved the chart understanding ability of MLLMs, there are still limited benchmarks to properly evaluate their performance on the charts, especially unannotated ones.

\subsection{Multimodal Benchmarks}
MLLMs have been fully evaluated on numerous traditional benchmarks~\citep{VQAv2,GQA,LVLM-eHub,OwlEval,MME,MM-Vet,Seed-Bench,MMBench}, while largely ignoring the requirement for complex visual chart understanding and reasoning. HallusionBench~\citep{Hallusionbench} exposes the susceptibility of formidable models like GPT-4V~\citep{GPT4} and LLaVA-1.5~\citep{Improvedllava} to severe hallucinations when confronted with complex charts. VisText~\citep{Vistext} introduces a benchmark to incorporate multi-level and fine-grained chart labeling, covering aspects such as chart construction, summary statistics, relations, and complex trends. SciCap~\citep{SciCap}, Chart2Text~\citep{Chart2Text}, AutoChart~\citep{AutoChart}, and ChartSumm~\citep{ChartSumm} address chart-to-text summarization tasks. ChartQA~\cite{ChartQA} and PlotQA~\cite{PlotQA} are currently mainstream benchmark datasets for evaluating the chart comprehension abilities of MLLMs, which focus on three commonly encountered chart types. Chartllama~\cite{Chartllama} and ChartX~\cite{ChartX} expand the range of available chart types, while ChartY~\cite{OneChart} significantly expands the number of regular chart types with LLMs. However, these benchmarks have limited chart types, and their charts are always accompanied by detailed datapoint annotations, which allow MLLMs to obtain candidate answers via simple OCR. Comparatively, the advantages of ChartBench stem from its larger scale, more diverse chart types, richer plot styles, and high proportion of unannotated charts.

\section{ChartBench}
\vspace{-5pt}

\subsection{Data Processing Pipeline}
\begin{figure*}[t!]
    % \hsize=\textwidth
    \centering
      \begin{overpic}[width=\linewidth, grid=False]{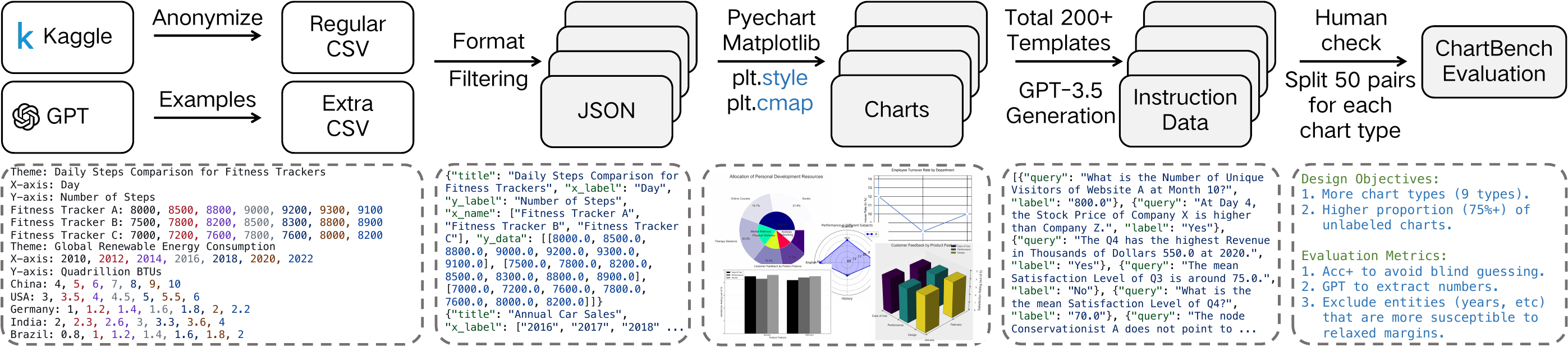}
      \end{overpic}
      \vspace{-10pt}
      \caption{Illustration of the overall data collection and annotation pipeline. We adopt desensitized and GPT-generated data. We employ a variety of charting methods, styles, and color combinations to ensure chart diversity. We provide over 200 question templates and GPT-generated questions to ensure question diversity. Each sample in the test set undergoes manual checks to prevent errors.}
      \vspace{-10pt}
    \label{fig_pipeline}
\end{figure*}

Fig.~\ref{fig_pipeline} illustrates the specific data processing flow of Chartbench. The core idea is \textit{to generate unannotated charts of various types and their corresponding instruction data.} 1) \itemnote{Data collection}. To design charts reflecting real-world scenarios, we gather themes and data suitable for scientific research from Kaggle, anonymizing all real names and identifiable entities to ensure privacy. To ensure the diversity of chart types, we also use LLMs~\cite{ChatGPT, deepseek, qwen} to generate realistic virtual themes and data for additional chart types. 2) \itemnote{Data filtering}. We establish standard JSON formats for 42 chart types and filter out all table data that does not conform to these standards to ensure proper chart generation. 3) \itemnote{Chart generation}. With effective data filtering, we plot various charts using various chart plotting libraries (such as \textit{Matplotlib}, etc.). We randomly applied different plotting styles and color schemes to ensure chart diversity and provide 9 major categories and 42 subcategories of charts (Tab.~\ref{tab_chartbench_distribution}). Refer to Appendix~\ref{sec_apdx_chartbench_stat} \& \ref{apdx_sec_thumbnails} for detailed descriptions and thumbnail visualizations. Specifically, we designate a proportion of charts without data point markers, which is a significant feature of ChartBench. 4) \itemnote{Instructions generation}. We set 5 different tasks for each type of chart and adopt appropriate metrics for evaluation. Detailed instruction tasks will be explained in Sec.~\ref{sec_method_qa_gene}. 5) \itemnote{Dataset splitting}. We randomly select 50 samples for each chart type to form the benchmark, with the specific distribution shown in Tab.~\ref{tab_chartbench_distribution}. Since the plots are generated by code, the plotting style inevitably appears somewhat rigid. Hence, while maintaining consistent basic settings, \textit{we choose part data from the test split to be plotted using online plotting websites to ensure a certain domain gap}. Subsequently, we conduct expert reviews to eliminate defective samples (e.g., label occlusions) for test split. This process yields both the metadata and rendered chart images.

\subsection{Automatic Instructions Generation}
\label{sec_method_qa_gene}
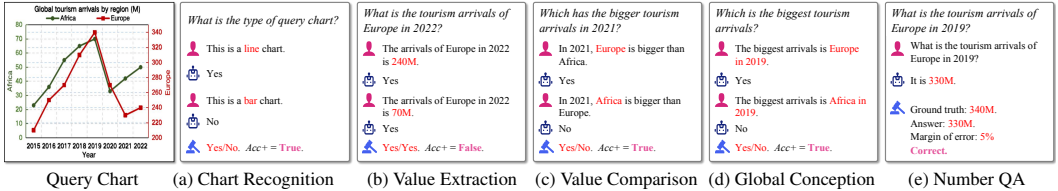
\begin{figure*}[t]
  \centering
  \begin{overpic}[width=\linewidth, grid=False]{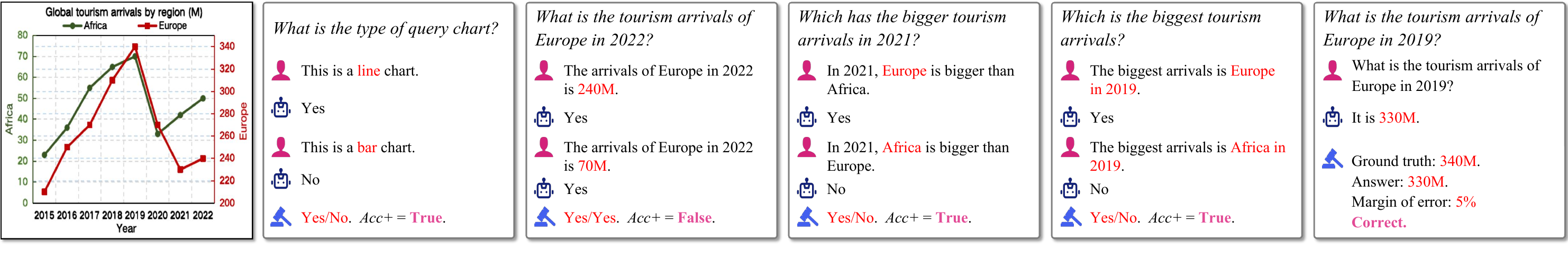}
  \put(4,-1){\scriptsize Query Chart}
  \put(16,-1){\scriptsize (a) Chart Recognition}
  \put(34.5,-1){\scriptsize (b) Value Extraction}
  \put(50.5,-1){\scriptsize (c) Value Comparison}
  \put(67,-1){\scriptsize (d) Global Conception}
  \put(86,-1){\scriptsize (e) Number QA}
  \end{overpic}
  % \vspace{1pt}
  \caption{Illustration of five proposed tasks. Tasks (a-d) are with \textit{Acc+} and (e) with GPT-acc metric.}
\vspace{-15pt}  
\label{fig_QA_types}
\end{figure*}

ChartBench consists of 5 tasks, encompassing \textit{perception} and \textit{conception}~\cite{MME} tasks. \textit{Perception} tasks primarily entail perceiving and processing raw data to extract valuable features and information. Conversely, \textit{conception} tasks involve processing and comprehending abstract concepts and higher-level information. \textit{Perception} tasks primarily encompass two types of QAs: 1) \itemnote{Chart type Recognition} (CR, Fig.~\ref{fig_QA_types}\textcolor{red}{a}) task aims to evaluate the MLLMs' capability to identify chart types accurately. 2) \itemnote{Value Extraction} (VE, Fig.~\ref{fig_QA_types}\textcolor{red}{b}) task aims to assess whether MLLMs can correctly extract the relevant values when confronted with complex visual logic. Without annotated data, MLLMs are required to rely on legends, axes, and corresponding graphical elements to provide answers. \textit{Conception} tasks include two types of QAs: 3) \itemnote{Value Comparison} (VC, Fig.~\ref{fig_QA_types}\textcolor{red}{c}) assesses MLLMs' visual reasoning by requiring them to rely solely on graphical elements, not metadata, to determine comparison answers. 4) \itemnote{Global Conception} (GC, Fig.~\ref{fig_QA_types}\textcolor{red}{d}) task assesses the ability to perceive global indicators, such as maximum values, from a holistic standpoint. Nevertheless, considering the excessive number of negative samples in the VE task, we additionally use a tolerance evaluation method like ChartQA~\cite{ChartQA}. Values within a certain error range are considered correct, which we refer to as the 5) \itemnote{NumberQA} task (NQA, Fig.~\ref{fig_QA_types}\textcolor{red}{e}). In summary, the MLLMs are not required to identify all the chart metadata or element layouts. On the contrary, simply observing graphic elements and identifying key components is sufficient to arrive at accurate conclusions.

\subsection{Dataset Analysis}

\begin{figure}
    \centering
    \begin{subfigure}[b]{0.31\textwidth}
        \includegraphics[width=\textwidth]{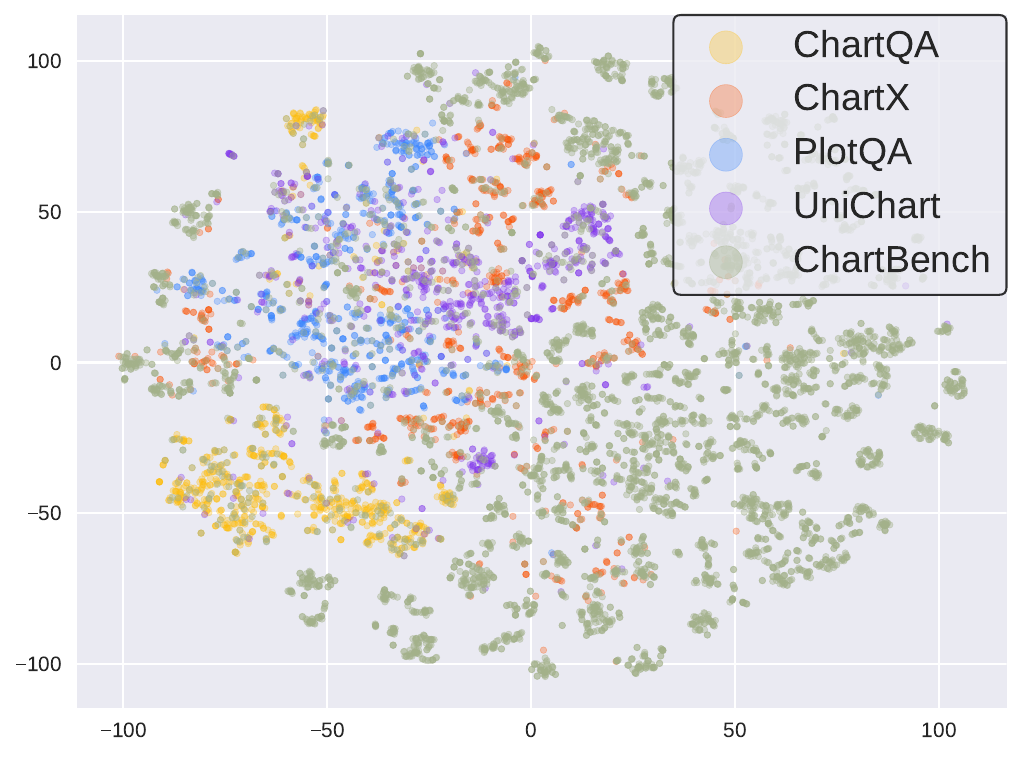}
        \caption{Chart}
        \label{fig_tsne_chart}
    \end{subfigure}
    \hfill
    \begin{subfigure}[b]{0.31\textwidth}
        \includegraphics[width=\textwidth]{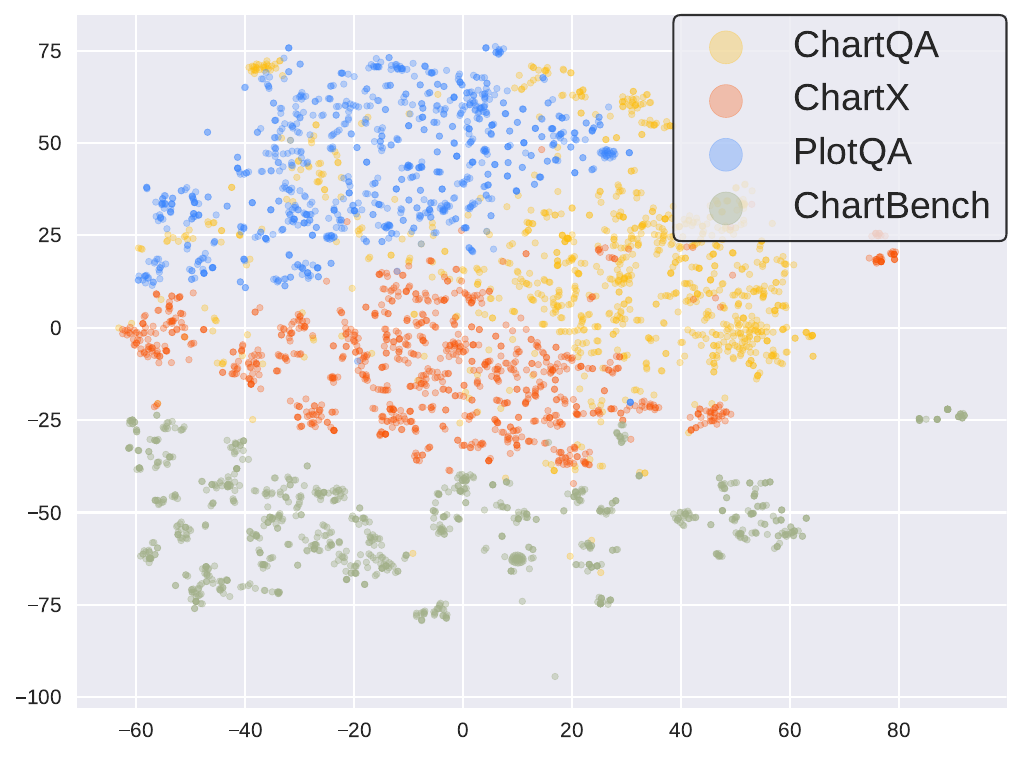}
        \caption{CSV}
        \label{fig_tsne_csv}
    \end{subfigure}
    \hfill
    \begin{subfigure}[b]{0.31\textwidth}
        \includegraphics[width=\textwidth]{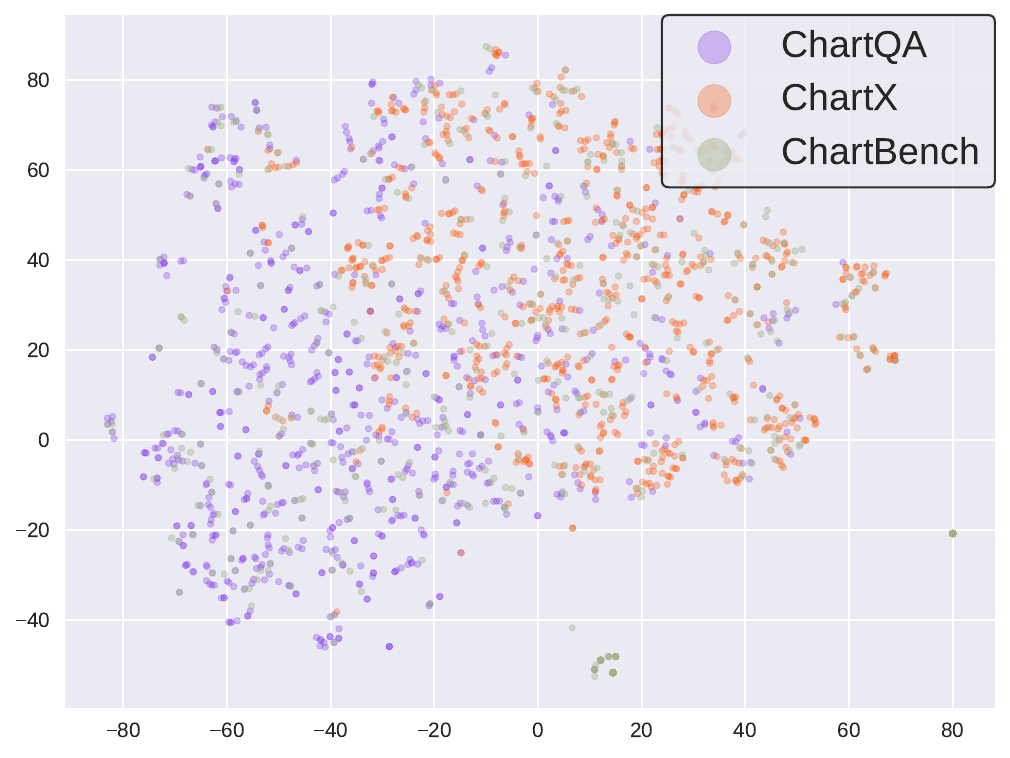}
        \caption{Query}
        \label{fig_tsne_query}
    \end{subfigure}
    \vspace{-5pt}
    \caption{t-SNE~\cite{tSNE} visualisation of CLIP encoding features~\cite{clip}. ChartBench (a) covers extensive distribution of charts, particularly with the unannotated chart; (b) stands apart from other datasets in terms of both topic and table data; (c) maintains consistent query manners with other datasets.}
    \vspace{-15pt}
    \label{fig_tsne}
\end{figure}

Fig.~\ref{fig_tsne} illustrates the distribution of chart, meta CSV, and query data, respectively. We randomly sample 10,000 data points respectively and extract corresponding features via CLIP (ViT-B/16) encoder~\cite{clip}. We adopt t-SNE~\cite{tSNE} for feature dimensionality reduction for visualizations. 1) \itemnote{Chart distribution}. As shown in Fig.~\ref{fig_tsne_chart}, ChartBench covers the main range of charts from previous benchmarks. ChartBench and ChartX~\cite{ChartX} are quite similar in distribution trends. However, ChartBench adopts more plot style ( e.g., \textit{classic, solarize, mpl, bmh, seaborn, ggplot}, etc) to achieve style diversification. ChartQA significantly distinguishes it from other datasets for real-world charts. Our ChartBench supplements this aspect by including charts created using online plotting websites. 2) \itemnote{CSV Distribution}. Our raw data is stored in CSV format. As shown in Fig.~\ref{fig_tsne_csv}, the CSVs of each dataset exhibit different distributions, indicating significant variations in table information. Considering the text truncation length of the CLIP text encoder, this distribution also reflects the differences between the original data topics, as the leading data often includes titles, and labels for the \textit{x} and \textit{y} axes. 3) \itemnote{Query Distribution}. As shown in Fig.~\ref{fig_tsne_query}, the query style of ChartBench is generally consistent with that of ChartQA~\cite{ChartQA} and ChartX~\cite{ChartX}. Note that we only display the QA task features of each dataset. Maintaining a similar querying style helps in comparing model performance across different datasets.

\subsection{Evaluation Metrics}
\label{sec_method_metrics}

\myparagraph{Improved \textit{Acc+}.} 
As shown in Fig.~\ref{fig_QA_types}, for a base query $Q_i$ on chart $c$, we expand $Q_i$ into correct ($Q^r_i$) and incorrect ($Q^w_i$) assertions using a given query prompt. ChartBench requires the MLLM $\mathcal{M}$ to determine the correctness of the queries, providing boolean outputs $A^r_i:=\mathcal{M}(Q^r_i;c)$ and $A^w_i:=\mathcal{M}(Q^w_i;c)$. Because of the concise outputs, we can use regular expression matching instead of additional LLM judgement~\cite{GPTscore}. We note that: 1) $Q^r_i$ and $Q^w_i$ differ only in the ground truth value, resulting in similar token sequences. 2) $A^r_i$ and $A^w_i$ are derived from independent inferences. 3) The incorrect value in $Q^w_i$ is randomly selected from metadata to maintain rationality. We define the improved \textit{Acc+} metric as follows: Given $N$ base queries in ChartBench, $\textit{Acc+} = \frac{1}{N}\sum_{i=1}^N \mathbbm{1}\left[\mathcal{M}(Q^r_i;c) \land \lnot \mathcal{M}(Q^w_i;c)\right]$, where $\land$, $\lnot$ and $\mathbbm{1}[x]$ are \textit{and}, \textit{not} and indicator function, respectively. The MLLM is considered to understand the query chart only if it accurately answers both $Q^r_i$ and $Q^w_i$ simultaneously. 

\myparagraph{Confusion Rate (CoR).} 
During the evaluation, we find that many MLLMs produce the same output for both assertions, likely because they fail to utilize the chart information. To assess this failure, we introduce the \textit{CoR} metric. Formally, $\textit{CoR} = \frac{1}{N}\sum_{i=1}^N \mathbbm{1}\left[\mathcal{M}(Q^r_i;c) \oplus \lnot \mathcal{M}(Q^w_i;c)\right]$, where $\oplus$ denotes the XOR operation. If an MLLM fails to use the visual information from charts, it tends to generate identical answers, resulting in \textit{CoR} approaching 100\%.

\myparagraph{GPT-acc.} 
While \textit{Acc+} is an efficient way to evaluate model responses, it falls short for specific numerical questions, as correctly answering a negative sample doesn't fully demonstrate the model's generalization ability and differs from methods used in datasets like ChartQA. To address this, we propose an improved error margin evaluation (5\%) from ChartQA~\cite{ChartQA}. Our improvements include: 1) using LLMs~\cite{ChatGPT, qwen, deepseek} to filter responses and extract numerical answers, avoiding pattern-matching errors due to extraneous text, and 2) restricting NQA task questions to exclude elements like years and months, which could make the error margin too lenient and the evaluation meaningless.

\section{Baselines}
\label{sec_chartcot}
\vspace{-5pt}

ChartBench primarily evaluates MLLMs' ability to understand unannotated charts. We propose two simple yet effective baselines that significantly improve MLLMs' performance.

\begin{figure}[t]
    \centering
    \begin{overpic}[width=\linewidth, grid=False]{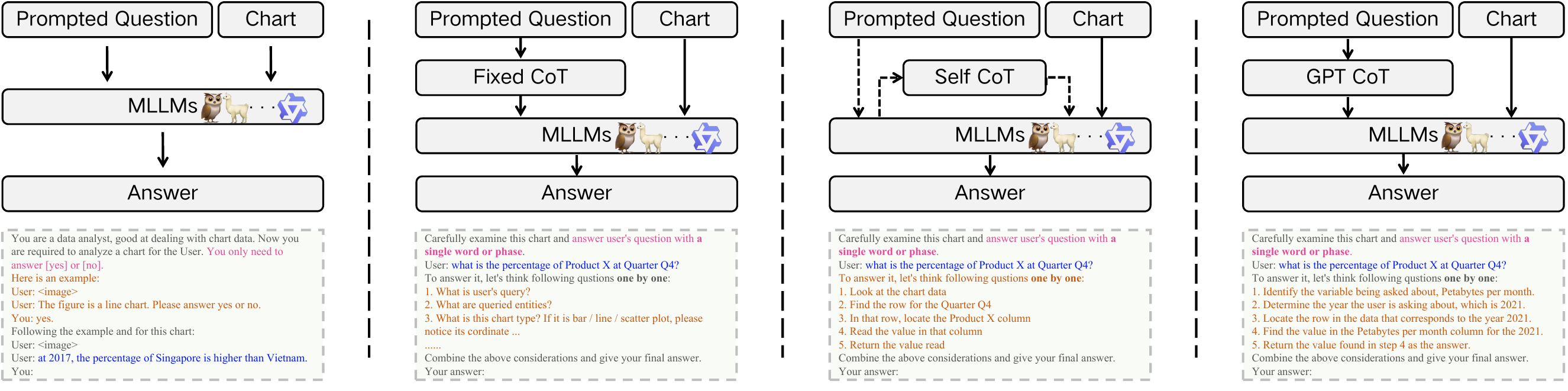}
    \put(7,-2){\scriptsize (a) Base}
    \put(32,-2){\scriptsize (b) Fixed CoT}
    \put(58,-2){\scriptsize (c) Self CoT}
    \put(85,-2){\scriptsize (d) GPT CoT}
    \end{overpic}
    \vspace{-5pt}
    \caption{Illustration of different Chain of Thought. (a) No CoT. (b) All charts utilize the same CoT template that we provide. (c) The CoT for each chart is generated by its own LLM given the prompted question. (d) The CoT for each chart is generated by GPT given the prompted question.}
    \vspace{-15pt}
    \label{fig_CoT}
\end{figure}
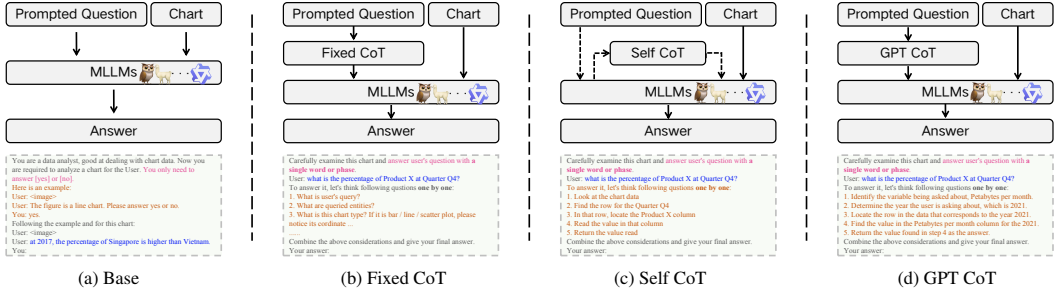

\itemnote{ChartCoT.} 
As shown in Fig.~\ref{fig_CoT}, we propose effective baselines based on Chain of Thought~\cite{CoT} to enhance the visual reasoning capability without model tuning. As shown in Fig.~\ref{fig_CoT}\textcolor{red}{b}, we design a series of questions that decompose user inquiries and employ prompts to mimic human visual reasoning for chart recognition. Additionally, we enable MLLMs to generate their own CoT (Fig.~\ref{fig_CoT}\textcolor{red}{c}) or seek assistance from stronger LLMs to generate CoTs (Fig.~\ref{fig_CoT}\textcolor{red}{d}). This approach significantly aids MLLMs in understanding charts, particularly in cases where visual logic is more complicated.

\itemnote{Supervised Fine-tuning.}
 We conduct a two-stage supervised fine-tuning (SFT) based on Qwen-VL-Chat and Internlm-XComposer-v2. In the first stage, we perform alignment training with chart and CSV pairs to update the connector parameters. In the second stage, we utilize instruction and chart pairs to fine-tune the LLM branch with LoRA~\cite{LoRA}. Considering that charts are not complex images compared to neutral images, we keep the visual encoder frozen during the SFT process. Please refer to Appendix~\ref{sec_apdx_settings} for detailed experimental settings.

\section{Experiments}
\vspace{-5pt}

We evaluate 18 open-sourced and 3 closed-sourced MLLMs (shown in Tab.~\ref{tab_overview}) on ChartBench. Detailed model architectures and configurations are provided in Appendix~\ref{apdx_sec_mllm_arch}. Notably, some models exhibited poor performance in certain areas, which may be due to suboptimal instruction prompts. We provide a detailed analysis of the model with this anomaly in Appendix~\ref{sec_apdx_model_performance_explain}.

\myparagraph{Results on ChartBench.}
Tab.~\ref{tab_overview} compares various MLLMs on the ChartQA and our ChartBench. Overall, MLLMs show consistent trends across both benchmarks, though individual models vary notably. Onechart~\cite{OneChart} performs well on ChartQA but struggles with ChartBench, extracting incomplete or overly long Python dictionaries, which hampers its LLM (llava-V1.6~\cite{llava}) from following instructions effectively. Qwen~\cite{qwen-vl} and other top-ranked MLLMs demonstrate consistent performance across both metrics, indicating accurate chart comprehension. However, models like BLIP2 and MiniGPT-v2 show significant deviations due to the broader and less standardized output required by NQA compared to Acc+, leading to many extraction failures despite filtering by stronger LLMs~\cite{GPT4, deepseek, qwen}. Unsurprisingly, models generally perform better on regular charts than on extra types, especially those with pre-alignment, such as ChartVLM~\cite{ChartX}, DocOwl~\cite{docowl}, and Internlm-XComposer-v2~\cite{internlm2}, since the alignment process primarily uses regular charts. This indicates that pre-alignment and SFT with chart data effectively enhance chart comprehension abilities.

\myparagraph{Results w.r.t. Task Types.} 
Tab.~\ref{tab_task_type} presents the performance of MLLMs on 5 type tasks, which are introduced in Sec.~\ref{sec_method_qa_gene}. All MLLMs perform exceptionally well on the easiest CR task, demonstrating their ability to recognize basic chart types effectively. LLaVA-v1.5~\cite{llava}, mPLUG-Owl~\cite{Mplug}, and Qwen-VL-Chat~\cite{qwen-vl} demonstrate significant advantages in the VC and GC conception tasks, benefiting from their chart-tuning data. VE is the most challenging task, which is the key distinction between ChartBench and ChartQA. VE task cannot be resolved merely through basic OCR and demands a series of visual and textual logical reasoning steps to reach the ultimate answer. Despite demonstrating strong overall performance, models like BLIP2~\cite{BLIP2} and ChartLlama~\cite{Chartllama} struggle with the VE task. This observation suggests that strong text recognition abilities are insufficient for high chart reasoning capabilities. Closed-source models outperform open-source models, partly due to their larger size and broader data coverage. Additionally, they utilize supplementary recognition tools instead of relying solely on end-to-end inference, as further detailed in Appendix~\ref{apdx_sec_gpt4v_vis}.

\begin{table}[t]
\centering
\caption{The zero-shot performance on ChartQA and our proposed ChartBench. We report average \textit{Acc+} for 4 yes-or-no tasks and GPT-acc for NQA task. Regular: line, pie, and bar plots. Extra: additional chart in Tab.~\ref{tab_chartbench_distribution}. ChartBench is more challenging for more unannotated charts.}
\resizebox{\linewidth}{!}{
\setlength{\tabcolsep}{8pt}
\begin{tabular}{@{}ccccccccccccc@{}}
\toprule[1.5pt]
\multicolumn{1}{c|}{\multirow{3}{*}{Models}} & \multicolumn{8}{c|}{ChartBench} & \multicolumn{4}{c}{ChartQA} \\ \cmidrule(l){2-13} 
\multicolumn{1}{l|}{} & \multicolumn{3}{c|}{Regular Type} & \multicolumn{3}{c|}{Extra Type} & \multirow{2}{*}{Avg.} & \multicolumn{1}{c|}{\multirow{2}{*}{Rank}} & \multirow{2}{*}{Human} & \multirow{2}{*}{Aug.} & \multirow{2}{*}{Avg.} & \multirow{2}{*}{Rank} \\ \cmidrule(lr){2-7}
\multicolumn{1}{l|}{} & \textit{Acc+} & NQA & \multicolumn{1}{c|}{Avg.} & \textit{Acc+} & NQA & \multicolumn{1}{c|}{Avg.} &  & \multicolumn{1}{c|}{} &  &  &  &  \\ \midrule
\multicolumn{13}{l}{\quad \textit{Open source MLLMs}} \\
\multicolumn{1}{l|}{VisualGLM~\cite{glm}} & 3.46 & 1.83 & \multicolumn{1}{c|}{3.13} & 4.22 & 4.84 & \multicolumn{1}{c|}{4.35} & 3.68 & \multicolumn{1}{c|}{\#18} & 18.96 & 6.80 & 12.88 & \#12 \\
\multicolumn{1}{l|}{Shikra~\cite{shikra}} & 8.59 & 2.35 & \multicolumn{1}{c|}{7.34} & 7.50 & 9.05 & \multicolumn{1}{c|}{7.81} & 7.55 & \multicolumn{1}{c|}{\#17} & 16.24 & 7.28 & 11.76 & \#15 \\
\multicolumn{1}{l|}{OneChart~\cite{OneChart}} & 12.34 & 2.26 & \multicolumn{1}{c|}{10.33} & 8.75 & 3.37 & \multicolumn{1}{c|}{7.68} & 9.12 & \multicolumn{1}{c|}{\#16} & \textbf{85.30} & 49.10 & 67.20 & \#5\\
\multicolumn{1}{l|}{InstructBLIP~\cite{InstructBLIP}} & 17.96 & 0.87 & \multicolumn{1}{c|}{14.55} & 5.50 & 5.37 & \multicolumn{1}{c|}{5.47} & 10.43 & \multicolumn{1}{c|}{\#15} & 15.92 & 7.92 & 11.92 & \#14 \\
\multicolumn{1}{l|}{ChartVLM~\cite{ChartX}} & 8.02 & \textbf{43.74} & \multicolumn{1}{c|}{15.24} & 5.92 & 18.21 & \multicolumn{1}{c|}{8.37} & 12.06 & \multicolumn{1}{c|}{\#14} & 42.08 & 82.48 & 62.28 & \#6 \\
\multicolumn{1}{l|}{Internlm-XComposer~\cite{internlm-xcomposer}} & 19.70 & 1.22 & \multicolumn{1}{c|}{16.01} & 10.11 & 5.79 & \multicolumn{1}{c|}{9.25} & 12.94 & \multicolumn{1}{c|}{\#13} & 13.20 & 7.84 & 10.52 & \#16 \\
\multicolumn{1}{l|}{CogVLM-Chat~\cite{cogvlm}} & 14.41 & 12.96 & \multicolumn{1}{c|}{14.12} & 11.89 & 13.68 & \multicolumn{1}{c|}{12.25} & 13.26 & \multicolumn{1}{c|}{\#12} & 34.24 & 28.56 & 31.40 & \#9 \\
\multicolumn{1}{l|}{SPHINX~\cite{sphinx}} & 17.87 & 6.17 & \multicolumn{1}{c|}{15.54} & 17.92 & 12.74 & \multicolumn{1}{c|}{16.89} & 16.13 & \multicolumn{1}{c|}{\#11} & 21.44 & 11.20 & 16.32 & \#11 \\
\multicolumn{1}{l|}{BLIP2~\cite{BLIP2}} & 21.65 & 0.96 & \multicolumn{1}{c|}{17.53} & 18.44 & 4.84 & \multicolumn{1}{c|}{15.74} & 16.70 & \multicolumn{1}{c|}{\#10} & 13.52 & 6.00 & 9.76 & \#17 \\
\multicolumn{1}{l|}{CogAgent~\cite{cogagent}} & 20.39 & 26.61 & \multicolumn{1}{c|}{21.63} & 14.36 & 25.79 & \multicolumn{1}{c|}{16.64} & 19.35 & \multicolumn{1}{c|}{\#9} & 54.08 & 80.56 & 67.32 & \#4 \\
\multicolumn{1}{l|}{MiniGPT-v2~\cite{Minigptv2}} & 22.37 & 2.43 & \multicolumn{1}{c|}{18.40} & 25.06 & 5.26 & \multicolumn{1}{c|}{21.11} & 19.61 & \multicolumn{1}{c|}{\#8} & 15.60 & 8.48 & 12.04 & \#13 \\
\multicolumn{1}{l|}{ChartLlama~\cite{Chartllama}} & 22.02 & 16.87 & \multicolumn{1}{c|}{21.00} & 22.56 & 18.32 & \multicolumn{1}{c|}{21.71} & 21.30 & \multicolumn{1}{c|}{\#7} & 58.40 & \textbf{93.12} & \textbf{75.76} & \#1 \\
\multicolumn{1}{l|}{mPLUG-Owl-bloomz~\cite{Mplug}} & 27.80 & 2.35 & \multicolumn{1}{c|}{22.73} & 25.47 & 6.21 & \multicolumn{1}{c|}{21.64} & 22.21 & \multicolumn{1}{c|}{\#6} & 7.84 & 4.88 & 6.36 & \#18 \\
\multicolumn{1}{l|}{LLaVA-v1.5~\cite{llava}} & 25.61 & 8.09 & \multicolumn{1}{c|}{22.12} & 27.39 & 15.26 & \multicolumn{1}{c|}{24.97} & 23.39 & \multicolumn{1}{c|}{\#5} & 22.64 & 13.04 & 17.84 & \#10 \\
\multicolumn{1}{l|}{Qwen-VL-Chat~\cite{qwen-vl}} & 29.46 & 23.57 & \multicolumn{1}{c|}{28.28} & 26.56 & 21.05 & \multicolumn{1}{c|}{25.46} & 26.98 & \multicolumn{1}{c|}{\#4} & 42.48 & 75.20 & 58.84 & \#7 \\
\multicolumn{1}{l|}{DocOwl-v1.5~\cite{docowl}} & 35.27 & 37.30 & \multicolumn{1}{c|}{35.67} & 26.86 & 29.47 & \multicolumn{1}{c|}{27.38} & 31.89 & \multicolumn{1}{c|}{\#3} & 48.24 & 86.72 & 67.48 & \#3 \\
\multicolumn{1}{l|}{Mini-Gemini~\cite{minigemini}} & 39.57 & 25.57 & \multicolumn{1}{c|}{36.78} & 31.81 & 25.79 & \multicolumn{1}{c|}{30.61} & 33.96 & \multicolumn{1}{c|}{\#2} & 44.32 & 57.04 & 50.68 & \#8 \\
\multicolumn{1}{l|}{Internlm-XComposer-v2~\cite{internlm2}} & \textbf{57.89} & 40.96 & \multicolumn{1}{c|}{\textbf{54.52}} & \textbf{41.75} & \textbf{31.58} & \multicolumn{1}{c|}{\textbf{39.73}} & \textbf{47.78} & \multicolumn{1}{c|}{\#1} & 63.12 & 81.92 & 72.64 & \#2 \\ \midrule
\multicolumn{13}{l}{\quad \textit{Closed source MLLMs}} \\
\multicolumn{1}{l|}{ERNIE~\cite{wenxin}} & 47.39 & 25.74 & \multicolumn{1}{c|}{43.08} & 46.39 & 33.37 & \multicolumn{1}{c|}{43.82} & 43.37 & \multicolumn{1}{c|}{\#3} & - & - & - &  \\
\multicolumn{1}{l|}{GPT-4V~\cite{GPT4}} & 53.26 & 33.04 & \multicolumn{1}{c|}{49.23} & 55.83 & 40.00 & \multicolumn{1}{c|}{52.68} & 50.74 & \multicolumn{1}{c|}{\#2} & - & - & 78.50 & \#2 \\
\multicolumn{1}{l|}{GPT-4O~\cite{GPT4}} & \textbf{65.00} & \textbf{40.00} & \multicolumn{1}{c|}{\textbf{60.02}} & \textbf{63.33} & \textbf{41.05} & \multicolumn{1}{c|}{\textbf{58.89}} & \textbf{59.45} & \multicolumn{1}{c|}{\#1} & - & - & \textbf{85.70} & \#1 \\ 
\bottomrule[1.5pt]
\end{tabular}
}
\vspace{-15pt}
\label{tab_overview}
\end{table}
\begin{table}[t]
\centering
\caption{The zero-shot performance w.r.t. task types, i.e., Chart Recognition (CR), Value Extraction (VE), Value Comparison (VC), Global Conception (GC), and Number QA (NQA). $\uparrow$ / $\downarrow$ indicates that higher/lower is the better, respectively.}
\resizebox{\linewidth}{!}{
\setlength{\tabcolsep}{12pt}
\begin{tabular}{@{}ccccccccccc@{}}
\toprule[1.5pt]
\multicolumn{1}{c|}{\multirow{2}{*}{Models}} & \multicolumn{2}{c|}{CR} & \multicolumn{2}{c|}{VE} & \multicolumn{2}{c|}{VC} & \multicolumn{2}{c|}{GC} & \multicolumn{1}{c|}{\multirow{2}{*}{NQA$\uparrow$}} & \multirow{2}{*}{Avg.$\uparrow$} \\ \cmidrule(lr){2-9}
\multicolumn{1}{l|}{} & \textit{Acc+} $\uparrow$ & \multicolumn{1}{c|}{\textit{CoR}$\downarrow$} & \textit{Acc+}$\uparrow$ & \multicolumn{1}{c|}{\textit{CoR}$\downarrow$} & \textit{Acc+}$\uparrow$ & \multicolumn{1}{c|}{\textit{CoR}$\downarrow$} & \textit{Acc+}$\uparrow$ & \multicolumn{1}{c|}{\textit{CoR}$\downarrow$} & \multicolumn{1}{c|}{} &  \\ \midrule
\multicolumn{11}{l}{\quad \textit{Open source MLLMs}} \\
\multicolumn{1}{l|}{VisualGLM~\cite{glm}} & 16.29 & \multicolumn{1}{c|}{79.19} & 0.00 & \multicolumn{1}{c|}{99.67} & 0.00 & \multicolumn{1}{c|}{99.81} & 0.00 & \multicolumn{1}{c|}{99.71} & \multicolumn{1}{c|}{3.19} & 3.68 \\
\multicolumn{1}{l|}{Shikra~\cite{shikra}} & 2.10 & \multicolumn{1}{c|}{93.57} & 11.90 & \multicolumn{1}{c|}{80.71} & 10.62 & \multicolumn{1}{c|}{87.71} & 7.86 & \multicolumn{1}{c|}{82.71} & \multicolumn{1}{c|}{5.38} & 7.55 \\
\multicolumn{1}{l|}{OneChart~\cite{OneChart}} & 3.71 & \multicolumn{1}{c|}{94.33} & 15.48 & \multicolumn{1}{c|}{82.14} & 17.57 & \multicolumn{1}{c|}{73.71} & 11.38 & \multicolumn{1}{c|}{85.67} & \multicolumn{1}{c|}{2.76} & 9.12 \\
\multicolumn{1}{l|}{InstructBLIP~\cite{InstructBLIP}} & 49.57 & \multicolumn{1}{c|}{36.67} & 0.00 & \multicolumn{1}{c|}{100.00} & 0.05 & \multicolumn{1}{c|}{99.81} & 0.00 & \multicolumn{1}{c|}{99.90} & \multicolumn{1}{c|}{2.90} & 10.43 \\
\multicolumn{1}{l|}{ChartVLM~\cite{ChartX}} & 0.00 & \multicolumn{1}{c|}{100.00} & 9.05 & \multicolumn{1}{c|}{85.48} & 10.05 & \multicolumn{1}{c|}{83.81} & 8.52 & \multicolumn{1}{c|}{86.19} & \multicolumn{1}{c|}{32.19} & 12.06 \\
\multicolumn{1}{l|}{Internlm-XComposer~\cite{internlm-xcomposer}} & 42.29 & \multicolumn{1}{c|}{56.95} & 6.86 & \multicolumn{1}{c|}{85.14} & 2.48 & \multicolumn{1}{c|}{96.57} & 9.67 & \multicolumn{1}{c|}{78.48} & \multicolumn{1}{c|}{3.29} & 12.94 \\
\multicolumn{1}{l|}{CogVLM-Chat~\cite{cogvlm}} & 29.14 & \multicolumn{1}{c|}{69.33} & 2.81 & \multicolumn{1}{c|}{94.29} & 14.19 & \multicolumn{1}{c|}{80.71} & 7.33 & \multicolumn{1}{c|}{90.14} & \multicolumn{1}{c|}{13.29} & 13.26 \\
\multicolumn{1}{l|}{SPHINX~\cite{sphinx}} & 38.48 & \multicolumn{1}{c|}{51.38} & 10.38 & \multicolumn{1}{c|}{80.67} & 14.33 & \multicolumn{1}{c|}{77.38} & 9.62 & \multicolumn{1}{c|}{80.90} & \multicolumn{1}{c|}{9.14} & 16.13 \\
\multicolumn{1}{l|}{BLIP2~\cite{BLIP2}} & 60.05 & \multicolumn{1}{c|}{37.05} & 4.24 & \multicolumn{1}{c|}{89.29} & 14.05 & \multicolumn{1}{c|}{78.86} & 3.86 & \multicolumn{1}{c|}{90.00} & \multicolumn{1}{c|}{2.71} & 16.70 \\
\multicolumn{1}{l|}{MiniGPT-v2~\cite{Minigptv2}} & 29.05 & \multicolumn{1}{c|}{49.24} & 22.00 & \multicolumn{1}{c|}{\textbf{55.14}} & 24.29 & \multicolumn{1}{c|}{53.33} & 18.10 & \multicolumn{1}{c|}{61.76} & \multicolumn{1}{c|}{3.71} & 19.35 \\
\multicolumn{1}{l|}{CogAgent~\cite{cogagent}} & 62.57 & \multicolumn{1}{c|}{37.10} & 1.19 & \multicolumn{1}{c|}{94.90} & 7.33 & \multicolumn{1}{c|}{88.24} & 1.19 & \multicolumn{1}{c|}{94.76} & \multicolumn{1}{c|}{26.24} & 19.61 \\
\multicolumn{1}{l|}{ChartLlama~\cite{Chartllama}} & 49.86 & \multicolumn{1}{c|}{44.19} & 8.38 & \multicolumn{1}{c|}{84.14} & 20.43 & \multicolumn{1}{c|}{69.48} & 10.67 & \multicolumn{1}{c|}{83.81} & \multicolumn{1}{c|}{17.52} & 21.30 \\
\multicolumn{1}{l|}{mPLUG-Owl-bloomz~\cite{Mplug}} & 32.33 & \multicolumn{1}{c|}{51.24} & 23.14 & \multicolumn{1}{c|}{76.76} & 25.33 & \multicolumn{1}{c|}{69.29} & 26.48 & \multicolumn{1}{c|}{71.00} & \multicolumn{1}{c|}{4.10} & 22.21 \\
\multicolumn{1}{l|}{LLaVA-v1.5~\cite{llava}} & 47.86 & \multicolumn{1}{c|}{36.24} & 15.81 & \multicolumn{1}{c|}{66.24} & 26.05 & \multicolumn{1}{c|}{56.48} & 16.52 & \multicolumn{1}{c|}{66.57} & \multicolumn{1}{c|}{11.33} & 23.39 \\
\multicolumn{1}{l|}{Qwen-VL-Chat~\cite{qwen-vl}} & 51.67 & \multicolumn{1}{c|}{42.71} & 11.14 & \multicolumn{1}{c|}{84.57} & 27.29 & \multicolumn{1}{c|}{63.14} & 21.71 & \multicolumn{1}{c|}{74.86} & \multicolumn{1}{c|}{22.43} & 26.98 \\
\multicolumn{1}{l|}{DocOwl-v1.5~\cite{docowl}} & 30.43 & \multicolumn{1}{c|}{65.05} & 34.48 & \multicolumn{1}{c|}{58.24} & 31.10 & \multicolumn{1}{c|}{55.19} & 30.48 & \multicolumn{1}{c|}{63.19} & \multicolumn{1}{c|}{33.76} & 31.89 \\
\multicolumn{1}{l|}{Mini-Gemini~\cite{minigemini}} & \textbf{80.52} & \multicolumn{1}{c|}{\textbf{17.86}} & 17.62 & \multicolumn{1}{c|}{70.43} & 26.00 & \multicolumn{1}{c|}{59.38} & 22.00 & \multicolumn{1}{c|}{71.10} & \multicolumn{1}{c|}{25.67} & 33.96 \\
\multicolumn{1}{l|}{Internlm-XComposer-v2~\cite{internlm2}} & 68.29 & \multicolumn{1}{c|}{30.24} & \textbf{36.63} & \multicolumn{1}{c|}{57.71} & \textbf{54.63} & \multicolumn{1}{c|}{\textbf{27.71}} & \textbf{45.80} & \multicolumn{1}{c|}{\textbf{51.46}} & \multicolumn{1}{c|}{\textbf{36.71}} & \textbf{47.78} \\ \midrule
\multicolumn{11}{l}{\quad \textit{Closed source MLLMs}} \\
\multicolumn{1}{l|}{ERNIE~\cite{wenxin}} & 65.24 & \multicolumn{1}{c|}{19.52} & \textbf{44.76} & \multicolumn{1}{c|}{\textbf{44.76}} & 32.86 & \multicolumn{1}{c|}{41.43} & 47.14 & \multicolumn{1}{c|}{47.62} & \multicolumn{1}{c|}{29.24} & 43.37 \\
\multicolumn{1}{l|}{GPT-4V~\cite{GPT4}} & 96.19 & \multicolumn{1}{c|}{2.86} & 30.95 & \multicolumn{1}{c|}{63.33} & 48.57 & \multicolumn{1}{c|}{34.76} & 46.19 & \multicolumn{1}{c|}{47.62} & \multicolumn{1}{c|}{36.19} & 50.74 \\
\multicolumn{1}{l|}{GPT-4O~\cite{GPT4}} & \textbf{97.62} & \multicolumn{1}{c|}{\textbf{1.43}} & 43.33 & \multicolumn{1}{c|}{\textbf{44.76}} & \textbf{66.19} & \multicolumn{1}{c|}{\textbf{16.19}} & \textbf{53.33} & \multicolumn{1}{c|}{\textbf{41.43}} & \multicolumn{1}{c|}{\textbf{40.48}} & \textbf{59.45} \\ \bottomrule[1.5pt]
\end{tabular}
}
\vspace{-20pt}
\label{tab_task_type}
\end{table}
\begin{table}[ht]
\centering
\caption{The zero-shot \textit{CoR} (\%) performance w.r.t. chart types. Higher \textit{CoR} means more severe hallucinations. \textit{CoR} and \textit{Acc+} exhibit a negative correlation.}
\resizebox{\linewidth}{!}{
\setlength{\tabcolsep}{8pt}
\begin{tabular}{@{}ccccccccccccc@{}}
\toprule[1.5pt]
\multicolumn{1}{c|}{\multirow{2}{*}{Models}} & \multicolumn{4}{c|}{Regular Type} & \multicolumn{7}{c|}{Extra Type} & \multirow{2}{*}{\textit{CoR}} \\ \cmidrule(lr){2-12}
\multicolumn{1}{l|}{} & Line & Bar & Pie & \multicolumn{1}{c|}{Avg.} & Area & Box & Radar & Scatter & Node & Combin. & \multicolumn{1}{c|}{Avg.} &  \\ \midrule
\multicolumn{13}{l}{\quad \textit{Open source MLLMs}} \\
\multicolumn{1}{l|}{VisualGLM~\cite{glm}} & 89.20 & 98.04 & 99.38 & \multicolumn{1}{c|}{96.27} & 93.50 & 90.50 & 97.50 & 91.33 & 80.50 & 94.62 & \multicolumn{1}{c|}{92.39} & 94.60 \\
\multicolumn{1}{l|}{ChartVLM~\cite{ChartX}} & 85.80 & 87.46 & 92.25 & \multicolumn{1}{c|}{87.95} & 88.00 & 90.33 & 89.88 & 91.17 & 91.00 & 89.50 & \multicolumn{1}{c|}{89.83} & 88.87 \\
\multicolumn{1}{l|}{Shikra~\cite{shikra}} & 85.80 & 82.19 & 98.25 & \multicolumn{1}{c|}{85.93} & 84.83 & 85.00 & 86.00 & 84.33 & 72.00 & 95.38 & \multicolumn{1}{c|}{85.89} & 86.18 \\
\multicolumn{1}{l|}{InstructBLIP~\cite{InstructBLIP}} & 75.50 & 82.58 & 79.50 & \multicolumn{1}{c|}{80.41} & 88.33 & 85.50 & 91.00 & 86.00 & 90.50 & 89.62 & \multicolumn{1}{c|}{88.58} & 84.10 \\
\multicolumn{1}{l|}{OneChart~\cite{OneChart}} & 80.10 & 84.46 & 89.38 & \multicolumn{1}{c|}{84.36} & 89.83 & 87.33 & 93.62 & 90.17 & 33.50 & 89.38 & \multicolumn{1}{c|}{87.08} & 83.96 \\
\multicolumn{1}{l|}{CogVLM-Chat~\cite{cogvlm}} & 87.20 & 83.38 & 79.38 & \multicolumn{1}{c|}{83.52} & 85.33 & 86.67 & 77.88 & 84.17 & 79.50 & 89.88 & \multicolumn{1}{c|}{84.13} & 83.62 \\
\multicolumn{1}{l|}{Internlm-XComposer~\cite{internlm-xcomposer}} & 79.40 & 73.92 & 68.62 & \multicolumn{1}{c|}{74.20} & 93.33 & 79.83 & 77.00 & 84.17 & 91.00 & 92.25 & \multicolumn{1}{c|}{85.84} & 79.29 \\
\multicolumn{1}{l|}{CogAgent~\cite{cogagent}} & 81.40 & 76.00 & 89.00 & \multicolumn{1}{c|}{79.59} & 84.33 & 82.67 & 90.12 & 87.50 & \textbf{7.00} & 84.00 & \multicolumn{1}{c|}{81.50} & 78.75 \\
\multicolumn{1}{l|}{mPLUG-Owl-bloomz~\cite{Mplug}} & 69.20 & 79.54 & 76.12 & \multicolumn{1}{c|}{76.57} & 82.50 & 78.50 & 80.00 & 77.83 & 70.00 & 77.50 & \multicolumn{1}{c|}{78.24} & 77.35 \\
\multicolumn{1}{l|}{BLIP2~\cite{BLIP2}} & 66.40 & 79.96 & 72.75 & \multicolumn{1}{c|}{75.57} & 92.50 & 85.83 & 78.12 & 73.17 & 16.00 & 66.88 & \multicolumn{1}{c|}{71.92} & 73.80 \\
\multicolumn{1}{l|}{SPHINX~\cite{sphinx}} & 73.80 & 75.73 & 58.00 & \multicolumn{1}{c|}{72.07} & 82.00 & 86.17 & 71.00 & 73.17 & 63.50 & 65.25 & \multicolumn{1}{c|}{73.47} & 72.58 \\
\multicolumn{1}{l|}{ChartLlama~\cite{Chartllama}} & 65.60 & 74.27 & 74.50 & \multicolumn{1}{c|}{72.34} & 81.50 & 78.83 & 72.62 & 66.00 & 28.50 & 68.62 & \multicolumn{1}{c|}{68.47} & 70.40 \\
\multicolumn{1}{l|}{Qwen-VL-Chat~\cite{qwen-vl}} & 56.00 & 73.62 & 57.50 & \multicolumn{1}{c|}{66.68} & 68.67 & 66.67 & 57.25 & 74.50 & 74.00 & 66.25 & \multicolumn{1}{c|}{66.92} & 66.32 \\
\multicolumn{1}{l|}{DocOwl-v1.5~\cite{docowl}} & 47.10 & 63.69 & 63.62 & \multicolumn{1}{c|}{59.91} & 80.50 & 61.33 & 64.62 & 59.67 & 53.00 & 52.00 & \multicolumn{1}{c|}{62.44} & 60.42 \\
\multicolumn{1}{l|}{LLaVA-v1.5~\cite{llava}} & 51.20 & 59.69 & 54.87 & \multicolumn{1}{c|}{56.89} & 61.67 & 58.50 & 60.00 & 59.17 & 29.00 & 56.00 & \multicolumn{1}{c|}{55.79} & 56.38 \\
\multicolumn{1}{l|}{MiniGPT-v2~\cite{Minigptv2}} & 52.20 & 57.35 & 56.75 & \multicolumn{1}{c|}{56.07} & 57.17 & \textbf{56.00} & 52.75 & 51.50 & 47.00 & 54.25 & \multicolumn{1}{c|}{53.47} & 54.87 \\
\multicolumn{1}{l|}{Mini-Gemini~\cite{minigemini}} & 55.70 & 53.92 & 51.25 & \multicolumn{1}{c|}{53.84} & \textbf{53.50} & 62.67 & 57.75 & 50.83 & 61.00 & 57.88 & \multicolumn{1}{c|}{56.92} & 54.69 \\
\multicolumn{1}{l|}{Internlm-XComposer-v2~\cite{internlm2}} & \textbf{27.40} & \textbf{44.65} & \textbf{32.50} & \multicolumn{1}{c|}{\textbf{38.52}} & 55.33 & 58.33 & \textbf{47.88} & \textbf{43.17} & 29.00 & \textbf{39.75} & \multicolumn{1}{c|}{\textbf{47.22}} & \textbf{41.78} \\ \midrule
\multicolumn{13}{l}{\quad \textit{Closed source MLLMs}} \\
\multicolumn{1}{l|}{ERNIE~\cite{wenxin}} & 34.00 & 41.15 & 27.50 & \multicolumn{1}{c|}{37.05} & \textbf{46.67} & 45.00 & 51.25 & 33.33 & 25.00 & 33.75 & \multicolumn{1}{c|}{40.26} & 38.33 \\
\multicolumn{1}{l|}{GPT-4V~\cite{GPT4}} & 21.00 & 52.69 & 37.50 & \multicolumn{1}{c|}{42.73} & 58.33 & 38.33 & 23.75 & 25.00 & \textbf{0.00} & 33.75 & \multicolumn{1}{c|}{31.32} & 37.14 \\
\multicolumn{1}{l|}{GPT-4O~\cite{GPT4}} & \textbf{9.00} & \textbf{37.31} & \textbf{20.00} & \multicolumn{1}{c|}{\textbf{27.73}} & 50.00 & \textbf{28.33} & \textbf{20.00} & \textbf{16.67} & \textbf{0.00} & \textbf{28.75} & \multicolumn{1}{c|}{\textbf{25.26}} & \textbf{25.95} \\ \bottomrule[1.5pt]
\end{tabular}
}
\vspace{-15pt}
\label{tab_cor}
\end{table}

\myparagraph{Error analysis.} 
Tab.~\ref{tab_cor} presents the results on \textit{CoR}, which reflects the MLLM's failure to utilize chart information. We find that existing MLLMs tend to give identical answers to similar questions about charts. Internlm-XComposer-v2~\cite{internlm2} shows the lowest CoR (41.78\%), which means nearly half of the responses fail to distinguish between positive and negative questions. This indicates that random guessing without the chart is common among open-source models due to their inability to utilize chart information. 
\textit{CoR} generally shows a negative correlation with performance, although there are exceptions. Qwen~\cite{qwen-vl} demonstrates better \textit{Acc+} compared to MiniGPT-v2~\cite{Minigptv2} with higher \textit{CoR}. For closed-source MLLMs, although GPT-4V~\cite{GPT4} outperforms ERNIE~\cite{wenxin} in terms of \textit{Acc+}, their \textit{CoR} are similar. More granular analysis reveals that ERNIE performs better on challenging VE tasks, which is the weakest area for GPT-4V.

\begin{table}[ht]
\centering
\caption{The performance on with and without annotation charts. \textit{w/i} and \textit{w/o} indicate with and without annotation, respectively. $\dagger$: \textit{Acc+}. $\ddagger$: GPT-acc. MLLMs are better with annotated charts.}
\resizebox{\linewidth}{!}{
\setlength{\tabcolsep}{7pt}
\begin{tabular}{@{}cccccccccccccc@{}}
\toprule[1.5pt]
\multicolumn{1}{c|}{\multirow{2}{*}{Models}} & \multicolumn{2}{c|}{CR$\dagger$} & \multicolumn{2}{c|}{VE$\dagger$} & \multicolumn{2}{c|}{VC$\dagger$} & \multicolumn{2}{c|}{GC$\dagger$} & \multicolumn{2}{c|}{NQA$\ddagger$} & \multicolumn{3}{c}{Avg.} \\ \cmidrule(l){2-14} 
\multicolumn{1}{l|}{} & \textit{w/i} & \multicolumn{1}{c|}{\textit{w/o}} & \textit{w/i} & \multicolumn{1}{c|}{\textit{w/o}} & \textit{w/i} & \multicolumn{1}{c|}{\textit{w/o}} & \textit{w/i} & \multicolumn{1}{c|}{\textit{w/o}} & \textit{w/i} & \multicolumn{1}{c|}{\textit{w/o}} & \textit{w/i} & \textit{w/o} & $\Delta$ \\ \midrule
\multicolumn{14}{l}{\quad \textit{Open source MLLMs}} \\ 
\multicolumn{1}{l|}{Internlm-XComposer~\cite{internlm-xcomposer}} & 30.50 & \multicolumn{1}{c|}{53.00} & 8.00 & \multicolumn{1}{c|}{7.50} & 1.00 & \multicolumn{1}{c|}{2.75} & 10.00 & \multicolumn{1}{c|}{8.50} & 10.60 & \multicolumn{1}{c|}{1.75} & 12.02 & 14.70 & -2.68 \\
\multicolumn{1}{l|}{mPLUG-Owl-bloomz~\cite{Mplug}} & 37.50 & \multicolumn{1}{c|}{41.50} & 22.50 & \multicolumn{1}{c|}{27.50} & 27.25 & \multicolumn{1}{c|}{30.25} & 27.50 & \multicolumn{1}{c|}{29.25} & 9.40 & \multicolumn{1}{c|}{3.75} & 24.83 & 26.45 & -1.62 \\
\multicolumn{1}{l|}{Shikra~\cite{shikra}} & 1.25 & \multicolumn{1}{c|}{1.00} & 6.00 & \multicolumn{1}{c|}{10.25} & 2.25 & \multicolumn{1}{c|}{6.50} & 5.00 & \multicolumn{1}{c|}{8.50} & 15.80 & \multicolumn{1}{c|}{1.50} & 6.06 & 5.55 & +0.51 \\
\multicolumn{1}{l|}{MiniGPT-v2~\cite{Minigptv2}} & 31.25 & \multicolumn{1}{c|}{31.00} & 24.50 & \multicolumn{1}{c|}{22.25} & 27.25 & \multicolumn{1}{c|}{26.50} & 16.50 & \multicolumn{1}{c|}{19.75} & 7.80 & \multicolumn{1}{c|}{2.75} & 21.46 & 20.45 & +1.01 \\
\multicolumn{1}{l|}{ChartVLM~\cite{ChartX}} & 0.00 & \multicolumn{1}{c|}{0.00} & 12.22 & \multicolumn{1}{c|}{10.00} & 9.33 & \multicolumn{1}{c|}{11.00} & 12.44 & \multicolumn{1}{c|}{10.25} & 57.00 & \multicolumn{1}{c|}{\textbf{46.50}} & 18.20 & 15.55 & +2.65 \\
\multicolumn{1}{l|}{InstructBLIP~\cite{InstructBLIP}} & 59.50 & \multicolumn{1}{c|}{54.75} & 0.00 & \multicolumn{1}{c|}{0.00} & 0.25 & \multicolumn{1}{c|}{0.00} & 0.00 & \multicolumn{1}{c|}{0.00} & 10.40 & \multicolumn{1}{c|}{1.00} & 14.03 & 11.15 & +2.88 \\
\multicolumn{1}{l|}{BLIP2~\cite{BLIP2}} & 78.25 & \multicolumn{1}{c|}{69.00} & 4.00 & \multicolumn{1}{c|}{5.00} & 26.50 & \multicolumn{1}{c|}{21.50} & 5.00 & \multicolumn{1}{c|}{7.50} & 6.80 & \multicolumn{1}{c|}{1.75} & 24.11 & 20.95 & +3.16 \\
\multicolumn{1}{l|}{VisualGLM~\cite{glm}} & 24.75 & \multicolumn{1}{c|}{16.25} & 0.00 & \multicolumn{1}{c|}{0.00} & 0.00 & \multicolumn{1}{c|}{0.00} & 0.00 & \multicolumn{1}{c|}{0.00} & 9.20 & \multicolumn{1}{c|}{0.75} & 6.79 & 3.40 & +3.39 \\
\multicolumn{1}{l|}{SPHINX~\cite{sphinx}} & 43.75 & \multicolumn{1}{c|}{41.00} & 11.75 & \multicolumn{1}{c|}{12.25} & 18.50 & \multicolumn{1}{c|}{17.00} & 15.00 & \multicolumn{1}{c|}{8.75} & 23.00 & \multicolumn{1}{c|}{5.25} & 22.40 & 16.85 & +5.55 \\
\multicolumn{1}{l|}{DocOwl-v1.5~\cite{docowl}} & 47.11 & \multicolumn{1}{c|}{49.25} & 60.89 & \multicolumn{1}{c|}{\textbf{42.25}} & 43.11 & \multicolumn{1}{c|}{41.50} & 38.22 & \multicolumn{1}{c|}{43.75} & 61.60 & \multicolumn{1}{c|}{40.75} & 50.19 & 43.50 & +6.69 \\
\multicolumn{1}{l|}{LLaVA-v1.5~\cite{llava}} & 55.25 & \multicolumn{1}{c|}{43.50} & 17.75 & \multicolumn{1}{c|}{16.00} & 28.50 & \multicolumn{1}{c|}{31.50} & 15.50 & \multicolumn{1}{c|}{16.25} & 31.80 & \multicolumn{1}{c|}{5.50} & 29.76 & 22.55 & +7.21 \\
\multicolumn{1}{l|}{CogVLM-Chat~\cite{cogvlm}} & 31.25 & \multicolumn{1}{c|}{27.00} & 3.50 & \multicolumn{1}{c|}{2.00} & 22.75 & \multicolumn{1}{c|}{19.25} & 14.00 & \multicolumn{1}{c|}{9.00} & 37.40 & \multicolumn{1}{c|}{5.75} & 21.78 & 12.60 & +9.18 \\
\multicolumn{1}{l|}{OneChart~\cite{OneChart}} & 4.00 & \multicolumn{1}{c|}{3.50} & 36.67 & \multicolumn{1}{c|}{14.50} & 21.78 & \multicolumn{1}{c|}{16.00} & 25.11 & \multicolumn{1}{c|}{9.25} & 4.40 & \multicolumn{1}{c|}{2.25} & 18.39 & 9.10 & +9.29 \\
\multicolumn{1}{l|}{ChartLlama~\cite{Chartllama}} & 57.00 & \multicolumn{1}{c|}{53.50} & 15.75 & \multicolumn{1}{c|}{7.00} & 33.00 & \multicolumn{1}{c|}{24.25} & 20.00 & \multicolumn{1}{c|}{13.00} & 42.20 & \multicolumn{1}{c|}{12.75} & 33.59 & 22.10 & +11.49 \\
\multicolumn{1}{l|}{CogAgent~\cite{cogagent}} & 64.67 & \multicolumn{1}{c|}{64.75} & 2.89 & \multicolumn{1}{c|}{0.00} & 16.00 & \multicolumn{1}{c|}{13.25} & 2.44 & \multicolumn{1}{c|}{0.25} & 61.60 & \multicolumn{1}{c|}{11.50} & 29.52 & 17.95 & +11.57 \\
\multicolumn{1}{l|}{Mini-Gemini~\cite{minigemini}} & 79.33 & \multicolumn{1}{c|}{\textbf{74.00}} & 20.00 & \multicolumn{1}{c|}{16.75} & 32.89 & \multicolumn{1}{c|}{33.75} & 30.89 & \multicolumn{1}{c|}{22.00} & 59.20 & \multicolumn{1}{c|}{14.75} & 44.46 & 32.25 & +12.21 \\
\multicolumn{1}{l|}{Qwen-VL-Chat~\cite{qwen-vl}} & 68.00 & \multicolumn{1}{c|}{53.50} & 26.50 & \multicolumn{1}{c|}{7.50} & 47.75 & \multicolumn{1}{c|}{35.00} & 31.50 & \multicolumn{1}{c|}{33.50} & 54.80 & \multicolumn{1}{c|}{14.00} & 45.71 & 28.70 & +17.01 \\
\multicolumn{1}{l|}{Internlm-XComposer-v2~\cite{internlm2}} & \textbf{83.00} & \multicolumn{1}{c|}{64.25} & \textbf{75.25} & \multicolumn{1}{c|}{39.75} & \textbf{70.00} & \multicolumn{1}{c|}{\textbf{66.00}} & \textbf{67.75} & \multicolumn{1}{c|}{\textbf{66.25}} & \textbf{69.80} & \multicolumn{1}{c|}{37.75} & \textbf{73.16} & \textbf{54.80} & \textbf{+18.36} \\ \midrule
\multicolumn{14}{l}{\quad \textit{Closed source MLLMs}} \\
\multicolumn{1}{l|}{ERNIE~\cite{wenxin}} & 67.50 & \multicolumn{1}{c|}{72.50} & 32.50 & \multicolumn{1}{c|}{\textbf{45.00}} & 42.50 & \multicolumn{1}{c|}{37.50} & 52.50 & \multicolumn{1}{c|}{52.50} & 52.20 & \multicolumn{1}{c|}{7.25} & 49.44 & 42.95 & +6.49 \\
\multicolumn{1}{l|}{GPT-4O~\cite{GPT4}} & \textbf{95.00} & \multicolumn{1}{c|}{95.00} & \textbf{87.50} & \multicolumn{1}{c|}{37.50} & \textbf{72.50} & \multicolumn{1}{c|}{\textbf{80.00}} & \textbf{87.50} & \multicolumn{1}{c|}{\textbf{60.00}} & 74.00 & \multicolumn{1}{c|}{\textbf{32.50}} & \textbf{83.30} & \textbf{61.00} & +22.30 \\ 
\multicolumn{1}{l|}{GPT-4V~\cite{GPT4}} & 92.50 & \multicolumn{1}{c|}{\textbf{97.50}} & 72.50 & \multicolumn{1}{c|}{7.50} & 67.50 & \multicolumn{1}{c|}{57.50} & 72.50 & \multicolumn{1}{c|}{37.50} & \textbf{82.00} & \multicolumn{1}{c|}{15.00} & 77.40 & 43.00 & \textbf{+34.40} \\
\bottomrule[1.5pt]
\end{tabular}
}
\vspace{-15pt}
\label{tab_data_anno}
\end{table}
\begin{table}[t]
\centering
\caption{Performance gain of chart chain of thought on various MLLMs. CoTs prove to be simple and effective ways to improve the performance on ChartBench. $\dagger$: \textit{Acc+}. $\ddagger$: GPT-acc.}
\resizebox{\linewidth}{!}{
\setlength{\tabcolsep}{4pt}
\begin{tabular}{@{}c|l|lll||lllll|l@{}}
\toprule[1.5pt]
Models & Method & \textit{w/i} & \textit{w/o} & $\Delta$ & CR$\dagger$ & VE$\dagger$ & VC$\dagger$ & GC$\dagger$ & NQA$\ddagger$ & Avg. \\ \midrule
\multirow{4}{*}{MiniGPT-v2~\cite{Minigptv2}} & Base & 21.46 & 20.45 & 1.01 & 29.02 & 22.29 & 24.59 & 18.29 & 3.71 & 19.58 \\
 & CoT-fix & 25.25\good{+3.79} & 21.33\good{+0.88} & 3.92\bad{+2.91} & 36.76\good{+7.74} & 29.22\good{+6.93} & 25.14\good{+0.55} & 26.37\good{+8.08} & 5.20\good{+1.49} & 24.54\good{+4.96} \\
 & CoT-self & 22.44\good{+0.98} & 20.12\bad{-0.33} & 2.32\bad{+1.31} & 34.52\good{+5.50} & 27.83\good{+5.54} & 26.02\good{+1.43} & 24.44\good{+6.15} & 4.40\good{+0.69} & 23.44\good{+3.86} \\
 & CoT-GPT & 26.66\good{+5.20} & 21.52\good{+1.07} & 5.14\bad{+4.13} & 37.72\good{+8.70} & 29.31\good{+7.02} & 26.66\good{+2.07} & 27.62\good{+9.33} & 5.55\good{+1.84} & 25.37\good{+5.79} \\ \midrule
\multirow{4}{*}{Qwen-VL-Chat~\cite{qwen-vl}} & Base & 45.71 & 28.70 & 17.01 & 52.54 & 10.78 & 27.46 & 21.95 & 22.43 & 27.03 \\
 & CoT-fix & 50.12\good{+4.42} & 29.80\good{+1.10} & 20.32\bad{+3.31} & 64.54\good{+12.00} & 15.85\good{+5.07} & 28.44\good{+0.98} & 29.22\good{+7.27} & 24.98\good{+2.55} & 32.61\good{+5.58} \\
 & CoT-self & 47.77\good{+2.07} & 26.74\bad{-1.96} & 21.03\bad{+4.02} & 56.52\good{+3.98} & 11.24\good{+0.46} & 26.42\bad{-1.04} & 24.33\good{+2.38} & 22.64\good{+0.21} & 28.23\good{+1.20} \\
 & CoT-GPT & 51.22\good{+5.52} & 30.02\good{+1.32} & 21.20\bad{+4.19} & 66.64\good{+14.10} & 16.02\good{+5.24} & 29.33\good{+1.87} & 28.82\good{+6.87} & 26.72\good{+4.29} & 33.51\good{+6.48} \\ \midrule
\multirow{4}{*}{Internlm-XComposer-v2~\cite{internlm2}} & Base & 73.16 & 54.80 & 18.36 & 68.29 & 36.63 & 54.63 & 45.80 & 36.71 & 48.41 \\
 & CoT-fix & 75.22\good{+2.06} & 55.74\good{+0.94} & 19.48\bad{+1.12} & 69.22\good{+0.93} & 36.76\good{+0.13} & 58.23\good{+3.60} & 46.11\good{+0.31} & 36.52\bad{-0.19} & 49.37\good{+0.96} \\
 & CoT-self & 73.54\good{+0.38} & 54.62\bad{-0.18} & 18.92\bad{+0.56} & 69.92\good{+1.63} & 35.32\bad{-1.31} & 55.21\good{+0.58} & 46.02\good{+0.22} & 36.32\bad{-0.39} & 48.56\good{+0.15} \\
 & CoT-GPT & 76.23\good{+3.07} & 55.12\good{+0.32} & 21.11\bad{+2.75} & 70.92\good{+2.63} & 37.33\good{+0.70} & 58.82\good{+4.19} & 47.46\good{+1.66} & 37.22\good{+0.51} & 50.35\good{+1.94} \\ \bottomrule[1.5pt]
\end{tabular}
}
\vspace{-15pt}
\label{tab_cot}
\end{table}

\begin{table}[t]
\centering
\caption{Performance gain of supervised fine-tuning on Qwen-VL-Chat and Internlm-XComposer-v2.}
\resizebox{\linewidth}{!}{
\setlength{\tabcolsep}{3pt}
\begin{tabular}{@{}c|lll||lll|lll|l@{}}
\toprule[1.5pt]
\multirow{2}{*}{Models} & \multirow{2}{*}{\textit{w/i}} & \multirow{2}{*}{\textit{w/o}} & \multirow{2}{*}{$\Delta$} & \multicolumn{3}{c|}{Regular} & \multicolumn{3}{c|}{Extra} & \multicolumn{1}{c}{\multirow{2}{*}{Avg.}} \\ \cmidrule(lr){5-10}
 &  &  &  & Acc+ & NQA & Avg. & Acc+ & NQA & Avg. & \multicolumn{1}{c}{} \\ \midrule
Qwen-VL-Chat~\cite{qwen-vl} & 45.71 & 28.70 & 17.01 & 29.46 & 23.57 & 28.28 & 26.56 & 21.05 & 25.46 & 26.98 \\
Qwen-VL-Chat+SFT & 60.00\good{+14.29} & 43.65\good{+14.95} & 16.35\good{-0.66} & 46.39\good{+16.93} & 25.65\good{+2.08} & 42.26\good{+13.98} & 40.18\good{+13.62} & 25.89\good{+4.84} & 37.33\good{+11.87} & 39.99\good{+13.01} \\ \midrule
Internlm-XComposer-v2~\cite{internlm2} & 73.16 & 54.80 & 18.36 & 57.89 & 40.96 & 54.52 & 41.75 & 31.58 & 39.73 & 47.78 \\ 
Internlm-XComposer-v2+SFT & 87.16\good{+14.00} & 68.20\good{+13.40} & 18.96\bad{+0.60} & 72.66\good{+14.77} & 43.81\good{+2.85} & 66.91\good{+12.39} & 62.74\good{+21.00} & 45.37\good{+13.79} & 59.28\good{+19.55} & 63.40\good{+15.65} \\ \bottomrule[1.5pt]
\end{tabular}

}
\vspace{-15pt}
\label{tab_sft}
\end{table}

\myparagraph{Results w.r.t. Datapoint annotations.}
Tab.~\ref{tab_data_anno} presents the MLLMs' performance on annotated and unannotated charts. We report only the comparison results\textit{ between the \textit{w/i} and \textit{w/o} chart versions from the same table} to ensure fair comparisons. Almost all models perform better on annotated charts. As MLLM capabilities increase, the performance gap between annotated and unannotated charts widens significantly, such as Internlm-XComposer-v2 (+18.36\%) and GPT-4V (+34.40\%). This is because OCR on annotated charts is an easier task for advanced MLLMs, while their performance on unannotated charts is limited. To further enhance MLLM capabilities, more unannotated charts are needed, highlighting the importance of our ChartBench.

\myparagraph{CoT Performance.}
Tab.~\ref{tab_cot} shows the performance of the CoT-based baseline, which generally improves performance without parameter updates. Because many models encounter difficulties in following instructions, we show the results on MiniGPT-v2, Qwen-VL-Chat, and Internlm-XComposer-v2. The fixed prompt ameliorates all tasks, especially for weaker models like MiniGPT-v2 and Qwen-VL-Chat. CoT-self is less effective because the quality and length of the self-generated CoT are uncontrollable, which hinders models from following instructions. CoT-GPT ensures CoT quality and is customized for each question type and thus performs the best. See chain of thought examples in Fig.~\ref{fig_CoT}.

\myparagraph{SFT Performance.}
Tab.~\ref{tab_sft} shows the performance of the SFT-based baseline. Each model undergoes 2 epochs of alignment and 1 epoch of SFT with a learning rate of $1e-5$. Due to the commonality of chart images, we freeze the visual encoder and update only the connector and LLM branch using LoRA~\cite{LoRA}. We balance \textit{NQA} and \textit{Acc+} instructions to avoid predictive bias. The improvement in \textit{Acc+} is particularly notable. SFT significantly boosts performance on ChartBench (Qwen-VL-Chat +13.01\%, Internlm-XComposer-v2 +15.62\%) and shows gains on ChartQA as well. Notably, Internlm-XComposer-v2, the best open-source model on ChartBench, achieves performance on par with the SOTA GPT-4o after alignment and SFT.

\section{Discussion}
\label{sec_discussion}
\vspace{-10pt}

\myparagraph{Instruction following.} 
Some models encounter difficulties in following instructions. For instance, mPLUG~\cite{Mplug} provides overly detailed responses which explains its performance on ChartQA. LLaVA-v1.6 has difficulty accurately understanding the instructions when the dictionaries extracted by OneChart~\cite{OneChart} are too lengthy. Models like Shikra~\cite{shikra} often simply reiterate the original question. Meanwhile, models like CogVLM~\cite{cogvlm} produce hallucinatory responses unrelated to the query. Therefore, instruction design greatly impacts the performance of models; the same model can yield vastly different results with different prompt templates.

\myparagraph{MLLM performance}.
MLLMs exhibit several common deficiencies in chart comprehension. 1) Since MLLMs are typically trained on \textit{images} and \textit{descriptive statements}, they prioritize giving descriptive responses to charts over numbers. This is the opposite of human graph recognition, where specific elements are identified first, followed by the final answer. 2) Some MLLMs fail to effectively follow complex instructions, which hinders their application of intricate CoT strategies. 3) Data hallucinations that occurred in VE and NQA tasks show that the data extracted by models is not yet entirely reliable, leading to errors when answers involve specific numbers. 

\myparagraph{CoT v.s. SFT.} 
Both CoT and SFT effectively improve MLLMs' capabilities, but their impacts vary. CoT shows greater improvement for weaker MLLMs (e.g., 6.48\% for Qwen-VL-Chat v.s. 1.94\% for Internlm-XComposer-v2 in Tab.~\ref{tab_cot}). The main improvement of CoT comes from unannotated charts, and Qwen-VL-Chat benefits more than Internlm-XComposer-v2. As a result, CoT provides limited improvement for MLLMs that already exhibit high performance on annotated charts. Enhancing performance on unannotated charts through CoT remains a challenging task. 
In contrast, as shown in Tab.~\ref{tab_sft}, SFT provides more significant improvements for the more powerful model Internlm-XComposer-v2 compared to Qwen-VL-Chat (Avg. gain 15.65\% v.s. 13.01\%, respectively). The improvements are comparable for both annotated and unannotated charts ($\Delta$ -0.66\% v.s. +0.60\%, respectively). This indicates that existing models are required to enhance the fundamental ability to understand unannotated charts and researchers should prioritize such data in the MLLM training.

\myparagraph{Limitations.}
1) ChartBench is required to evaluate more models, and we will continue to follow the rapidly evolving area. 2) Models are highly sensitive to prompt templates, and thus the best prompt template for each model is required to be explored further. 3) The training methods and model architectures for chart perception and reasoning are worth further exploration.

\section{Conclusion}
\vspace{-10pt}
In this paper, we introduce ChartBench to evaluate the chart comprehension abilities of MLLMs. ChartBench significantly expands chart types and requires MLLMs to infer numbers using visual cues like color or legends. We propose improved \textit{Acc+} for accurate, automated assessments, avoiding manual effort or costly LLM evaluations. We further offer two effective baselines to show how the chain of thought and supervised fine-tuning ameliorate MLLMs on charts. Our evaluation of 21 mainstream MLLMs reveals their limitations in chart interpretation and provides some insights for further directions. We aim to highlight the MLLM's ability to understand charts without data annotations. ChartBench and its code will be publicly available for research.

\clearpage
\small{
\bibliographystyle{plain}
\bibliography{main}
}

\clearpage
\addtocontents{toc}{\protect\setcounter{tocdepth}{2}}
\newpage
\onecolumn
\appendix

\begin{center}
    \Large \textbf{ChartBench: A Benchmark for Complex Visual Reasoning \\ in Charts}
    \Large \\ \textbf{Supplementary Materials}
\end{center}
\vspace{30pt}

\begin{center}
\toccolor
\begin{minipage}[t]{0.95\textwidth}
\begingroup
\renewcommand{\cftsecleader}{\cftdotfill{\cftdotsep}}
\setlength{\cftsecnumwidth}{3em}
\setlength{\cftsecindent}{1.5em}
\setlength{\cftparskip}{5pt}
\setlength{\cftbeforesecskip}{5pt}
\cftsetindents{section}{0em}{3em}
\setcounter{tocdepth}{2}
\tableofcontents
\endgroup
\end{minipage}
\defaultcolor
\end{center}
\clearpage

\section{ChartBench Statistics}
\label{sec_apdx_chartbench_stat}
\subsection{Design Principle}
ChartBench has two fundamental design principles. 1) \itemnote{Wider range of chart types}. ChartBench expands the 3 common chart types (line, bar, and pie)~\cite{ChartQA, PlotQA, OneChart} to representative 9 chart types in the real world (see Tab.~\ref{tab_chartbench_distribution} and thumbnails in Appendix~\ref{apdx_sec_thumbnails}). In the train and test sets, conventional charts account for 61.4\% and 54.8\%, respectively, while the newly added charts account for 38.6\% and 45.2\%. ChartBench further divides 9 major categories into 42 subcategories, allowing for a more detailed analysis of MLLM performance. 2) \itemnote{More intuitive visual logic}. Unlike existing benchmarks, ChartBench primarily focuses on perception and \textit{visual} logical reasoning. It emphasizes evaluating the ability to extract value from unlabeled charts rather than simple OCR or localization tasks. We assess MLLMs' core visual reasoning skills directly without converting charts into textual descriptions for further textual reasoning. Previous benchmarks mainly provided annotated charts, which led to some approaches extracting tables first and then transforming the problem into purely text-based logic. In contrast, ChartBench includes a larger proportion of unlabeled charts, accounting for 84.96\% and 76.20\% in train and test splits, respectively, in Tab.~\ref{tab_chartbench_distribution}. MLLMs must accurately extract values based on color or line shape to identify categories and their corresponding coordinate systems, rather than relying on OCR for answer candidates, which offers a more realistic assessment of MLLMs' visual reasoning abilities of charts.

\begin{table}[hb]
\caption{ChartBench training set detailed statistics. We provide statistics based on chart types and more granular image types. Each image will have two kinds of questions: \textit{Acc+} and Number QA.}
\resizebox{\linewidth}{!}{
\setlength{\tabcolsep}{10pt}
\begin{tabular}{@{}cccccccc@{}}
\toprule[1.5pt]
\multirow{2}{*}{Data Split} & \multirow{2}{*}{\begin{tabular}[c]{@{}c@{}}\#Image\\ Number\end{tabular}} & \multirow{2}{*}{Chart Type} & \multirow{2}{*}{\begin{tabular}[c]{@{}c@{}}\#Image\\ Number\end{tabular}} & \multirow{2}{*}{Image Type} & \multicolumn{3}{c}{Number} \\ \cmidrule(l){6-8} 
 &  &  &  &  & \#Image & \#\textit{Acc+} QA & \#NQA \\ \midrule
\multicolumn{1}{c|}{\multirow{23}{*}{Regular}} & \multicolumn{1}{c|}{\multirow{23}{*}{40,887}} & \multicolumn{1}{c|}{\multirow{5}{*}{Line}} & \multicolumn{1}{c|}{\multirow{5}{*}{7,830}} & \multicolumn{1}{c|}{multi-line plot} & 1,744 & 13,952 & 1,744 \\
\multicolumn{1}{c|}{} & \multicolumn{1}{c|}{} & \multicolumn{1}{c|}{} & \multicolumn{1}{c|}{} & \multicolumn{1}{c|}{multi-line plot (w/i anno)} & 1,744 & 13,952 & 1,744 \\
\multicolumn{1}{c|}{} & \multicolumn{1}{c|}{} & \multicolumn{1}{c|}{} & \multicolumn{1}{c|}{} & \multicolumn{1}{c|}{single line plot} & 1,744 & 13,952 & 1,744 \\
\multicolumn{1}{c|}{} & \multicolumn{1}{c|}{} & \multicolumn{1}{c|}{} & \multicolumn{1}{c|}{} & \multicolumn{1}{c|}{single line plot (w/i anno)} & 1,744 & 13,952 & 1,744 \\
\multicolumn{1}{c|}{} & \multicolumn{1}{c|}{} & \multicolumn{1}{c|}{} & \multicolumn{1}{c|}{} & \multicolumn{1}{c|}{line with error plot} & 854 & 6,832 & 854 \\ \cmidrule(l){3-8} 
\multicolumn{1}{c|}{} & \multicolumn{1}{c|}{} & \multicolumn{1}{c|}{\multirow{13}{*}{Bar}} & \multicolumn{1}{c|}{\multirow{13}{*}{24,580}} & \multicolumn{1}{c|}{horizontal single bar plot} & 1,891 & 15,128 & 1,891 \\
\multicolumn{1}{c|}{} & \multicolumn{1}{c|}{} & \multicolumn{1}{c|}{} & \multicolumn{1}{c|}{} & \multicolumn{1}{c|}{horizontal single bar plot (w/i anno)} & 1,891 & 15,128 & 1,891 \\
\multicolumn{1}{c|}{} & \multicolumn{1}{c|}{} & \multicolumn{1}{c|}{} & \multicolumn{1}{c|}{} & \multicolumn{1}{c|}{horizontal multi-bar plot} & 1,891 & 15,128 & 1,891 \\
\multicolumn{1}{c|}{} & \multicolumn{1}{c|}{} & \multicolumn{1}{c|}{} & \multicolumn{1}{c|}{} & \multicolumn{1}{c|}{horizontal stacked bar plot} & 1,891 & 15,128 & 1,891 \\
\multicolumn{1}{c|}{} & \multicolumn{1}{c|}{} & \multicolumn{1}{c|}{} & \multicolumn{1}{c|}{} & \multicolumn{1}{c|}{horizontal stacked bar in percentage plot} & 1,890 & 15,120 & 1,890 \\
\multicolumn{1}{c|}{} & \multicolumn{1}{c|}{} & \multicolumn{1}{c|}{} & \multicolumn{1}{c|}{} & \multicolumn{1}{c|}{vertical single bar plot} & 1,891 & 15,128 & 1,891 \\
\multicolumn{1}{c|}{} & \multicolumn{1}{c|}{} & \multicolumn{1}{c|}{} & \multicolumn{1}{c|}{} & \multicolumn{1}{c|}{vertical single bar plot (w/i anno)} & 1,891 & 15,128 & 1,891 \\
\multicolumn{1}{c|}{} & \multicolumn{1}{c|}{} & \multicolumn{1}{c|}{} & \multicolumn{1}{c|}{} & \multicolumn{1}{c|}{vertical multi-bar plot} & 1,891 & 15,128 & 1,891 \\
\multicolumn{1}{c|}{} & \multicolumn{1}{c|}{} & \multicolumn{1}{c|}{} & \multicolumn{1}{c|}{} & \multicolumn{1}{c|}{vertical stacked bar plot} & 1,891 & 15,128 & 1,891 \\
\multicolumn{1}{c|}{} & \multicolumn{1}{c|}{} & \multicolumn{1}{c|}{} & \multicolumn{1}{c|}{} & \multicolumn{1}{c|}{vertical stacked bar in percentage plot} & 1,890 & 15,120 & 1,890 \\
\multicolumn{1}{c|}{} & \multicolumn{1}{c|}{} & \multicolumn{1}{c|}{} & \multicolumn{1}{c|}{} & \multicolumn{1}{c|}{3D multi-bar plot} & 1,891 & 15,128 & 1,891 \\
\multicolumn{1}{c|}{} & \multicolumn{1}{c|}{} & \multicolumn{1}{c|}{} & \multicolumn{1}{c|}{} & \multicolumn{1}{c|}{3D stacked bar plot} & 1,891 & 15,128 & 1,891 \\
\multicolumn{1}{c|}{} & \multicolumn{1}{c|}{} & \multicolumn{1}{c|}{} & \multicolumn{1}{c|}{} & \multicolumn{1}{c|}{3D stacked bar in percentage plot} & 1,890 & 15,120 & 1,890 \\ \cmidrule(l){3-8} 
\multicolumn{1}{c|}{} & \multicolumn{1}{c|}{} & \multicolumn{1}{c|}{\multirow{5}{*}{Pie}} & \multicolumn{1}{c|}{\multirow{5}{*}{8,477}} & \multicolumn{1}{c|}{ring plot} & 1,989 & 15,912 & 1,989 \\
\multicolumn{1}{c|}{} & \multicolumn{1}{c|}{} & \multicolumn{1}{c|}{} & \multicolumn{1}{c|}{} & \multicolumn{1}{c|}{ring plot (w/i anno)} & 1,989 & 15,912 & 1,989 \\
\multicolumn{1}{c|}{} & \multicolumn{1}{c|}{} & \multicolumn{1}{c|}{} & \multicolumn{1}{c|}{} & \multicolumn{1}{c|}{inter sun plot} & 521 & 4,168 & 521 \\
\multicolumn{1}{c|}{} & \multicolumn{1}{c|}{} & \multicolumn{1}{c|}{} & \multicolumn{1}{c|}{} & \multicolumn{1}{c|}{sector plot} & 1,989 & 15,912 & 1,989 \\
\multicolumn{1}{c|}{} & \multicolumn{1}{c|}{} & \multicolumn{1}{c|}{} & \multicolumn{1}{c|}{} & \multicolumn{1}{c|}{pie plot} & 1,989 & 15,912 & 1,989 \\ \midrule
\multicolumn{1}{c|}{\multirow{19}{*}{Extra}} & \multicolumn{1}{c|}{\multirow{19}{*}{25,737}} & \multicolumn{1}{c|}{\multirow{3}{*}{Area}} & \multicolumn{1}{c|}{\multirow{3}{*}{5,613}} & \multicolumn{1}{c|}{area plot} & 1,871 & 14,968 & 1,871 \\
\multicolumn{1}{c|}{} & \multicolumn{1}{c|}{} & \multicolumn{1}{c|}{} & \multicolumn{1}{c|}{} & \multicolumn{1}{c|}{area in percentage plot} & 1,871 & 14,968 & 1,871 \\
\multicolumn{1}{c|}{} & \multicolumn{1}{c|}{} & \multicolumn{1}{c|}{} & \multicolumn{1}{c|}{} & \multicolumn{1}{c|}{stacked area plot} & 1,871 & 14,968 & 1,871 \\ \cmidrule(l){3-8} 
\multicolumn{1}{c|}{} & \multicolumn{1}{c|}{} & \multicolumn{1}{c|}{\multirow{3}{*}{Box}} & \multicolumn{1}{c|}{\multirow{3}{*}{4,068}} & \multicolumn{1}{c|}{stock plot} & 1,356 & 10,848 & 1,356 \\
\multicolumn{1}{c|}{} & \multicolumn{1}{c|}{} & \multicolumn{1}{c|}{} & \multicolumn{1}{c|}{} & \multicolumn{1}{c|}{vertical box plot} & 1,356 & 10,848 & 1,356 \\
\multicolumn{1}{c|}{} & \multicolumn{1}{c|}{} & \multicolumn{1}{c|}{} & \multicolumn{1}{c|}{} & \multicolumn{1}{c|}{horizontal box plot} & 1,356 & 10,848 & 1,356 \\ \cmidrule(l){3-8} 
\multicolumn{1}{c|}{} & \multicolumn{1}{c|}{} & \multicolumn{1}{c|}{\multirow{4}{*}{Radar}} & \multicolumn{1}{c|}{\multirow{4}{*}{3,056}} & \multicolumn{1}{c|}{single radar plot} & 764 & 6,112 & 764 \\
\multicolumn{1}{c|}{} & \multicolumn{1}{c|}{} & \multicolumn{1}{c|}{} & \multicolumn{1}{c|}{} & \multicolumn{1}{c|}{single radar plot (w/i anno)} & 764 & 6,112 & 764 \\
\multicolumn{1}{c|}{} & \multicolumn{1}{c|}{} & \multicolumn{1}{c|}{} & \multicolumn{1}{c|}{} & \multicolumn{1}{c|}{multi-radar plot} & 764 & 6,112 & 764 \\
\multicolumn{1}{c|}{} & \multicolumn{1}{c|}{} & \multicolumn{1}{c|}{} & \multicolumn{1}{c|}{} & \multicolumn{1}{c|}{multi-radar with fill plot} & 764 & 6,112 & 764 \\ \cmidrule(l){3-8} 
\multicolumn{1}{c|}{} & \multicolumn{1}{c|}{} & \multicolumn{1}{c|}{\multirow{3}{*}{Scatter}} & \multicolumn{1}{c|}{\multirow{3}{*}{2,046}} & \multicolumn{1}{c|}{2D scatter plot} & 784 & 6,272 & 784 \\
\multicolumn{1}{c|}{} & \multicolumn{1}{c|}{} & \multicolumn{1}{c|}{} & \multicolumn{1}{c|}{} & \multicolumn{1}{c|}{2D scatter smooth plot} & 784 & 6,272 & 784 \\
\multicolumn{1}{c|}{} & \multicolumn{1}{c|}{} & \multicolumn{1}{c|}{} & \multicolumn{1}{c|}{} & \multicolumn{1}{c|}{3D scatter} & 478 & 3,824 & 478 \\ \cmidrule(l){3-8} 
\multicolumn{1}{c|}{} & \multicolumn{1}{c|}{} & \multicolumn{1}{c|}{\multirow{2}{*}{Node}} & \multicolumn{1}{c|}{\multirow{2}{*}{3,978}} & \multicolumn{1}{c|}{undirected node plot} & 1,989 & 15,912 & 1,989 \\
\multicolumn{1}{c|}{} & \multicolumn{1}{c|}{} & \multicolumn{1}{c|}{} & \multicolumn{1}{c|}{} & \multicolumn{1}{c|}{directed node plot} & 1,989 & 15,912 & 1,989 \\ \cmidrule(l){3-8} 
\multicolumn{1}{c|}{} & \multicolumn{1}{c|}{} & \multicolumn{1}{c|}{\multirow{4}{*}{Combination}} & \multicolumn{1}{c|}{\multirow{4}{*}{6,976}} & \multicolumn{1}{c|}{line \& line plot (dual coordinate)} & 1,744 & 13,952 & 1,744 \\
\multicolumn{1}{c|}{} & \multicolumn{1}{c|}{} & \multicolumn{1}{c|}{} & \multicolumn{1}{c|}{} & \multicolumn{1}{c|}{bar \& line plot (dual coordinate)} & 1,744 & 13,952 & 1,744 \\
\multicolumn{1}{c|}{} & \multicolumn{1}{c|}{} & \multicolumn{1}{c|}{} & \multicolumn{1}{c|}{} & \multicolumn{1}{c|}{pie \& bar combinated plot} & 1,744 & 13,952 & 1,744 \\
\multicolumn{1}{c|}{} & \multicolumn{1}{c|}{} & \multicolumn{1}{c|}{} & \multicolumn{1}{c|}{} & \multicolumn{1}{c|}{pie \& pie combinated plot} & 1,744 & 13,952 & 1,744 \\ \midrule
\multicolumn{1}{c|}{Total} & \multicolumn{1}{c|}{66,624} & \multicolumn{1}{c|}{Total} & \multicolumn{1}{c|}{66,624} & \multicolumn{1}{c|}{Total} & 66,624 & 532,992 & 66,624 \\ \bottomrule[1.5pt]
\end{tabular}
}
\label{tab_training_set_detail}
\end{table}

\begin{table}[ht]
\caption{ChartBench test set detailed statistics. We provide statistics based on chart types and more granular image types. Each image will have two kinds of questions: \textit{Acc+} and Number QA.}
\resizebox{\linewidth}{!}{
\begin{tabular}{@{}cccccccc@{}}
\toprule[1.5pt]
\multirow{2}{*}{Data Split} & \multirow{2}{*}{\begin{tabular}[c]{@{}c@{}}\#Image\\ Number\end{tabular}} & \multirow{2}{*}{Chart Type} & \multirow{2}{*}{\begin{tabular}[c]{@{}c@{}}\#Image\\ Number\end{tabular}} & \multirow{2}{*}{Image Type} & \multicolumn{3}{c}{Number} \\ \cmidrule(l){6-8} 
 &  &  &  &  & \#Image & \#\textit{Acc+} QA & \#NQA \\ \midrule
\multicolumn{1}{c|}{\multirow{23}{*}{Regular}} & \multicolumn{1}{c|}{\multirow{23}{*}{1,150}} & \multicolumn{1}{c|}{\multirow{5}{*}{Line}} & \multicolumn{1}{c|}{\multirow{5}{*}{250}} & \multicolumn{1}{c|}{multi-line plot} & 50 & 400 & 50 \\
\multicolumn{1}{c|}{} & \multicolumn{1}{c|}{} & \multicolumn{1}{c|}{} & \multicolumn{1}{c|}{} & \multicolumn{1}{c|}{multi-line plot (w/i anno)} & 50 & 400 & 50 \\
\multicolumn{1}{c|}{} & \multicolumn{1}{c|}{} & \multicolumn{1}{c|}{} & \multicolumn{1}{c|}{} & \multicolumn{1}{c|}{single line plot} & 50 & 400 & 50 \\
\multicolumn{1}{c|}{} & \multicolumn{1}{c|}{} & \multicolumn{1}{c|}{} & \multicolumn{1}{c|}{} & \multicolumn{1}{c|}{single line plot (w/i anno)} & 50 & 400 & 50 \\
\multicolumn{1}{c|}{} & \multicolumn{1}{c|}{} & \multicolumn{1}{c|}{} & \multicolumn{1}{c|}{} & \multicolumn{1}{c|}{line with error plot} & 50 & 400 & 50 \\ \cmidrule(l){3-8} 
\multicolumn{1}{c|}{} & \multicolumn{1}{c|}{} & \multicolumn{1}{c|}{\multirow{13}{*}{Bar}} & \multicolumn{1}{c|}{\multirow{13}{*}{650}} & \multicolumn{1}{c|}{horizontal single bar plot} & 50 & 400 & 50 \\
\multicolumn{1}{c|}{} & \multicolumn{1}{c|}{} & \multicolumn{1}{c|}{} & \multicolumn{1}{c|}{} & \multicolumn{1}{c|}{horizontal single bar plot (w/i anno)} & 50 & 400 & 50 \\
\multicolumn{1}{c|}{} & \multicolumn{1}{c|}{} & \multicolumn{1}{c|}{} & \multicolumn{1}{c|}{} & \multicolumn{1}{c|}{horizontal multi-bar plot} & 50 & 400 & 50 \\
\multicolumn{1}{c|}{} & \multicolumn{1}{c|}{} & \multicolumn{1}{c|}{} & \multicolumn{1}{c|}{} & \multicolumn{1}{c|}{horizontal stacked bar plot} & 50 & 400 & 50 \\
\multicolumn{1}{c|}{} & \multicolumn{1}{c|}{} & \multicolumn{1}{c|}{} & \multicolumn{1}{c|}{} & \multicolumn{1}{c|}{horizontal stacked bar in percentage plot} & 50 & 400 & 50 \\
\multicolumn{1}{c|}{} & \multicolumn{1}{c|}{} & \multicolumn{1}{c|}{} & \multicolumn{1}{c|}{} & \multicolumn{1}{c|}{vertical single bar plot} & 50 & 400 & 50 \\
\multicolumn{1}{c|}{} & \multicolumn{1}{c|}{} & \multicolumn{1}{c|}{} & \multicolumn{1}{c|}{} & \multicolumn{1}{c|}{vertical single bar plot (w/i anno)} & 50 & 400 & 50 \\
\multicolumn{1}{c|}{} & \multicolumn{1}{c|}{} & \multicolumn{1}{c|}{} & \multicolumn{1}{c|}{} & \multicolumn{1}{c|}{vertical multi-bar plot} & 50 & 400 & 50 \\
\multicolumn{1}{c|}{} & \multicolumn{1}{c|}{} & \multicolumn{1}{c|}{} & \multicolumn{1}{c|}{} & \multicolumn{1}{c|}{vertical stacked bar plot} & 50 & 400 & 50 \\
\multicolumn{1}{c|}{} & \multicolumn{1}{c|}{} & \multicolumn{1}{c|}{} & \multicolumn{1}{c|}{} & \multicolumn{1}{c|}{vertical stacked bar in percentage plot} & 50 & 400 & 50 \\
\multicolumn{1}{c|}{} & \multicolumn{1}{c|}{} & \multicolumn{1}{c|}{} & \multicolumn{1}{c|}{} & \multicolumn{1}{c|}{3D multi-bar plot} & 50 & 400 & 50 \\
\multicolumn{1}{c|}{} & \multicolumn{1}{c|}{} & \multicolumn{1}{c|}{} & \multicolumn{1}{c|}{} & \multicolumn{1}{c|}{3D stacked bar plot} & 50 & 400 & 50 \\
\multicolumn{1}{c|}{} & \multicolumn{1}{c|}{} & \multicolumn{1}{c|}{} & \multicolumn{1}{c|}{} & \multicolumn{1}{c|}{3D stacked bar in percentage plot} & 50 & 400 & 50 \\ \cmidrule(l){3-8} 
\multicolumn{1}{c|}{} & \multicolumn{1}{c|}{} & \multicolumn{1}{c|}{\multirow{5}{*}{Pie}} & \multicolumn{1}{c|}{\multirow{5}{*}{250}} & \multicolumn{1}{c|}{ring plot} & 50 & 400 & 50 \\
\multicolumn{1}{c|}{} & \multicolumn{1}{c|}{} & \multicolumn{1}{c|}{} & \multicolumn{1}{c|}{} & \multicolumn{1}{c|}{ring plot (w/i anno)} & 50 & 400 & 50 \\
\multicolumn{1}{c|}{} & \multicolumn{1}{c|}{} & \multicolumn{1}{c|}{} & \multicolumn{1}{c|}{} & \multicolumn{1}{c|}{inter sun plot} & 50 & 400 & 50 \\
\multicolumn{1}{c|}{} & \multicolumn{1}{c|}{} & \multicolumn{1}{c|}{} & \multicolumn{1}{c|}{} & \multicolumn{1}{c|}{sector plot} & 50 & 400 & 50 \\
\multicolumn{1}{c|}{} & \multicolumn{1}{c|}{} & \multicolumn{1}{c|}{} & \multicolumn{1}{c|}{} & \multicolumn{1}{c|}{pie plot} & 50 & 400 & 50 \\ \midrule
\multicolumn{1}{c|}{\multirow{19}{*}{Extra}} & \multicolumn{1}{c|}{\multirow{19}{*}{950}} & \multicolumn{1}{c|}{\multirow{3}{*}{Area}} & \multicolumn{1}{c|}{\multirow{3}{*}{150}} & \multicolumn{1}{c|}{area plot} & 50 & 400 & 50 \\
\multicolumn{1}{c|}{} & \multicolumn{1}{c|}{} & \multicolumn{1}{c|}{} & \multicolumn{1}{c|}{} & \multicolumn{1}{c|}{area in percentage plot} & 50 & 400 & 50 \\
\multicolumn{1}{c|}{} & \multicolumn{1}{c|}{} & \multicolumn{1}{c|}{} & \multicolumn{1}{c|}{} & \multicolumn{1}{c|}{stacked area plot} & 50 & 400 & 50 \\ \cmidrule(l){3-8} 
\multicolumn{1}{c|}{} & \multicolumn{1}{c|}{} & \multicolumn{1}{c|}{\multirow{3}{*}{Box}} & \multicolumn{1}{c|}{\multirow{3}{*}{150}} & \multicolumn{1}{c|}{stock plot} & 50 & 400 & 50 \\
\multicolumn{1}{c|}{} & \multicolumn{1}{c|}{} & \multicolumn{1}{c|}{} & \multicolumn{1}{c|}{} & \multicolumn{1}{c|}{vertical box plot} & 50 & 400 & 50 \\
\multicolumn{1}{c|}{} & \multicolumn{1}{c|}{} & \multicolumn{1}{c|}{} & \multicolumn{1}{c|}{} & \multicolumn{1}{c|}{horizontal box plot} & 50 & 400 & 50 \\ \cmidrule(l){3-8} 
\multicolumn{1}{c|}{} & \multicolumn{1}{c|}{} & \multicolumn{1}{c|}{\multirow{4}{*}{Radar}} & \multicolumn{1}{c|}{\multirow{4}{*}{200}} & \multicolumn{1}{c|}{single radar plot} & 50 & 400 & 50 \\
\multicolumn{1}{c|}{} & \multicolumn{1}{c|}{} & \multicolumn{1}{c|}{} & \multicolumn{1}{c|}{} & \multicolumn{1}{c|}{single radar plot (w/i anno)} & 50 & 400 & 50 \\
\multicolumn{1}{c|}{} & \multicolumn{1}{c|}{} & \multicolumn{1}{c|}{} & \multicolumn{1}{c|}{} & \multicolumn{1}{c|}{multi-radar plot} & 50 & 400 & 50 \\
\multicolumn{1}{c|}{} & \multicolumn{1}{c|}{} & \multicolumn{1}{c|}{} & \multicolumn{1}{c|}{} & \multicolumn{1}{c|}{multi-radar with fill plot} & 50 & 400 & 50 \\ \cmidrule(l){3-8} 
\multicolumn{1}{c|}{} & \multicolumn{1}{c|}{} & \multicolumn{1}{c|}{\multirow{3}{*}{Scatter}} & \multicolumn{1}{c|}{\multirow{3}{*}{150}} & \multicolumn{1}{c|}{2D scatter plot} & 50 & 400 & 50 \\
\multicolumn{1}{c|}{} & \multicolumn{1}{c|}{} & \multicolumn{1}{c|}{} & \multicolumn{1}{c|}{} & \multicolumn{1}{c|}{2D scatter smooth plot} & 50 & 400 & 50 \\
\multicolumn{1}{c|}{} & \multicolumn{1}{c|}{} & \multicolumn{1}{c|}{} & \multicolumn{1}{c|}{} & \multicolumn{1}{c|}{3D scatter} & 50 & 400 & 50 \\ \cmidrule(l){3-8} 
\multicolumn{1}{c|}{} & \multicolumn{1}{c|}{} & \multicolumn{1}{c|}{\multirow{2}{*}{Node}} & \multicolumn{1}{c|}{\multirow{2}{*}{100}} & \multicolumn{1}{c|}{undirected node plot} & 50 & 400 & 50 \\
\multicolumn{1}{c|}{} & \multicolumn{1}{c|}{} & \multicolumn{1}{c|}{} & \multicolumn{1}{c|}{} & \multicolumn{1}{c|}{directed node plot} & 50 & 400 & 50 \\ \cmidrule(l){3-8} 
\multicolumn{1}{c|}{} & \multicolumn{1}{c|}{} & \multicolumn{1}{c|}{\multirow{4}{*}{Combination}} & \multicolumn{1}{c|}{\multirow{4}{*}{200}} & \multicolumn{1}{c|}{line \& line plot (dual coordinate)} & 50 & 400 & 50 \\
\multicolumn{1}{c|}{} & \multicolumn{1}{c|}{} & \multicolumn{1}{c|}{} & \multicolumn{1}{c|}{} & \multicolumn{1}{c|}{bar \& line plot (dual coordinate)} & 50 & 400 & 50 \\
\multicolumn{1}{c|}{} & \multicolumn{1}{c|}{} & \multicolumn{1}{c|}{} & \multicolumn{1}{c|}{} & \multicolumn{1}{c|}{pie \& bar combinated plot} & 50 & 400 & 50 \\
\multicolumn{1}{c|}{} & \multicolumn{1}{c|}{} & \multicolumn{1}{c|}{} & \multicolumn{1}{c|}{} & \multicolumn{1}{c|}{pie \& pie combinated plot} & 50 & 400 & 50 \\ \midrule
\multicolumn{1}{c|}{Total} & \multicolumn{1}{c|}{2,100} & \multicolumn{1}{c|}{Total} & \multicolumn{1}{c|}{2,100} & \multicolumn{1}{c|}{Total} & 2,100 & 16,800 & 2,100 \\ \bottomrule[1.5pt]
\end{tabular}
}
\vspace{-10pt}
\label{tab_test_set_detail}
\end{table}

\subsection{Chart Taxonomy}
ChartBench primarily focuses on the following evaluation aspects: 
1) \itemnote{Bar charts} are the most common and have been the focus of ChartQA and ChartLLaMA. ChartBench includes basic variations such as horizontal and vertical bar orientations, data complexity (single and multiple groups of data), and different representations (regular, percentage, stacked, and 3D bar charts). 
2) \itemnote{Line charts} are commonly used chart types to reflect data trends. ChartBench includes error line charts as well as regular single or multiple groups, with or without annotations line charts. 
3) \itemnote{Pie charts} primarily show the data proportional distribution. ChartBench includes single, nested, doughnut pie charts, and irregular sector charts. 
4) \itemnote{Radar charts} have a straightforward distribution structure and are used to represent multiple attributes of an entity. ChartBench incorporates diverse data complexities (single or multiple groups) and representations (with or without fillings). 
5) \itemnote{Box charts} primarily depict the statistical distribution of a substantial volume of data points. ChartBench collects horizontal and vertical box plots, as well as authentic candlestick charts depicting real stock prices. 
6) \itemnote{Scatter charts} mainly depict the distribution of discrete data. ChartBench includes simple single or multi-group scatter plots, 3D bubble plots, and scatter plots with interpolated smoothing lines. 
7) \itemnote{Area charts} employ color fillings to visually convey the magnitude and distribution of data. ChartBench encompasses single or multiple groups area plots, stacked and percentage stacked area charts. 
8) \itemnote{Node charts} primarily illustrate the logical relationships between nodes. ChartBench includes directed and undirected graphs, as well as simple and complex node-link diagrams. 
9) \itemnote{Combination charts} combine the above-mentioned chart types. ChartBench includes dual coordinate system charts (e.g. line and bar charts), multi-level pie charts, and combinations between bar and pie charts.

\subsection{Data Splitting}
\label{sec_apdx_data_split_detail}
Tab.~\ref{tab_training_set_detail} and Tab.~\ref{tab_test_set_detail} show the hierarchical relationship and quantity of each type of chart in detail. Note that the distribution of train and test set is slightly different because we guarantee that each subclass in the test split has 50 data points. For each chart, we generate questions on 5 different tasks to evaluate MLLMs' basic performance on perception and cognition. Notice that some categories have two variants, i.e., \textit{w/i} and \textit{w/o} annotations. Although the dataset mainly consists of unannotated charts, we only report the results of comparisons between the \textit{w/i} and \textit{w/o} chart versions derived from the same table in our experiments to ensure fair comparisons.

\section{Participating MLLMs}

\subsection{Architecture}
\label{apdx_sec_mllm_arch}
\begin{table*}[ht]
\centering
\caption{Open-sourced model architecture. Note that we classify connector components such as QFormer~\citep{BLIP2} as the visual branch for brevity. Mem.: the maximum GPU memory usage during inference. Time: the average inference time per QA. Due to the multiple visual encoders in SPHINX~\cite{sphinx}, which extract more robust visual representations, \textit{mixed} refers to QFormer~\citep{BLIP2}, OpenCLIP ViT-L/14~\citep{openclip}, OpenCLIP ConvNeXt-XXL~\citep{openclip,cherti2023reproducible}, DINOv2-ViT-g/14~\citep{dinov2} and MLP.}
% \vspace{-10pt}
\setlength{\tabcolsep}{5pt}
\resizebox{\linewidth}{!}{
\begin{tabular}{@{}c|c|cc|cc|cc@{}}
\toprule[1.5pt]
Models & Total Size & LLM Branch & LLM Size & Visual Branch & Visual Size & Peak Memory (G) & Inference Time (s) \\ \midrule
BLIP2~\cite{BLIP2} & 12.1B & FlanT5-XXL & 11B & EVA-CLIP-g/14 & 1B & 39.60 & 0.176 \\
CogVLM-Chat~\cite{cogvlm} & 17B & Vicuna-7B & 7B & EVA-02-CLIP-E/14 & 4.4B & 39.60 & 1.455 \\
InstructBLIP~\cite{InstructBLIP} & 8.2B & Vicuna-7B & 7B & EVA-CLIP-g/14 & 1B & 36.50 & 0.895 \\
Internlm-XComposer~\cite{internlm-xcomposer} & 8.2B & InternLM-Chat-7B & 7B & EVA-CLIP-g/14 & 1B & 22.20 & 0.707 \\
LLaVA-v1.5~\cite{llava} & 13.4B & Vicuna-13B & 13B & CLIP ViT-L/14@336px & 304M & 16.50 & 0.534 \\
MiniGPT-v2~\cite{Minigptv2} & 8.1B & LLaMA2-Chat-7B & 7B & EVA-ViT-g/14 & 1B & 17.20 & 0.236 \\
mPLUG-Owl-bloomz~\cite{Mplug} & 7.4B & Bloomz-7B & 7B & CLIP ViT-L/14 & 304M & 16.00 & 0.284 \\
Qwen-VL-Chat~\cite{qwen-vl} & 9.6B & Qwen-7B & 7.7B & OpenCLIP ViT-G/14 & 1.9B & 21.00 & 0.269 \\
Shikra~\cite{shikra} & 7.4B & Vicuna-7B & 7B & CLIP ViT-L/14 & 304M & 15.60 & 0.561 \\
SPHINX~\cite{sphinx} & 15.7B & LLaMA-13B & 13B & Mixed & 2.7B & 29.6 * 2 & 0.581 \\
VisualGLM~\cite{glm} & 7.8B & ChatGLM-6B & 6.2B & EVA-CLIP-g/14 & 1B & 16.00 & 0.201 \\
ChartLlama~\cite{Chartllama} & 13.4B & Vicuna-13B & 13B & CLIP ViT-L/14@336px & 304M & 29.00 & 0.593 \\
DocOwl-v1.5~\cite{docowl} & 8.1B & Bloomz-7B & 7B & CLIP ViT-L/14 & 304M & 37.5 & 0.483 \\
Mini-Gemini~\cite{minigemini} & 14B & Vicuna-13B & 13B & ConvNext-L + CLIP ViT-L/14 & 502M & 32.45 & 3.951 \\
Internlm-XComposer-v2~\cite{internlm2} & 8B & InternLM2-7B & 7B & CLIP ViT-L/14 & 304M & 23.72 & 0.945 \\
OneChart~\cite{OneChart} & 13.4B & Vicuna-13B & 13B & SAM-base ViT & 304M & 37.62 & 2.201 \\
ChartVLM~\cite{ChartX} & 7.4B & Vicuna-7B & 7B & Pix2Struct-base & 304M & 17.83 & 2.831 \\
CogAgent~\cite{cogagent} & 7.4B & Vicuna-7B & 7B & EVA2-CLIP-L & 304M & 18.82 & 2.548 \\ \bottomrule[1.5pt]
\end{tabular}

}
\label{tab_mllm_arch}
\vspace{-10pt}
\end{table*}

We evaluate 18 main-stream open-sourced and 3 closed-sourced MLLMs on ChartBench. The open-source models include \itemnote{BLIP2}~\cite{BLIP2}, \itemnote{CogVLM-Chat}~\cite{cogvlm}, \itemnote{InstructBLIP}~\cite{InstructBLIP}, \itemnote{InternLM-XComposer}~\citep{internlm-xcomposer}, \itemnote{LLaVA-v1.5}~\cite{Improvedllava}, \itemnote{MiniGPT-v2}~\cite{Minigptv2}, \itemnote{mPLUG-Owl-bloomz}~\cite{Mplug}, \itemnote{Qwen-VL-Chat}~\cite{qwen-vl}, \itemnote{Shikra}~\cite{shikra}, \itemnote{SPHINX}~\cite{sphinx}, \itemnote{VisualGLM}~\citep{glm,cogview}, \itemnote{ChartLlama}~\cite{Chartllama}, \itemnote{DocOwl-v1.5}~\cite{docowl}, \itemnote{Mini-Gemini}~\cite{minigemini}, \itemnote{Internlm-XComposer-v2}~\cite{internlm2}, \itemnote{OneChart}~\cite{OneChart}, \itemnote{ChartVLM}~\cite{ChartX},  \itemnote{CogAgent}~\cite{cogagent}, while the closed-source models contain \itemnote{Baidu ERNIE}~\cite{wenxin}, \itemnote{GPT-4V / GPT-4O}~\cite{GPT4}. Some close-sourced models do not provide efficient APIs, so we randomly sample a subset for evaluations. Tab.~\ref{tab_mllm_arch} summarizes the visual and LLM branch architecture, along with memory costs and inference latency on NVIDIA A100-40G GPUs.

\itemnote{BLIP2}~\citep{BLIP2} proposes a lightweight Query Transformer to leverage off-the-shelf frozen image encoders and LLMs, which is pre-trained via a two-stage strategy. We test \textit{BLIP-2 ViT-g FlanT5-xxl}~\citep{eva,FlanT5}.

\itemnote{CogVLM-Chat}~\citep{cogvlm} bridges the gap between the frozen vision encoder and LLM by integrating a visual expert module in the transformer block. We test the version \noindent\textit{CogVLM-Chat-17B}, which leverages Vicuna-7B finetuned from LLaMA2~\citep{llama-2} and EVA-02-CLIP-E/14~\citep{eva-clip} as unimodal encoders.

\itemnote{InstructBLIP}~\citep{InstructBLIP} extends the framework of instruction tuning to the BLIP2, and demonstrates its appealing ability of generalization. We carry out evaluations on version \textit{IntructBLIP-7B}, which uses EVA-CLIP-g/14 as vision encoder and Vicuna-7B as text encoder.

\itemnote{InternLM-XComposer}~\citep{internlm-xcomposer} is an instruction-tuned MLLM based on InternLM~\citep{internlm}. It is empowered by tuning on extensive multimodal multilingual concepts with carefully crafted strategies. We test the released version of \textit{InternLM-XComposer-7B} with InternLM-Chat-7B~\citep{internlm} and EVA-CLIP-g/14.

\itemnote{LLaVA-v1.5}~\citep{Improvedllava} is a variant of LLaVA~\citep{llava} with exquisite modifications, such as curated datasets, larger input resolution, modality connector and prompt engineering. We test the version of \textit{LLaVA-v1.5-13B} with Vicuna-13B and CLIP ViT-L/14@336px~\citep{clip}.

\itemnote{MiniGPT-v2}~\citep{Minigptv2} proposes a three-stage training paradigm and uses unique identifiers for different tasks, building a unified interface for multiple vision-language tasks. We test \textit{MiniGPT-v2-7B} version, leveraging LLaMA2-Chat-7B and EVA-ViT-g/14 as unimodal encoders.

\itemnote{mPLUG-Owl-bloomz}~\citep{Mplug} equips LLM with visual abilities by modularized learning of LLM, visual knowledge module, and visual abstractor module. We conduct evaluations on \textit{mPLUG-Owl-bloomz-7B} version with Bloomz-7B~\citep{bloomz} and CLIP ViT-L/14.

\itemnote{Qwen-VL-Chat}~\citep{qwen-vl} is trained with alignment techniques, which support more flexible interaction, such as multiple image inputs, multi-round question answering and creative capability. We test the version of \textit{Qwen-VL-Chat-7B} with Qwen-7B~\citep{qwen} and OpenCLIP ViT-G/14~\citep{openclip,cherti2023reproducible}.

\itemnote{Shikra}~\citep{shikra} proposes to tackle spatial coordinate inputs and outputs in natural language without extra plug-in models or vocabularies. We test the version \textit{Shikra-7B} which uses Vicuna-7B and CLIP ViT-L/14.

\itemnote{SPHINX}~\citep{sphinx} showcases the superior capability of multi-modal understanding with a joint mixing of model weights, tuning tasks, visual embeddings, and sub-images of different scales. We conduct the test on version \textit{SPHINX-13B}, whose visual branch (note as mixed in Tab.~\ref{tab_mllm_arch}) is a mixture of QFormer, OpenCLIP ViT-L/14, OpenCLIP ConvNeXt-XXL and DINOv2-ViT-g/14~\citep{dinov2} and LLM branch is LLaMA-13B~\citep{llama}.

\itemnote{VisualGLM}~\citep{glm,cogview} is an open-source, multi-modal dialogue language model. We test \textit{VisualGLM-6B} based on ChatGLM-6B~\citep{glm} and EVA-CLIP-g/14.

\itemnote{ChartLlama}~\citep{Chartllama} proposes to endow \textit{LLaVA-v1.5} with the capability of chart understanding and generation. We evaluate \textit{ChartLlama-13B}, which uses Vicuna-13B and CLIP ViT-L/14@336px.

\itemnote{DocOwl-v1.5}~\citep{docowl} propose to merge visual tokens horizontally to handle high-resolution images and align all data with markdown. We evaluate the DocOwl-Omni version in our experiments, which is good at document/webpage parsing and VQA with concise answers.

\itemnote{Mini-Gemini}~\citep{minigemini} adopt two visual encoders to handle low and high-resolution images. This approach is applicable to a variety of LLMs, and we select the Mini-Gemini-Vicuna-13B for evaluation.

\itemnote{Internlm-XComposer-v2}~\cite{internlm2} introduces a Partial LoRA approach, applying additional LoRA parameters only to image tokens. This preserves the integrity of the model's pre-trained language knowledge while enabling precise vision understanding and literary-level text composition. Compared to the first version, the performance of Internlm-XComposer-v2 has been significantly improved.

\itemnote{OneChart}~\cite{OneChart} introduces an auxiliary token placed at the beginning of the token sequence, along with an additional decoder. This decoder will provide a Python dictionary about chart metadata. OneChart needs to be used in conjunction with other MLLMS, so we choose LLaVA-v1.6, which is the best model in the paper.

\itemnote{ChartVLM}~\cite{ChartX} extracts metadata of chart based on Pix2Struct~\cite{Pix2Str}. It employs an instruction adapter to dynamically select tasks based on user instructions and provides two decoders for the base and complex queries. ChartVLM has two variants and we select ChartVLM-Base-7.3B for evaluations.

\itemnote{CogAgent}~\cite{cogagent} is a visual-linguistic model specialized in GUI understanding and planning while retaining strong capabilities across general cross-modal tasks. By leveraging both low and high-resolution image encoders, CogAgent supports input at $1120\times 1120$ resolution, enabling it to recognize even tiny page elements and text.

\subsection{Model Performance Explaination}
\label{sec_apdx_model_performance_explain}
\itemnote{OneChart}~\citep{OneChart} is a hierarchical architecture model. It trains a decoder to convert charts to CSV tables as a prompt for LLaVA-V1.6 to inference. OneChart's performance on ChartBench is abnormal and inconsistent with its performance on ChartQA. Unlike ChartQA, the metadata in ChartBench is longer, and the charts do not have data point annotations. In this case, the Python dictionary extracted by OneChart is inaccurate and results in generally longer table prompts. After analyzing specific cases, we find that OneChart always fails to follow instructions on the cases with longer prompts, even for simple yes-or-no binary outputs.

\itemnote{ChartVLM}~\cite{ChartX} is a multi-decoder structure. The router selects the corresponding decoder according to the difficulty of the current query. However, ChartVLM shows the opposite performance on \textit{Acc+} and NQA tasks (Tab.~\ref{tab_overview} 8.02\% v.s. 43.74\% in regular charts and 5.92\% v.s. 18.21\% in extra charts). Case studies show that ChartVLM tends to generate numbers or phrases, ignoring various yes/no prompt constraints. As a result, the current metric cannot parse the output of ChartVLM. However, it is worth noting that although some of ChartVLM's outputs are not strictly yes or no, they are consistent with the correct answers. While LLMs can be used to correct this bias, we have retained the original results for a fair comparison.

\itemnote{ChartLlama}~\cite{Chartllama} is a supervised fine-tuning model with LoRA~\cite{LoRA} based on LLaVA-v1.5~\cite{llava} with a large number of generated chart instruction data. As shown in Tab.~\ref{tab_overview}, ChartLlama is the best-performing model on ChartQA, but it fails to catch up with LLaVA-v1.5 on ChartBench. Notice that ChartLlama is still better than LLaVA-v1.5 on NQA tasks but performs poorly on \textit{Acc+} tasks that mainly require yes/no answers. This indicates that ChartLlama's ability to extract values is relatively good, but SFT may reduce the model's ability to follow instructions, causing it to consistently provide numerical answers instead of yes/no responses.

\itemnote{mPLUG-Owl-bloomz}~\cite{Mplug} performs well on the ChartBench generally. However, when asked to provide a concise answer consisting of only one word or phrase, it becomes difficult to control the length of the output. It tends to generate descriptive statements, which explains its poor performance on the NQA tasks of ChartBench and ChartQA. Even if we apply LLMs to extract the key information from its output statements, the results are still unsatisfactory. Considering the model's impressive performance on \textit{Acc+} tasks, we believe that mPLUG-Owl-bloomz shares a similar issue with ChartVLM. The excessive emphasis on descriptive summaries during the supervised fine-tuning process hinders the model's ability to generate short and concise content. This limitation arises from the training procedure, which prioritizes detailed and elaborate explanations rather than producing succinct answers. As a result, when tasked with generating brief responses, the model struggles to control the length of its output and tends to generate lengthy and descriptive statements instead. This issue adversely affects its performance on tasks that require concise answers, such as the ChartQA and NQA tasks in ChartBench.

\section{Experimental Settings}
\label{sec_apdx_settings}
\subsection{Evaluation Implementation}
We locally deploy 18 open-source MLLMs and conduct evaluations on A100-40G GPUs. To maintain consistency, we strictly utilize a single GPU to evaluate the \textit{Chat} version of each MLLM with the corresponding system prompt. We employ the zero-shot evaluation manner to avoid any potential data leakage and guarantee fair comparisons. It is to highlight that the choice of prompts remarkably influences the MLLMs' response. Hence, we extensively conduct experiments with several prompts and select the one yielding the best performance (see detail in Tab.~\ref{tab_zeroshot_prompt}). For NQA task, all models adopt the same constraints as ChartQA, i.e., 

\begin{center}
    \texttt{user\textbackslash nAnswer the question using a single word or phrase. \{\}\textbackslash nassistant:}
\end{center}

Although this prompt is clear enough, some models will not be generated efficiently, so we have made some adjustments to this instruction to guide the output style of models.

\subsection{Zero-shot Prompt}
\label{apdx_sec_prompt}
\begin{table}[ht]
\centering
\caption{The mapping between the template and the MLLMs is displayed. Different prompt templates can greatly affect the performance. The values we report are the best results in each template. ICL: in-context learning style. \textcolor{ao(english)}{Green}: system prompt. \textcolor{magenta}{Pink}: \textit{Acc+} instruction. \textcolor{blue}{Blue}: the judgement based on the corresponding chart. The ground truth in the judgment has been \textbf{bolded}.}
% \vspace{-10pt}
\resizebox{1\linewidth}{!}{%
\begin{tabular}{c|c|p{12cm}}
\toprule[1.5pt]
Prompt Style & Model & \multicolumn{1}{c}{Prompt Example} \\ \midrule
\multirow{6}{*}{\newline BLIP2 style} & \multirow{6}{*}{\begin{tabular}[c]{@{}c@{}} BLIP2~\cite{BLIP2} \\ CogVLM~\cite{cogvlm} \\ MiniGPT-v2~\cite{Minigptv2} \\ Internlm-Xcomposer~\cite{internlm-xcomposer} \\ ChartVLM~\cite{ChartX} \\ CogAgent~\cite{cogagent} \\ DocOwl-v1.5~\cite{docowl} \\ Internlm-Xcomposer-v2~\cite{internlm2} \end{tabular}} & \quad \newline \textcolor{ao(english)}{ \newline \newline Question}: \textcolor{blue}{According to this chart, the Rainfall in Millimeters of Months Jul is around \textbf{100.0}}. \textcolor{magenta}{Please answer yes or no}. \textcolor{ao(english)}{Answer}: \newline \newline \newline \\  
\midrule
\multirow{6}{*}{LLaVA style} & \multirow{6}{*}{\begin{tabular}[c]{@{}c@{}} LLaVA-v1.5~\cite{Improvedllava} \\ ChartLlama~\cite{Chartllama} \\ Mini-Gemini~\cite{minigemini} \end{tabular}} & \textcolor{ao(english)}{You are a data analyst, good at dealing with chart data. Please determine whether the user's judgments on this chart are correct}. \textcolor{magenta}{You only need to answer} \textcolor{red}{{[}\textbf{yes}{]} or {[}\textbf{no}{]}}.\newline \textcolor{ao(english)}{The judgment from the User is}: \textcolor{blue}{According to this chart, the Rainfall in Millimeters of Months Jul is around \textbf{100.0}.}\newline \textcolor{magenta}{Please answer} \textcolor{red}{\textbf{yes or no}}.\newline \textcolor{ao(english)}{Your Answer}:
 % & ChartLlama &  
 \\ \midrule
\multirow{6}{*}{\begin{tabular}[c]{@{}c@{}}LLaVA style\\ no or yes\end{tabular}} & \multirow{6}{*}{\begin{tabular}[c]{@{}c@{}} Qwen-VL-Chat~\cite{qwen-vl} \\ SPHINX~\cite{sphinx} \\ OneChart~\cite{OneChart} \end{tabular}} & \textcolor{ao(english)}{You are a data analyst, good at dealing with chart data. Please determine whether the user's judgments on this chart are correct}. \textcolor{magenta}{You only need to answer} \textcolor{red}{{[}\textbf{no}{]} or {[}\textbf{yes}{]}}.\newline \textcolor{ao(english)}{The judgment from the User is}: \textcolor{blue}{According to this chart, the Rainfall in Millimeters of Months Jul is around \textbf{100.0}.}\newline \textcolor{magenta}{Please answer} \textcolor{red}{\textbf{no or yes}}.\newline \textcolor{ao(english)}{Your Answer}: \\ \midrule
\multirow{11}{*}{LLaVA style ICL} & \multirow{11}{*}{\begin{tabular}[c]{@{}c@{}} InstructBLIP~\cite{InstructBLIP} \\ mPLUG-Owl-bloomz~\cite{Mplug} \\ Shikra~\cite{shikra} \\ VisualGLM~\cite{glm} \end{tabular}}  & 
\textcolor{ao(english)}{You are a data analyst, good at dealing with chart data. Please determine whether the user's judgments on this chart are correct.} \textcolor{magenta}{You only need to answer {[}yes{]} or {[}no{]}}.\newline \textcolor{ao(english)}{Here is an example}:\newline \textcolor{ao(english)}{User: \textless{}\textbf{image}\textgreater \newline User: The figure is a line chart.\newline You: yes.\newline \newline Following the above example:\newline The query from the User is}: \textcolor{blue}{According to this chart, the Rainfall in Millimeters of Months Jul is around \textbf{100.0}.}\newline \textcolor{ao(english)}{Your Answer}: \\ \bottomrule[1.5pt]
\end{tabular}
}
\vspace{-10pt}
\label{tab_zeroshot_prompt}
\end{table}

During the evaluation on ChartBench, we observe that the zero-shot performance of MLLMs is heavily influenced by the prompt templates, which indirectly reflects the current lack of robustness in MLLMs. To ensure fairness, we select the most appropriate templates used by each MLLM's official implementation for testing. In Tab.~\ref{tab_zeroshot_prompt}, we provide the corresponding mappings between the MLLMs and the prompt templates that yield the best \textit{Acc+} metric. We also test more than 10 other prompt templates, but fail to produce the best Acc+, which thus are not summarized in the table.

It is worth noting that the MLLMs tend to randomly answer the judgment questions in ChartBench if they cannot accurately comprehend the chart. Specifically, we observe a tendency for these models to favor the first option (e.g., \textit{yes} in a yes-or-no scenario). Therefore, we provide two sets of LLaVA-style prompt templates, differing only in the order of the yes-or-no options. We have performed similar operations on other templates as well, but none of the MLLMs exhibited optimal performance on these prompt templates. Therefore, we did not include specific details about them in Tab.~\ref{tab_zeroshot_prompt}.

ICL stands for \textit{In Context Learning}. We only adopt the template format as shown in Tab.~\ref{tab_zeroshot_prompt} to standardize the output of MLLMs. We do not conduct actually ICL for our evaluations. In other words, for \textit{LLaVA-style ICL}, we just adopt a single-turn dialogue, and only the queried chart is provided as the image input.

\subsection{Supervised Fine-tuning Implementation}
Using the ChartBench data, we propose an SFT baseline. Here, we introduce the basic setup of our training process. Considering the imbalance between the Acc+ and NQA content in the instruction data, we manually balance these two types of data to prevent the model from developing a prediction bias.

\itemnote{Qwen-VL-Chat.} We perform SFT for $3$ epochs using instructions. We keep the parameters of the vision encoder frozen and use LoRA to update only the LLM branch. Training is conducted with DeepSpeed's \textit{Zero2 configuration} in half-precision \textit{bf16}, with a weight decay of $0.05$. The optimizer is AdamW with adam\_beta2 set to $0.98$. The input image resolution is $448\times 448$, the batch size is $1$, and the learning rate is $2e-5$. The entire training process consumes $12$ A100 GPU days. We do not perform alignment training for the connector because Qwen-VL's connector is small and can be updated along with the LLM parameters.

\itemnote{Internlm-XComposer-v2.} We use the chart-CSV pair for alignment training over $2$ epochs, freezing the parameters of the ViT Encoder and LLM, and only updating the connector. Then, we perform $1$ epoch of supervised fine-tuning using the chart instruction data, updating both the connector and the LLM branch with LoRA. We set a learning rate of $1e-5$ and the AdamW optimizer (adam\_beta2=$0.95$). DeepSpeed's \textit{Zero2 configuration} is employed, with half-precision \textit{bf16} for parameter updates. The input image resolution is $490\times490$, and the batch size is set to $1$. This experiment approximately requires $15$ A100-GPU days.

\section{Additional Results}
\label{apdx_sec_addition_results}
In this section, we 1) expand the discussion to include the model's \textit{Acc+} (Tab.~\ref{tab_acc_plus}) and \textit{NQA} (Tab.~\ref{tab_nqa}) performance on each chart type, details of FixedCoT (Fig.~\ref{fig_fixed_CoT}), and the relationship between model performance and image resolution (Fig.~\ref{fig_resolution}); 2) provide results using accuracy as a metric (Tab.~\ref{tab_acc_chart_type} \& \ref{tab_acc_task}); 3) show evaluation results on ChartQA by image type (Tab.~\ref{tab_result_type_chartQA} \& \ref{tab_regular_contrast}); 4) present human evaluation results on ChartBench (Tab.~\ref{tab_human_eval}); 5) offer specific evaluation samples (Fig.~\ref{fig_show_case_unanno} \& \ref{fig_show_case_anno}); and 6) provide sample analyses of SOTA, i.e., GPT-4 (Fig.~\ref{fig_gpt4_vis}).

\subsection{Further Study}

\begin{table}[ht]
\centering
\caption{The zero-shot \textit{Acc+} (\%) performance w.r.t. chart types.}
\resizebox{\linewidth}{!}{
\setlength{\tabcolsep}{8pt}
\begin{tabular}{@{}ccccccccccccc@{}}
\toprule[1.5pt]
\multicolumn{1}{c|}{\multirow{2}{*}{Models}} & \multicolumn{4}{c|}{Regular Type} & \multicolumn{7}{c|}{Extra Type} & \multirow{2}{*}{\textit{\textbf{Acc+}}} \\ \cmidrule(lr){2-12}
\multicolumn{1}{l|}{} & Line & Bar & Pie & \multicolumn{1}{c|}{Avg.} & Area & Box & Radar & Scatter & Node & Combin. & \multicolumn{1}{c|}{Avg.} &  \\ \midrule
\multicolumn{13}{l}{\quad \textit{Open source MLLMs}} \\
\multicolumn{1}{l|}{VisualGLM~\cite{glm}} & 10.80 & 1.96 & 0.00 & \multicolumn{1}{c|}{3.46} & 1.17 & 8.50 & 0.25 & 3.33 & 15.50 & 5.13 & \multicolumn{1}{c|}{4.22} & 3.79 \\
\multicolumn{1}{l|}{ChartVLM~\cite{ChartX}} & 10.70 & 8.04 & 4.62 & \multicolumn{1}{c|}{8.02} & 7.67 & 6.67 & 5.25 & 5.50 & 0.00 & 6.50 & \multicolumn{1}{c|}{5.92} & 6.90 \\
\multicolumn{1}{l|}{Shikra~\cite{shikra}} & 7.40 & 10.62 & 4.50 & \multicolumn{1}{c|}{8.59} & 6.00 & 11.33 & 11.88 & 4.17 & 8.50 & 3.63 & \multicolumn{1}{c|}{7.50} & 8.11 \\
\multicolumn{1}{l|}{OneChart~\cite{OneChart}} & 15.10 & 12.27 & 9.12 & \multicolumn{1}{c|}{12.34} & 7.00 & 7.33 & 2.75 & 6.33 & \textbf{53.50} & 7.75 & \multicolumn{1}{c|}{8.75} & 12.04 \\
\multicolumn{1}{l|}{InstructBLIP~\cite{InstructBLIP}} & 24.40 & 15.04 & 19.10 & \multicolumn{1}{c|}{17.96} & 4.33 & 7.33 & 2.00 & 12.50 & 9.00 & 2.38 & \multicolumn{1}{c|}{5.50} & 12.49 \\
\multicolumn{1}{l|}{CogVLM~\cite{cogvlm}} & 10.50 & 14.58 & 17.90 & \multicolumn{1}{c|}{14.41} & 12.50 & 9.67 & 16.00 & 14.33 & 16.00 & 6.13 & \multicolumn{1}{c|}{11.89} & 13.30 \\
\multicolumn{1}{l|}{Internlm-XComposer~\cite{internlm-xcomposer}} & 16.00 & 20.42 & 21.50 & \multicolumn{1}{c|}{19.70} & 4.50 & 14.50 & 15.00 & 12.00 & 8.50 & 5.13 & \multicolumn{1}{c|}{10.11} & 15.49 \\
\multicolumn{1}{l|}{SPHINX~\cite{sphinx}} & 18.40 & 15.54 & 23.40 & \multicolumn{1}{c|}{17.87} & 12.00 & 8.17 & 19.00 & 17.17 & 31.00 & 25.88 & \multicolumn{1}{c|}{17.92} & 17.89 \\
\multicolumn{1}{l|}{CogAgent~\cite{cogagent}} & 18.60 & 23.96 & 11.00 & \multicolumn{1}{c|}{20.39} & 15.67 & 16.50 & 9.38 & 11.67 & 27.50 & 15.50 & \multicolumn{1}{c|}{14.36} & 18.07 \\
\multicolumn{1}{l|}{BLIP2~\cite{BLIP2}} & 29.60 & 17.35 & 24.90 & \multicolumn{1}{c|}{21.65} & 6.17 & 10.67 & 17.63 & 22.00 & 33.00 & 28.00 & \multicolumn{1}{c|}{18.44} & 20.24 \\
\multicolumn{1}{l|}{ChartLlama~\cite{Chartllama}} & 28.90 & 19.35 & 22.10 & \multicolumn{1}{c|}{22.02} & 16.50 & 13.33 & 25.00 & 28.50 & 25.50 & 26.38 & \multicolumn{1}{c|}{22.56} & 22.26 \\
\multicolumn{1}{l|}{MiniGPT-v2~\cite{Minigptv2}} & 26.70 & 21.54 & 20.20 & \multicolumn{1}{c|}{22.37} & 21.67 & 24.67 & 25.88 & 28.17 & 15.50 & 27.13 & \multicolumn{1}{c|}{25.06} & 23.55 \\
\multicolumn{1}{l|}{LLaVA-v1.5~\cite{llava}} & 34.40 & 24.73 & 19.10 & \multicolumn{1}{c|}{25.61} & 26.83 & 25.67 & 28.63 & 26.00 & 33.50 & 27.38 & \multicolumn{1}{c|}{27.39} & 26.39 \\
\multicolumn{1}{l|}{mPLUG-Owl-bloomz~\cite{Mplug}} & 37.50 & 24.73 & 26.10 & \multicolumn{1}{c|}{27.80} & 21.33 & 25.83 & 26.50 & 24.17 & 28.50 & 27.50 & \multicolumn{1}{c|}{25.47} & 26.78 \\
\multicolumn{1}{l|}{Qwen-VL-Chat~\cite{qwen-vl}} & 41.00 & 20.96 & 40.00 & \multicolumn{1}{c|}{29.46} & 28.83 & 24.17 & 35.00 & 19.50 & 18.50 & 25.50 & \multicolumn{1}{c|}{26.56} & 28.18 \\
\multicolumn{1}{l|}{DocOwl-v1.5~\cite{docowl}} & 49.10 & 31.08 & 31.62 & \multicolumn{1}{c|}{35.27} & 12.17 & 24.00 & 20.50 & 35.33 & 26.00 & 40.25 & \multicolumn{1}{c|}{26.86} & 31.62 \\
\multicolumn{1}{l|}{Mini-Gemini~\cite{minigemini}} & 37.60 & 40.19 & 40.00 & \multicolumn{1}{c|}{39.57} & \textbf{36.83} & 26.50 & 30.00 & 37.17 & 43.00 & 27.00 & \multicolumn{1}{c|}{31.81} & 36.54 \\
\multicolumn{1}{l|}{Internlm-XComposer-v2~\cite{internlm2}} & \textbf{70.60} & \textbf{51.50} & \textbf{62.75} & \multicolumn{1}{c|}{\textbf{57.89}} & 30.17 & \textbf{31.33} & \textbf{43.50} & \textbf{52.00} & 52.50 & \textbf{46.12} & \multicolumn{1}{c|}{\textbf{41.75}} & \textbf{51.34} \\ \midrule
\multicolumn{13}{l}{\quad \textit{Closed source MLLMs}} \\
\multicolumn{1}{l|}{ERNIE~\cite{wenxin}} & 44.00 & 45.00 & 57.00 & \multicolumn{1}{c|}{47.39} & \textbf{45.00} & 30.00 & 40.00 & 51.67 & 70.00 & 56.25 & \multicolumn{1}{c|}{46.39} & 46.95 \\
\multicolumn{1}{l|}{GPT-4V~\cite{GPT4}} & 74.00 & 41.54 & 63.00 & \multicolumn{1}{c|}{53.26} & 33.30 & 46.67 & \textbf{57.50} & 70.00 & \textbf{100.00} & 56.25 & \multicolumn{1}{c|}{55.83} & 54.39 \\
\multicolumn{1}{l|}{GPT-4O~\cite{GPT4}} & \textbf{86.00} & \textbf{51.92} & \textbf{78.00} & \multicolumn{1}{c|}{\textbf{65.00}} & 36.67 & \textbf{63.33} & \textbf{57.50} & \textbf{83.33} & \textbf{100.00} & \textbf{65.00} & \multicolumn{1}{c|}{\textbf{63.33}} & \textbf{64.27} \\ \bottomrule[1.5pt]
\end{tabular}
}
\label{tab_acc_plus}
\end{table}

\begin{table}[ht]
\centering
\caption{The zero-shot \textit{NQA} (\%) performance w.r.t. chart types.}
\resizebox{\linewidth}{!}{
\setlength{\tabcolsep}{8pt}
\begin{tabular}{@{}ccccccccccccc@{}}
\toprule[1.5pt]
\multicolumn{1}{c|}{\multirow{2}{*}{Models}} & \multicolumn{4}{c|}{Regular Type} & \multicolumn{7}{c|}{Extra Type} & \multirow{2}{*}{\textbf{NQA}} \\ \cmidrule(lr){2-12}
\multicolumn{1}{l|}{} & Line & Bar & Pie & \multicolumn{1}{c|}{Avg.} & Area & Box & Radar & Scatter & Node & Combin. & \multicolumn{1}{c|}{Avg.} &  \\ \midrule
\multicolumn{13}{l}{\quad \textit{Open source MLLMs}} \\
\multicolumn{1}{l|}{BLIP2~\cite{BLIP2}} & 0.80 & 1.38 & 0.00 & \multicolumn{1}{c|}{0.96} & 0.00 & 0.67 & 4.00 & 2.67 & 31.00 & 1.00 & \multicolumn{1}{c|}{4.84} & 2.71 \\
\multicolumn{1}{l|}{OneChart~\cite{OneChart}} & 1.20 & 2.31 & 3.20 & \multicolumn{1}{c|}{2.26} & 0.00 & 1.33 & 0.50 & 10.67 & 6.00 & 3.50 & \multicolumn{1}{c|}{3.37} & 2.76 \\
\multicolumn{1}{l|}{InstructBLIP~\cite{InstructBLIP}} & 0.40 & 1.23 & 0.40 & \multicolumn{1}{c|}{0.87} & 1.33 & 0.67 & 0.50 & 0.00 & 46.00 & 0.50 & \multicolumn{1}{c|}{5.37} & 2.90 \\
\multicolumn{1}{l|}{VisualGLM~\cite{glm}} & 1.20 & 2.77 & 0.00 & \multicolumn{1}{c|}{1.83} & 0.00 & 0.67 & 0.50 & 2.67 & 38.00 & 1.00 & \multicolumn{1}{c|}{4.84} & 3.19 \\
\multicolumn{1}{l|}{Internlm-XComposer~\cite{internlm-xcomposer}} & 0.80 & 1.54 & 0.80 & \multicolumn{1}{c|}{1.22} & 2.67 & 0.00 & 2.00 & 1.33 & 43.00 & 1.00 & \multicolumn{1}{c|}{5.79} & 3.29 \\
\multicolumn{1}{l|}{MiniGPT-v2~\cite{Minigptv2}} & 2.80 & 1.85 & 3.60 & \multicolumn{1}{c|}{2.43} & 2.00 & 0.67 & 3.00 & 3.33 & 30.00 & 2.50 & \multicolumn{1}{c|}{5.26} & 3.71 \\
\multicolumn{1}{l|}{mPLUG-Owl-bloomz~\cite{Mplug}} & 0.40 & 2.77 & 3.20 & \multicolumn{1}{c|}{2.35} & 0.00 & 0.67 & 11.00 & 0.67 & 33.00 & 1.00 & \multicolumn{1}{c|}{6.21} & 4.10 \\
\multicolumn{1}{l|}{Shikra~\cite{shikra}} & 2.40 & 1.85 & 3.60 & \multicolumn{1}{c|}{2.35} & 2.00 & 2.00 & 8.50 & 2.67 & 52.00 & 3.50 & \multicolumn{1}{c|}{9.05} & 5.38 \\
\multicolumn{1}{l|}{SPHINX~\cite{sphinx}} & 4.80 & 6.31 & 7.20 & \multicolumn{1}{c|}{6.17} & 2.00 & 0.67 & 15.00 & 13.33 & \textbf{53.00} & 7.00 & \multicolumn{1}{c|}{12.74} & 9.14 \\
\multicolumn{1}{l|}{LLaVA-v1.5~\cite{llava}} & 8.00 & 7.38 & 10.00 & \multicolumn{1}{c|}{8.09} & 1.33 & 2.00 & 23.00 & 13.33 & 50.00 & 12.00 & \multicolumn{1}{c|}{15.26} & 11.33 \\
\multicolumn{1}{l|}{CogVLM~\cite{cogvlm}} & 9.60 & 12.46 & 17.60 & \multicolumn{1}{c|}{12.96} & 3.33 & 1.33 & 26.00 & 14.67 & 23.00 & 13.00 & \multicolumn{1}{c|}{13.68} & 13.29 \\
\multicolumn{1}{l|}{ChartLlama~\cite{Chartllama}} & 18.40 & 16.77 & 15.60 & \multicolumn{1}{c|}{16.87} & 5.33 & 6.67 & 21.50 & 24.67 & 29.00 & 23.50 & \multicolumn{1}{c|}{18.32} & 17.52 \\
\multicolumn{1}{l|}{Qwen-VL-Chat~\cite{qwen-vl}} & 26.00 & 19.69 & 31.20 & \multicolumn{1}{c|}{23.57} & 6.00 & 7.33 & 26.00 & 29.33 & 23.00 & 30.50 & \multicolumn{1}{c|}{21.05} & 22.43 \\
\multicolumn{1}{l|}{Mini-Gemini~\cite{minigemini}} & 24.00 & 19.85 & \textbf{42.00} & \multicolumn{1}{c|}{25.57} & 8.67 & 10.67 & \textbf{33.00} & 27.33 & 46.00 & 31.50 & \multicolumn{1}{c|}{25.79} & 25.67 \\
\multicolumn{1}{l|}{CogAgent~\cite{cogagent}} & 39.20 & 18.92 & 34.00 & \multicolumn{1}{c|}{26.61} & 3.33 & 11.33 & 27.50 & 50.67 & 21.00 & 35.50 & \multicolumn{1}{c|}{25.79} & 26.24 \\
\multicolumn{1}{l|}{ChartVLM~\cite{ChartX}} & \textbf{66.80} & \textbf{38.62} & 34.00 & \multicolumn{1}{c|}{\textbf{43.74}} & 6.67 & 12.67 & 19.00 & 17.33 & 27.00 & 26.50 & \multicolumn{1}{c|}{18.21} & 32.19 \\
\multicolumn{1}{l|}{DocOwl-v1.5~\cite{docowl}} & 51.60 & 34.15 & 31.20 & \multicolumn{1}{c|}{37.30} & 12.67 & \textbf{20.67} & 30.50 & 39.33 & 44.00 & 33.00 & \multicolumn{1}{c|}{29.47} & 33.76 \\
\multicolumn{1}{l|}{Internlm-XComposer-v2~\cite{internlm2}} & 58.40 & 37.69 & 32.00 & \multicolumn{1}{c|}{40.96} & \textbf{16.67} & 1.33 & 26.50 & \textbf{56.67} & 42.00 & \textbf{46.50} & \multicolumn{1}{c|}{\textbf{31.58}} & \textbf{36.71} \\ \midrule
\multicolumn{13}{l}{\quad \textit{Closed source MLLMs}} \\
\multicolumn{1}{l|}{ERNIE~\cite{wenxin}} & 36.00 & 19.23 & 32.42 & \multicolumn{1}{c|}{25.74} & 5.32 & 13.33 & 20.00 & 60.00 & \textbf{100.00} & 30.00 & \multicolumn{1}{c|}{33.47} & 29.24 \\
\multicolumn{1}{l|}{GPT-4V~\cite{GPT4}} & 48.00 & 24.62 & \textbf{40.00} & \multicolumn{1}{c|}{33.04} & 6.67 & 26.67 & 25.00 & 66.67 & 80.00 & 50.00 & \multicolumn{1}{c|}{40.00} & 36.19 \\
\multicolumn{1}{l|}{GPT-4O~\cite{GPT4}} & \textbf{72.00} & \textbf{29.00} & 36.00 & \multicolumn{1}{c|}{\textbf{40.00}} & \textbf{7.00} & \textbf{47.00} & \textbf{35.00} & \textbf{73.00} & 20.00 & \textbf{60.00} & \multicolumn{1}{c|}{\textbf{41.05}} & \textbf{40.48} \\ 
\bottomrule[1.5pt]
\end{tabular}
}
\label{tab_nqa}
\end{table}

\myparagraph{Results w.r.t. Chart Types.} Tab.~\ref{tab_acc_plus} \& \ref{tab_nqa} illustrate the performance of \textit{Acc+} and \textit{GPT-acc} w.r.t. chart types. In general, the current MLLMs demonstrate limited proficiency in chart recognition and encounter significant challenges. For certain chart types (e.g., radar or combination chart), some MLLMs achieve close to 0\% \textit{Acc+}, indicating their inability to extract key information from charts and insensitivity to both positive and negative interrogations. Note that the \textit{Acc+} metric approaches 0\% under random guessing, as discussed in Sec.~\ref{sec_method_metrics}. We also provide results of the vanilla accuracy metric in Appendix~\ref{apdx_sec_accuracy}, where the baseline should be 50\%.

Specifically, some MLLMs like Qwen-VL-Chat and mPLUG-Owl demonstrate satisfying chart recognition capabilities, which may be attributed to their instruction tuning on chart data. The corresponding performance is lower than their reported results in ChartQA~\citep{ChartQA,Chartllama}, primarily because their chart recognition depends on OCR capability rather than robust visual logical reasoning. In ChartBench, the proportion of annotated charts is notably low (about 20\% in Tab.~\ref{tab_chartbench_distribution}). The majority of queries demand MLLMs to employ visual, logical reasoning, which is quite challenging for these models. VisualGLM and Shikra perform poorly, possibly due to their smaller LLM sizes and weaker visual encoding branches. MLLMs exhibit satisfactory performance on regular charts, but there is still substantial potential for improvement when it comes to handling more intricate graphics.

\begin{figure}[ht]
    \begin{minipage}{0.48\textwidth}
      \centering
      \begin{overpic}[width=\linewidth, grid=False]{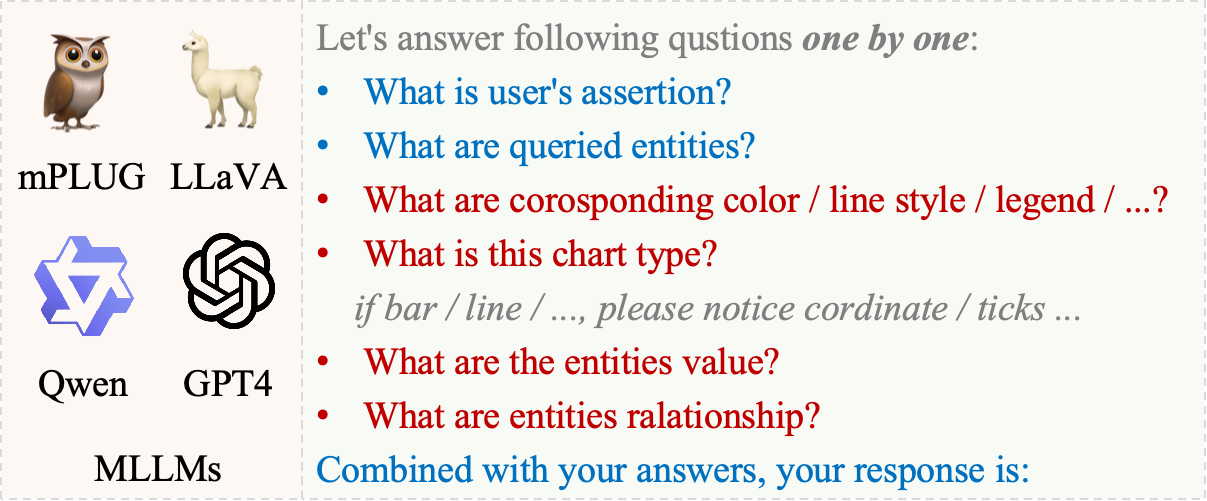}
      \end{overpic}
      \caption{The proposed FixedCoT. \textcolor{blue}{Blue} and \textcolor{red}{red} color questions indicate textual and visual reasoning, respectively.}
      \vspace{-15pt}
      \label{fig_fixed_CoT}
    \end{minipage}
    \hfill
    \begin{minipage}{0.48\textwidth}
      \centering
      \begin{overpic}[width=\linewidth, grid=False]{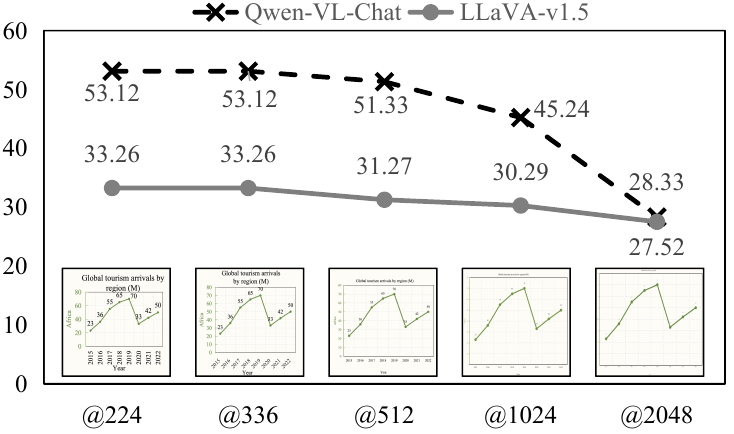}
      \end{overpic}
      \caption{The zero-shot \textit{Acc+} (\%) w.r.t. query chart resolution.}
      \vspace{-10pt}
      \label{fig_resolution}
    \end{minipage}
\end{figure}

\myparagraph{Fixed Chart CoT.}
In Fig.~\ref{fig_CoT}, we mention using a fixed template for CoT, with detailed content shown in Fig.~\ref{fig_fixed_CoT}. Thanks to the expanded chart types, we can summarize some common approaches to understanding each type of chart. For example, we can identify the main subject of the question and the objects being queried, then guide the model to focus on the locations and spatial relationships of these objects. Although we cannot specify the exact logical relationships between these elements (as they depend on the specific content of each chart), guiding the model to prioritize commonly occurring logic can still enhance overall performance.

\myparagraph{Chart resolution.}
The visual branch of MLLMs typically scales images to a fixed pixel size, e.g., Qwen-VL-Chat is 448px, and LLaVA-v1.5 is 336px by default. To investigate the impact of resolution, we select a part of annotated regular charts from ChartBench and adjust them to 5-level resolutions using \textit{Matplotlib} while keeping the font size unchanged. We ensure that each resolution is clear and legible for humans. Fig.~\ref{fig_resolution} illustrates the performance of Qwen-VL-Chat and LLaVA-v1.5 at different resolutions. As the resolution increases, the scaled annotations gradually become unreadable for OCR, resulting in a decline in MLLMs' performance. Qwen-VL-Chat exhibits larger performance drops than LLaVA-v1.5, indicating a greater reliance on OCR.

\subsection{Results of Accuracy Metric}
\label{apdx_sec_accuracy}

\begin{table}[ht]
\centering
\caption{The zero-shot \textit{Accuracy} (\%) performance w.r.t. chart types in ChartBench. We report the results of the best-performing prompt for each MLLM.}
\setlength{\tabcolsep}{10pt}
\resizebox{\linewidth}{!}{
\begin{tabular}{@{}l|cccc|ccccccc|c@{}}
\toprule[1.5pt]
\multicolumn{1}{c|}{\multirow{2}{*}{Models}} & \multicolumn{4}{c|}{Regular Type} & \multicolumn{7}{c|}{Extra Type} & \multirow{2}{*}{Avg.} \\ \cmidrule(lr){2-12}
\multicolumn{1}{c|}{} & Line & Bar & Pie & Avg. & Area & Box & Radar & Scatter & Node & Combin. & Avg. &  \\ \midrule
\multicolumn{13}{l}{\quad \textit{Open source MLLMs}} \\
mPLUG-Owl-bloomz~\cite{Mplug} & 56.55 & 49.87 & 49.19 & 51.26 & 46.75 & 48.50 & 50.44 & 47.58 & 47.25 & 49.06 & 48.47 & 49.94 \\
Shikra~\cite{shikra} & 50.35 & 50.75 & 50.00 & 50.52 & 51.33 & 50.17 & 49.94 & 47.17 & 47.75 & 49.94 & 49.53 & 50.05 \\
MiniGPT-v2~\cite{Minigptv2} & 52.80 & 50.21 & 48.88 & 50.56 & 50.25 & 52.67 & 52.25 & 53.92 & 39.00 & 54.25 & 51.29 & 50.79 \\
ChartVLM~\cite{ChartX} & 53.60 & 51.77 & 50.75 & 52.00 & 51.67 & 51.83 & 50.19 & 51.08 & 45.50 & 51.25 & 50.83 & 51.34 \\
VisualGLM~\cite{glm} & 55.40 & 50.98 & 49.69 & 51.75 & 47.92 & 53.75 & 49.00 & 49.00 & 55.75 & 52.44 & 51.01 & 51.37 \\
OneChart~\cite{OneChart} & 55.15 & 54.50 & 53.81 & 54.52 & 51.92 & 51.00 & 49.56 & 51.42 & \textbf{70.25} & 52.44 & 52.29 & 54.02 \\
InstructBLIP~\cite{InstructBLIP} & 62.15 & 56.33 & 59.13 & 58.16 & 48.50 & 50.08 & 47.50 & 55.50 & 54.25 & 47.19 & 49.97 & 54.45 \\
SPHINX~\cite{sphinx} & 55.40 & 53.40 & 52.25 & 53.65 & 53.00 & 51.25 & 54.50 & 53.75 & 62.75 & 58.50 & 55.34 & 54.51 \\
LLaVA-v1.5~\cite{Improvedllava} & 60.00 & 54.58 & 47.06 & 54.44 & 57.67 & 54.92 & 58.63 & 55.58 & 48.00 & 55.38 & 55.61 & 54.75 \\
Internlm-XComposer~\cite{internlm-xcomposer} & 55.70 & 57.38 & 55.31 & 56.62 & 51.17 & 54.42 & 53.50 & 54.08 & 54.00 & 51.25 & 52.95 & 54.96 \\
CogVLM~\cite{cogvlm} & 54.40 & 56.27 & 56.50 & 55.89 & 55.50 & 53.08 & 55.25 & 56.42 & 55.25 & 51.19 & 54.28 & 55.26 \\
BLIP2~\cite{BLIP2} & 62.80 & 57.33 & 60.38 & 59.13 & 52.42 & 53.58 & 56.69 & 58.58 & 41.00 & 61.44 & 55.17 & 57.45 \\
CogAgent~\cite{cogagent} & 59.30 & 61.96 & 55.50 & 60.18 & 57.83 & 57.83 & 54.44 & 55.42 & 31.00 & 57.50 & 55.11 & 57.45 \\
ChartLlama~\cite{Chartllama} & 61.70 & 56.48 & 57.50 & 57.85 & 57.25 & 52.75 & 61.31 & 61.50 & 39.75 & 60.69 & 56.95 & 57.54 \\
Qwen-VL-Chat~\cite{qwen-vl} & 69.00 & 57.77 & 66.50 & 61.91 & \textbf{63.17} & 57.50 & 63.62 & 56.75 & 55.50 & 58.63 & 59.59 & 61.11 \\
DocOwl-v1.5~\cite{docowl} & 72.65 & 62.92 & 63.44 & 65.23 & 52.42 & 54.67 & 52.81 & 65.17 & 52.50 & \textbf{66.25} & 58.08 & 61.83 \\
Mini-Gemini~\cite{minigemini} & 65.15 & 65.42 & 66.12 & 65.49 & 62.75 & 57.33 & 58.38 & 61.67 & 66.25 & 55.81 & 59.35 & 62.86 \\
Internlm-XComposer-v2~\cite{internlm2} & \textbf{84.30} & \textbf{73.83} & \textbf{79.00} & \textbf{77.15} & 57.83 & \textbf{60.50} & \textbf{67.44} & \textbf{73.58} & 67.00 & 66.00 & \textbf{65.36} & \textbf{72.23} \\ \midrule
\multicolumn{13}{l}{\quad \textit{Closed source MLLMs}} \\
ERNIE~\cite{wenxin} & 61.00 & 65.58 & 71.25 & 65.57 & \textbf{68.33} & 52.50 & 65.62 & 68.33 & 82.50 & 73.12 & 67.76 & 66.67 \\ 
GPT-4V~\cite{GPT4} & 84.50 & 68.08 & 78.75 & 73.75 & 62.50 & 65.83 & \textbf{69.38} & 82.50 & \textbf{100.00} & 73.12 & 73.82 & 74.11 \\
GPT-4O~\cite{GPT4} & \textbf{90.50} & \textbf{70.58} & \textbf{82.50} & \textbf{77.27} & 61.67 & \textbf{77.50} & 67.50 & \textbf{91.67} & \textbf{100.00} & \textbf{79.38} & \textbf{77.89} & \textbf{78.10} \\ 
\bottomrule[1.5pt]
\end{tabular}
}
\label{tab_acc_chart_type}
\vspace{-10pt}
\end{table}
\begin{table}[ht]
\centering
\setlength{\tabcolsep}{35pt}
\caption{The zero-shot \textit{Accuracy} (\%) performance w.r.t. chart tasks in ChartBench. We report the results of the best-performing prompt for each MLLM.}
\resizebox{\linewidth}{!}{
\begin{tabular}{@{}l|ccccc|c}
\toprule[1.5pt]
\multicolumn{1}{c|}{\multirow{2}{*}{Models}} & \multicolumn{5}{c|}{Task Type} & \multicolumn{1}{c}{\multirow{2}{*}{Avg.}} \\ \cmidrule(lr){2-6}
\multicolumn{1}{c|}{} & CR & VE & VC & GC & NQA &  \\ \midrule
\multicolumn{7}{l}{\textit{\quad Open source MLLMs}} \\
mPLUG-Owl-bloomz~\cite{Mplug} & 50.43 & 50.05 & 49.83 & 49.45 & 4.10 & 40.77 \\
Shikra~\cite{shikra} & 49.98 & 50.31 & 50.14 & 49.79 & 5.38 & 41.12 \\
MiniGPT-v2~\cite{Minigptv2} & 53.67 & 49.57 & 50.95 & 48.98 & 3.71 & 41.38 \\
VisualGLM~\cite{glm} & 55.88 & 49.83 & 49.90 & 49.86 & 3.19 & 41.73 \\
OneChart~\cite{OneChart} & 50.88 & 56.55 & 54.43 & 54.21 & 2.76 & 43.77 \\
InstructBLIP~\cite{InstructBLIP} & 67.90 & 50.00 & 49.95 & 49.95 & 2.90 & 44.14 \\
Internlm-XComposer~\cite{internlm-xcomposer} & 70.76 & 49.43 & 50.76 & 48.90 & 3.29 & 44.63 \\
SPHINX~\cite{sphinx} & 64.21 & 50.71 & 53.02 & 50.07 & 9.14 & 45.43 \\
LLaVA-v1.5~\cite{Improvedllava} & 65.98 & 48.93 & 54.29 & 49.81 & 11.33 & 46.07 \\
BLIP2~\cite{BLIP2} & 78.57 & 48.88 & 53.48 & 48.86 & 2.71 & 46.50 \\
CogVLM~\cite{cogvlm} & 64.07 & 49.98 & 54.57 & 52.40 & 13.29 & 46.86 \\
ChartVLM~\cite{ChartX} & 50.00 & 51.79 & 51.95 & 51.62 & 32.19 & 47.51 \\
ChartLlama~\cite{Chartllama} & 71.95 & 50.45 & 55.17 & 52.57 & 17.52 & 49.53 \\
CogAgent~\cite{cogagent} & 81.12 & 48.64 & 51.45 & 48.57 & 26.24 & 51.20 \\
Qwen-VL-Chat~\cite{qwen-vl} & 73.02 & 53.43 & 58.86 & 59.14 & 22.43 & 53.38 \\
Mini-Gemini~\cite{minigemini} & \textbf{88.95} & 52.17 & 55.48 & 54.83 & 25.67 & 55.42 \\
DocOwl-v1.5~\cite{docowl} & 62.95 & 63.60 & 58.69 & 62.07 & 33.76 & 56.21 \\
Internlm-XComposer-v2~\cite{internlm2} & 83.41 & \textbf{65.49} & \textbf{68.49} & \textbf{71.54} & \textbf{36.71} & \textbf{65.13} \\ \midrule
\multicolumn{7}{l}{\textit{\quad Closed source MLLMs}} \\ 
ERNIE~\cite{wenxin} & 75.00 & 67.14 & 53.57 & 70.95 & 16.19 & 56.57 \\
GPT-4V~\cite{GPT4} & 97.62 & 62.86 & 65.95 & 70.00 & 36.19 & 66.52 \\
GPT-4O~\cite{GPT4} & \textbf{98.33} & \textbf{65.71} & \textbf{74.29} & \textbf{74.05} & \textbf{40.48} & \textbf{70.57} \\
\bottomrule[1.5pt]
\end{tabular}
}
\vspace{-10pt}
\label{tab_acc_task}
\end{table}

Accuracy is the most widely used evaluation criterion for true/false or multiple-choice questions, but it has inherent limitations. Firstly, for difficult questions, accuracy struggles to distinguish between genuine answers and random guesses, both of which can yield performance close to the baseline (e.g., 50\% for true/false questions, 25\% for four-choice questions). Secondly, accuracy places high demands on data scale. In the case of the accuracy metric, if the test sample approaches infinity, the performance of random guessing would converge to the baseline. Conversely, with a small data scale, random guessing might produce results significantly higher than the baseline. Although ChartBench provides \textit{16.8K} judgment QA pairs (consisting of \textit{8.4K} original questions and their counterparts), this quantity still cannot completely eliminate the occurrence of the situations above (e.g., the accuracy of MiniGPT-v2 on Node chart in Tab.~\ref{tab_acc_chart_type}).

In Tab.~\ref{tab_acc_chart_type} and Tab.~\ref{tab_acc_task}, we present the results using Accuracy (abbreviated as \textit{Acc.}) as the metric. Overall, Internlm-Xcomposer-v2 continues to demonstrate the best performance, consistent with the trend shown by \textit{Acc+} in Tab.~\ref{tab_overview}. However, there are differences between accuracy and \textit{Acc+} in terms of specific details. InternLM-Xcomposer achieves 55.70\% accuracy in Tab~\ref{tab_acc_chart_type}, while its \textit{Acc+} performance is just 15.49\% (Tab.~\ref{tab_overview}), indicating that a significant portion of its correct answers are the result of random guessing. This is further confirmed by the \textit{CoR} metric in Tab.~\ref{tab_cor}. From Tab.~\ref{tab_acc_task}, it can be observed that accuracy does not effectively differentiate between tasks of varying difficulty, as it shows results close to the baseline of 50\% across all 5 tasks. Compared with Tab.~\ref{tab_task_type}, it is evident that the VE and GC tasks are notably more challenging, as they require MLLMs to rely on more visual cues for reasoning. The above analysis demonstrates that the improved \textit{Acc+} metric enables more robust evaluations.

Our improved metric, \textit{Acc+}, effectively addresses the two limitations of accuracy mentioned above. The \textit{Acc+} metric requires MLLMs to provide accurate judgments for both positive and negative perspectives regarding the base assertions. This innovative metric offers two distinct advantages. Firstly, it ensures consistency between positive and negative queries, with the only difference being the Ground Truth value. This precautionary approach reduces the chance of lucky guesses resulting from random choices, as MLLMs may produce identical responses for both query types if they fail to comprehend the chart. Secondly, the GT values for negative queries are derived from other data within the same chart, eliminating unrealistic scenarios and enhancing the validity of the evaluation process. Generally, the expected probability of random guessing is 25\% for vanilla \textit{Acc+}. However, for the MLLM that has insufficient chart recognition capabilities, \textit{the CoR tends to be 100\%, and thus the Acc+ tends to be 0\% instead of 25\% baseline}. This characteristic enables \textit{Acc+} to accurately reflect the model's chart comprehension ability even when the dataset is small in size.

\subsection{Results of ChartQA}
\begin{table*}[h]
\centering
\caption{The zero-shot \textit{Acc} (\%) performance w.r.t. chart types in ChartQA. For bar chart, We report the average score of horizontal and vertical bars in ChartQA.}
\resizebox{\linewidth}{!}{
\setlength{\tabcolsep}{10pt}
\begin{tabular}{@{}l|cccc|cccc|c@{}}
\toprule[1.5pt]
\multicolumn{1}{c|}{\multirow{2}{*}{Models}} & \multicolumn{4}{c|}{Human}          & \multicolumn{4}{c|}{Augmented}                                         & \multirow{2}{*}{Acc.} \\ 
\cmidrule(lr){2-9}
\multicolumn{1}{c|}{}                        & Line  & Bar   & Pie & \multicolumn{1}{c|}{Avg.}   & Line  & Bar   & Pie    & \multicolumn{1}{c|}{Avg.} &                       \\      
\midrule
BLIP2~\cite{BLIP2} & 14.34 & 9.69 & 7.24 & 10.40 & 6.20 & 5.18 & 0.00 & 5.20 & 7.80 \\
mPLUG-Owl-bloomz~\cite{Mplug} & 22.79 & 10.53 & 6.58 & 12.72 & 7.75 & 5.72 & 5.00 & 5.92 & 9.32 \\
Shikra~\cite{shikra} & 25.00 & 13.68 & 13.82 & 16.16 & 8.53 & 7.27 & 0.00 & 7.28 & 11.72 \\
InstructBLIP~\cite{InstructBLIP} & 29.78 & 11.86 & 10.53 & 15.60 & 10.08 & 9.81 & 10.00 & 9.84 & 12.72 \\
VisualGLM~\cite{glm} & 32.35 & 14.89 & 7.24 & 17.76 & 9.30 & 7.81 & 5.00 & 7.92 & 12.84 \\
Internlm-XComposer~\cite{internlm-xcomposer} & 31.99 & 13.20 & 9.21 & 16.80 & 9.30 & 9.17 & 20.00 & 9.36 & 13.08 \\
MiniGPT-v2~\cite{Minigptv2} & 33.09 & 16.22 & 11.18 & 19.28 & 9.30 & 10.99 & 10.00 & 10.80 & 15.04 \\
SPHINX~\cite{sphinx} & 35.66 & 17.68 & 16.45 & 21.44 & 10.08 & 11.35 & 10.00 & 11.20 & 16.32 \\
LLaVA-v1.5~\cite{Improvedllava} & 39.71 & 19.01 & 16.45 & 23.20 & 9.30 & 14.26 & 15.00 & 13.76 & 18.48 \\
CogVLM~\cite{cogvlm} & 48.90 & 29.41 & 34.21 & 34.24 & 17.83 & 29.88 & 25.00 & 28.56 & 31.40 \\
Mini-Gemini~\cite{minigemini} & 55.88 & 40.68 & 43.42 & 44.32 & 43.41 & 58.31 & 75.00 & 57.04 & 50.68 \\
Qwen-VL-Chat~\cite{qwen-vl} & 54.41 & 38.38 & 43.42 & 42.48 & 55.04 & 77.48 & 80.00 & 75.20 & 58.84 \\
ChartVLM~\cite{ChartX} & 48.90 & 39.59 & 43.42 & 42.08 & 69.77 & 83.92 & 85.00 & 82.48 & 62.28 \\
OneChart~\cite{OneChart} & - & - & - & \textbf{85.30} & - & - & - & 49.10 & 67.20 \\
CogAgent~\cite{cogagent} & 65.44 & 49.88 & 56.58 & 54.08 & 62.02 & 82.74 & 80.00 & 80.56 & 67.32 \\
DocOwl-v1.5~\cite{docowl} & 57.72 & 44.79 & 50.00 & 48.24 & 68.22 & 88.92 & 85.00 & 86.72 & 67.48 \\
Internlm-XComposer-v2~\cite{internlm2} & 65.81 & \textbf{61.38} & \textbf{67.76} & 63.12 & 78.29 & 82.11 & 95.00 & 81.92 & 72.64 \\
ChartLlama~\cite{Chartllama} & \textbf{68.75} & 53.63 & 65.79 & 58.40 & \textbf{79.84} & \textbf{94.55} & \textbf{100.00} & \textbf{93.12} & \textbf{75.76} \\
\bottomrule[1.5pt]
\end{tabular}
}
\vspace{-10pt}
\label{tab_result_type_chartQA}
\end{table*}
\begin{table*}[ht]
\centering
\caption{The zero-shot \textit{Acc+} (\%) and \textit{Acc} (\%) performance in ChartBench and ChartQA respectively w.r.t \textbf{\textit{regular}} chart types. We report the results of the best-performing prompt for each MLLM.}
\resizebox{\linewidth}{!}{
\setlength{\tabcolsep}{8pt}
\begin{tabular}{@{}l|cc|cc|cc|cc|c@{}}
\toprule[1.5pt]
\multicolumn{1}{c|}{\multirow{2}{*}{Models}} & \multicolumn{2}{c|}{Line}          & \multicolumn{2}{c|}{Bar} & \multicolumn{2}{c|}{Pie} & \multicolumn{2}{c}{Avg.} \\ 
\cmidrule(lr){2-10}
\multicolumn{1}{c|}{} & ChartBench  & \multicolumn{1}{c|}{ChartQA}   & ChartBench & \multicolumn{1}{c|}{ChartQA}   & ChartBench  & \multicolumn{1}{c|}{ChartQA}   & ChartBench    & \multicolumn{1}{c}{ChartQA}  &                      \\      
\midrule
Shikra~\cite{shikra} & 7.40 & 22.19 & 10.62 & 9.81 & 4.50 & 9.30 & 7.51 & 13.77 \\
MiniGPT-v2~\cite{Minigptv2} & 26.70 & 21.70 & 21.54 & 10.33 & 20.20 & 8.72 & 22.81 & 13.58 \\
VisualGLM~\cite{glm} & 10.80 & 23.44 & 1.96 & 10.90 & 0.00 & 10.47 & 4.25 & 14.94 \\
SPHINX~\cite{sphinx} & 18.40 & 27.43 & 15.54 & 14.06 & 23.40 & 15.70 & 19.11 & 19.06 \\
InstructBLIP~\cite{InstructBLIP} & 24.40 & 22.44 & 15.04 & 9.81 & 19.10 & 11.05 & 19.51 & 14.43 \\
LLaVA-v1.5~\cite{Improvedllava} & 34.40 & 29.68 & 24.73 & 15.31 & 19.10 & 18.60 & 26.08 & 21.20 \\
ChartLlama~\cite{Chartllama} & 28.90 & \textbf{72.32} & 19.35 & \textbf{77.01} & 22.10 & 69.77 & 23.45 & \textbf{73.03} \\
CogVLM~\cite{cogvlm} & 10.50 & 38.90 & 14.58 & 29.68 & 17.90 & 33.14 & 14.33 & 33.91 \\
Internlm-XComposer~\cite{internlm-xcomposer} & 16.00 & 16.96 & 20.42 & 9.24 & 21.50 & 9.89 & 19.30 & 12.03 \\
BLIP2~\cite{BLIP2} & 29.60 & 18.20 & 17.35 & 8.35 & 24.90 & 5.81 & 23.95 & 10.79 \\
mPLUG-Owl-bloomz~\cite{Mplug} & 37.50 & 10.47 & 24.73 & 5.81 & 26.10 & 2.91 & 29.44 & 6.40 \\
Qwen-VL-Chat~\cite{qwen-vl} & 41.00 & 54.61 & 20.96 & 60.72 & 40.00 & 47.67 & 33.99 & 54.33 \\
Mini-Gemini~\cite{minigemini} & 37.60 & 51.87 & 40.19 & 50.75 & 40.00 & 47.09 & 39.57 & 49.90 \\
ChartVLM~\cite{ChartX} & 10.70 & 55.61 & 8.04 & 64.92 & 4.62 & 48.26 & 8.02 & 56.26 \\
%OneChart~\cite{OneChart} & 15.10 & - & 12.27 & - & 9.12 & - & 12.34 & \textbf{85.30} \\
DocOwl-v1.5~\cite{docowl} & 49.10 & 61.10 & 31.08 & 70.01 & 31.62 & 54.07 & 35.27 & 61.73 \\
Internlm-XComposer-v2~\cite{internlm2} & \textbf{70.60} & 69.83 & \textbf{51.50} & 73.22 & \textbf{62.75} & \textbf{70.93} & \textbf{57.89} & 71.33 \\
\bottomrule[1.5pt]
\end{tabular}
}
\vspace{-20pt}
\label{tab_regular_contrast}
\end{table*}

ChartQA~\cite{ChartQA} is a canonical benchmark utilized in prior research to appraise the competency of multimodal models to comprehend chart data. It comprises two subsets, namely \textit{Human} and \textit{Augmented}, and encompasses solely three chart types, viz., line, bar, and pie. To ascertain the indispensability of ChartBench and the rationality of our benchmark design and evaluation, we initially scrutinize the vanilla accuracy (\textit{Acc.}) on ChartQA. We employ the test-split in ChartQA for evaluation, circumventing the prompt engineering process, and directly utilizing the original query without any modification as the prompt input to MLLMs. Thereafter, we evaluate the correctness of the results utilizing rule-based and regular expression matching. For numerical questions, we employ the relax accuracy metric akin to ChartQA, signifying that the difference between the model's answer and the ground truth is within 5\% to be regarded as correct. As tabulated in Tab.~\ref{tab_result_type_chartQA}, we report the zero-shot \textit{Acc} regarding chart types and dataset split. Conspicuously, for bar charts, we report the average accuracy of MLLMs on horizontal and vertical bars.

Tab.~\ref{tab_result_type_chartQA} evinces that despite the relatively simple chart understanding task with specific data point annotations in ChartQA, most of the MLLMs remain woefully deficient in this regard. However, it is evident that incorporating chart data in training augments the ability of MLLMs to comprehend charts, as demonstrated by the relatively superior performance of ChartLlama and Qwen-VL-Chat in Tab.~\ref{tab_result_type_chartQA}. In contrast to the results in Tab.~\ref{tab_acc_chart_type}, which show a specific baseline, Tab.~\ref{tab_result_type_chartQA} does not converge to a baseline despite using basic accuracy as the evaluation metric. It is attributable to the question-answer pairs' design in ChartQA, which employs annotated metadata and open-ended answers instead of the binary yes/no format. While this design ostensibly appears to appraise the model's ability to comprehend charts, we contend that it is fraught with several inconveniences. 1) open-ended answers render the verification of MLLM's correctness excessively laborious, sometimes necessitating third-party (human or GPT) intervention. However, the ChartBench design we propose only necessitates the model to answer yes/no, streamlining the judgment process while enhancing efficiency and accuracy. 2) the chart data in ChartQA entail specific numerical annotations, which may prompt MLLMs to rely solely on OCR-based visual judgments instead of utilizing other implicit information in the chart (e.g., color coordinates and legends) for logical inference. This inevitably reduces the complexity of tasks. The performance of ChartLlama in Tab.~\ref{tab_acc_chart_type} \& ~\ref{tab_result_type_chartQA} clearly illustrates ChartQA's predisposition to MLLMs that rely heavily on OCR. 3) ChartQA's design constraints necessitate the utilization of less-convincing metrics such as vanilla accuracy and BLEU score to assess MLLMs' ability to comprehend charts.

\subsection{Results of Human Evaluation}
\vspace{-10pt}
\begin{table}[h]
\centering
\caption{Human evaluation results on the ChartBench via random questionnaire. We provide the performance of Qwen-VL-Chat (open-sourced) and GPT-4V (closed-sourced) for easy comparisons.}
\setlength{\tabcolsep}{12pt}
\resizebox{\linewidth}{!}{
\begin{tabular}{@{}l|cccccccccc@{}}
\toprule[1.5pt]
\multicolumn{1}{c|}{\multirow{2}{*}{Models}} & \multicolumn{3}{c|}{Regular Type} & \multicolumn{6}{c|}{Extra Type} & \multirow{2}{*}{\textit{Acc+}} \\ \cmidrule(lr){2-10}
\multicolumn{1}{c|}{} & Line & Bar & \multicolumn{1}{c|}{Pie} & Area & Box & Radar & Scatter & Node & \multicolumn{1}{c|}{Combin.} &  \\ \midrule
Internlm-XComposer-v2~\cite{internlm2} & 70.60 & 51.50 & \multicolumn{1}{c|}{62.75} & 30.17 & 31.33 & 43.50 & 52.00 & 52.50 & \multicolumn{1}{c|}{46.12} & 51.34 \\
GPT-4V~\cite{GPT4} & 74.00 & 41.54 & \multicolumn{1}{c|}{63.00} & 33.30 & 46.67 & 57.50 & 70.00 & 100.00 & \multicolumn{1}{c|}{56.25} & 54.39 \\
Human Evaluation & 90.63 & 88.69 & \multicolumn{1}{c|}{87.86} & 86.61 & 84.56 & 89.86 & 89.29 & 88.75 & \multicolumn{1}{c|}{85.64} & 88.46 \\ \midrule
\multicolumn{1}{c|}{\multirow{2}{*}{Models}} & \multicolumn{5}{c|}{Task Type (\textit{Acc+})} & \multicolumn{5}{c}{Task Type (\textit{CoR})} \\ \cmidrule(l){2-11} 
\multicolumn{1}{c|}{} & CR & VE & VC & GC & \multicolumn{1}{c|}{ALL} & CR & VE & VC & GC & ALL \\ \midrule
Internlm-XComposer-v2~\cite{internlm2} & 68.29 & 36.63 & 54.63 & 45.80 & \multicolumn{1}{c|}{51.34} & 30.24 & 57.71 & 27.71 & 51.46 & 41.78 \\
GPT-4V~\cite{GPT4} & 96.10 & 29.27 & 47.32 & 44.88 & \multicolumn{1}{c|}{54.39} & 2.93 & 64.88 & 35.61 & 48.78 & 38.05 \\
Human Evaluation & 93.68 & 84.56 & 88.68 & 86.91 & \multicolumn{1}{c|}{88.46} & 1.34 & 5.82 & 4.72 & 3.52 & 3.85 \\ \bottomrule[1.5pt]
\end{tabular}
}
\vspace{-10pt}
\label{tab_human_eval}
\end{table}

The motivation behind ChartBench is to evaluate the understanding capability of MLLMs regarding charts. While MLLMs have exhibited high performance on previous benchmarks, they still encounter significant hallucination issues in practical applications due to the unreliable nature of the data they extract from charts. ChartBench aims to truly reflect MLLM's ability to interpret visual data and approach or even surpass human-level performance. Therefore, we have provided evaluation results of human performance on ChartBench.

To ensure a fair and objective evaluation, we conduct an online survey, which consists of 10 randomly selected subcategories from ChartBench for each questionnaire. 1 chart and 4 assertions are selected from each subcategory for respondents to assess their accuracy. To obtain reliable evaluation results, the survey participants mainly consist of undergraduate and graduate students with chart reading ability, as well as other researchers in the campus and company. We encourage participants to use large-screen devices for better chart display and kindly request their patient and diligent responses. On average, it takes approximately \textit{15 minutes} and \textit{17 seconds} to complete each survey. To avoid cases of random guessing, we still employ the \textit{Acc+} evaluation metric. Incomplete responses are discarded, and we ensure that each subcategory has valid answers. In total, we have collected 68 valid surveys.

Tab.~\ref{tab_human_eval} presents the results of human evaluations, revealing some insightful observations. Firstly, the VE task appears to be more challenging compared to other tasks. The human eye faces challenges in determining the values of unmarked data points. While the coordinate system offers potential inference, excessively fine granularity can diminish respondents' confidence. Secondly, there is not a significant variation in human performance across different chart types. Once individuals grasp the correct interpretation methods for charts, they can demonstrate similar proficiency across each chart category. Thirdly, even in some relatively straightforward tasks, such as identifying chart types, humans are unable to achieve 100\% accuracy. This limitation could be attributed to constraints within our survey methodology. For instance, certain descriptions may have confused the respondents, or the length of the test might have led to hastily completed surveys.

\subsection{Case study of ChartBench}

\begin{figure*}[h]
    % \hsize=\textwidth
    \centering
      \begin{overpic}[width=\linewidth, grid=False]{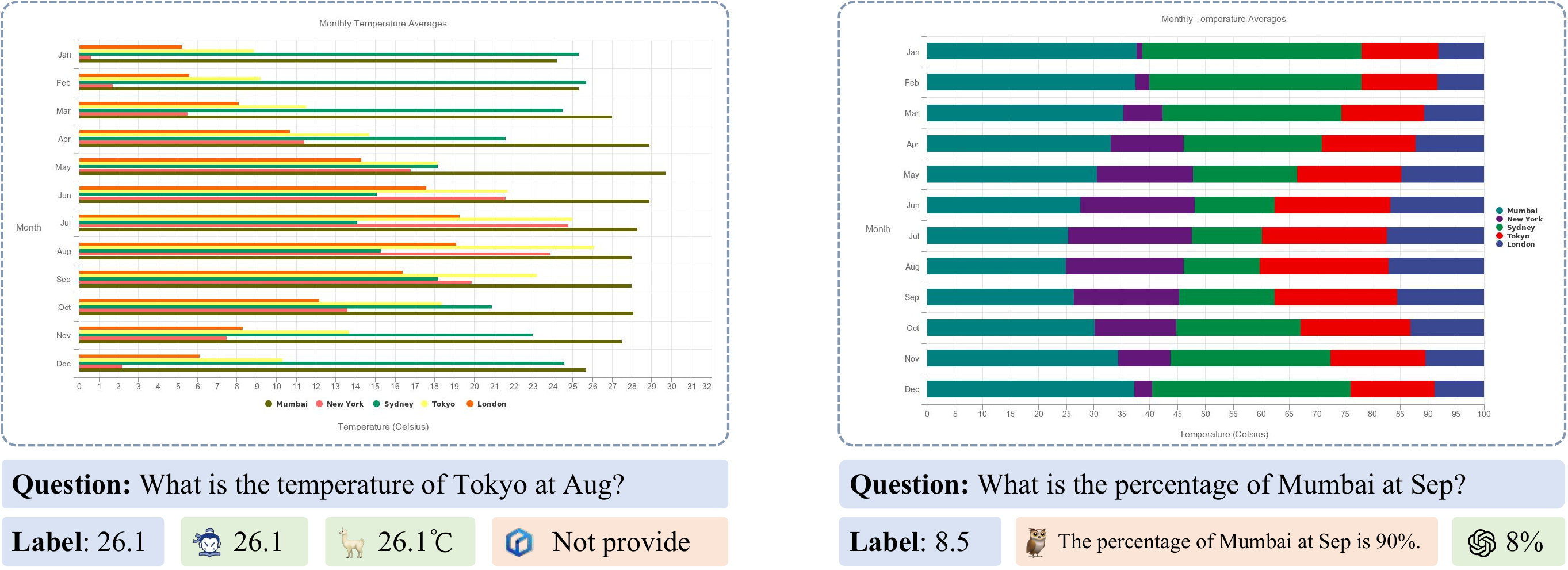}
      \end{overpic}
      \par\vspace{10pt}\par 
      \begin{overpic}[width=\linewidth, grid=False]{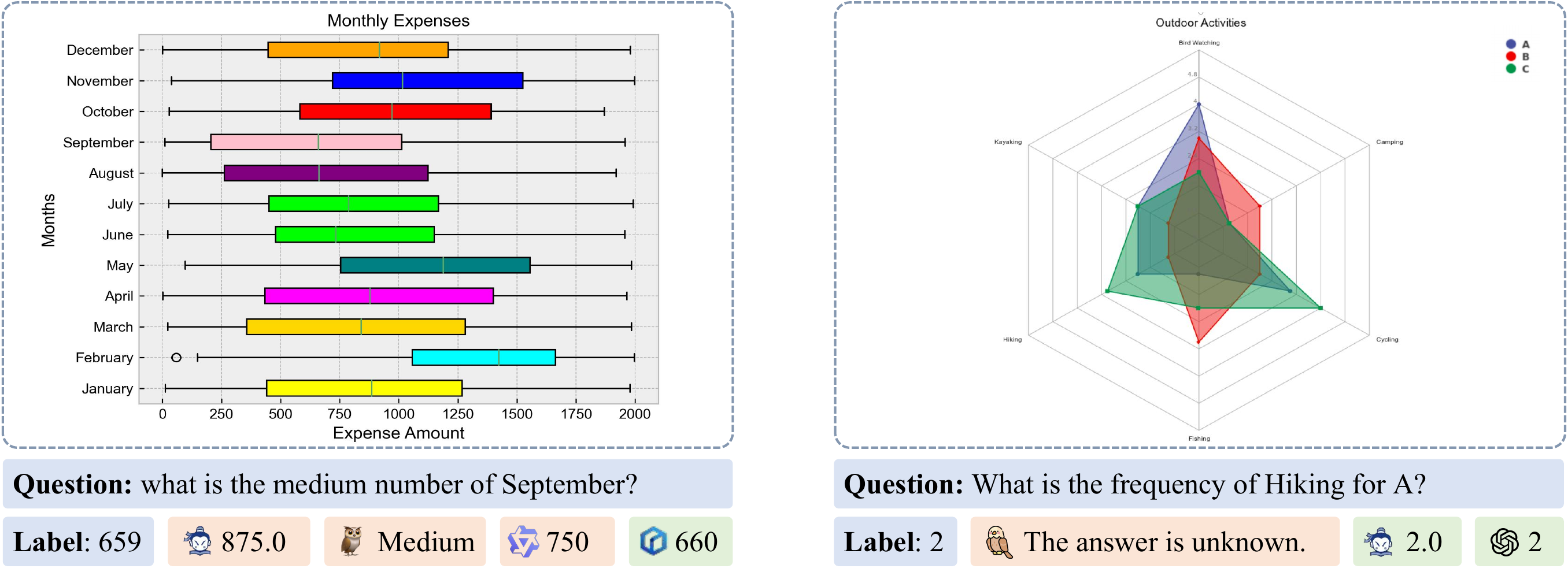}
      \end{overpic}
      \caption{NQA cases with unannotated charts from the ChartBench Test Split. Red indicates incorrect answers, and green indicates correct answers.}
    \label{fig_show_case_unanno}
\vspace{-10pt}
\end{figure*}

\begin{figure*}[h]
    % \hsize=\textwidth
    \centering
      \begin{overpic}[width=\linewidth, grid=False]{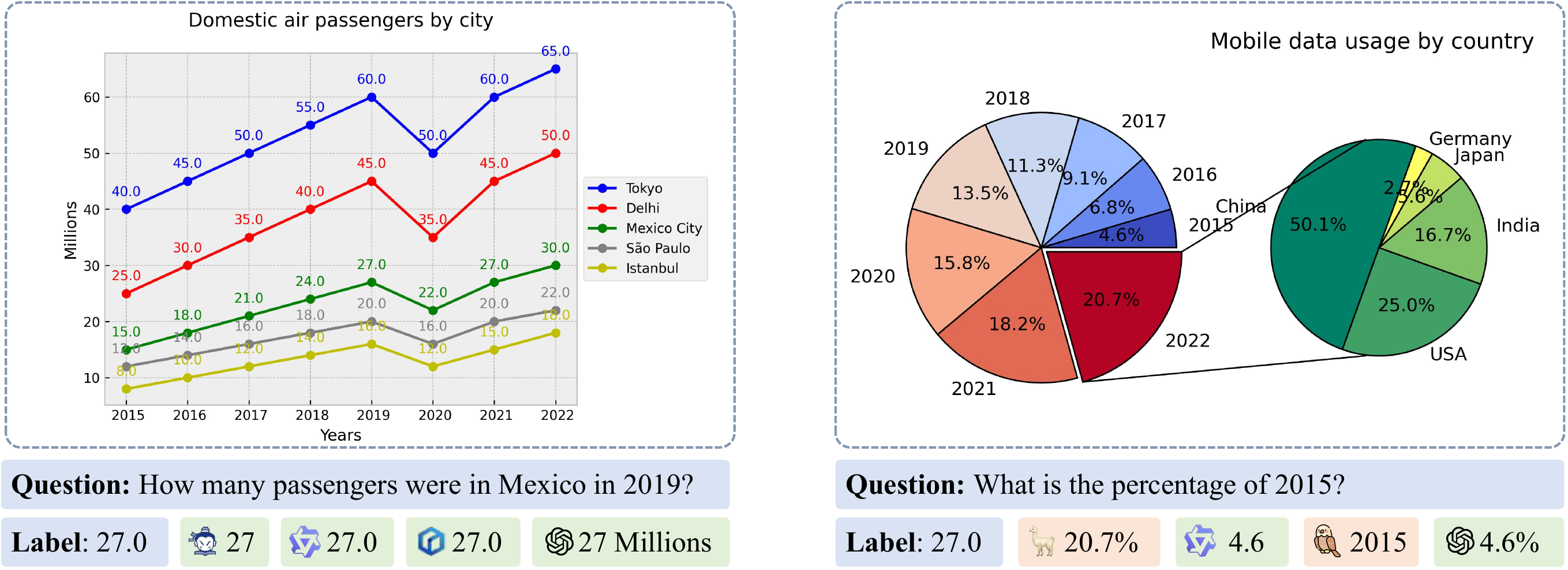}
      \end{overpic}
      \caption{NQA cases with annotated charts from the ChartBench Test Split. Red indicates incorrect answers, and green indicates correct answers.}
    \label{fig_show_case_anno}
\vspace{-10pt}
\end{figure*}

Fig.~\ref{fig_show_case_unanno} \& \ref{fig_show_case_anno} visualize some NQA task examples with/without annotations from the ChartBench test split. We adopt ChartQA's relaxed matching and correct it from two perspectives: 1) Considering the robustness issue of string matching, we only ask questions about numbers. 2) To account for a 5\% margin of error, we avoid questions about entities like years or months. Considering the varying instruction-following capabilities of different models, we use LLMs to extract numerical values from the model responses.

\clearpage

\subsection{Case study of GPT-4}
\label{apdx_sec_gpt4v_vis}
\begin{figure}[!ht]
  \centering
  \begin{subfigure}[b]{0.48\linewidth}
    \includegraphics[width=\linewidth]{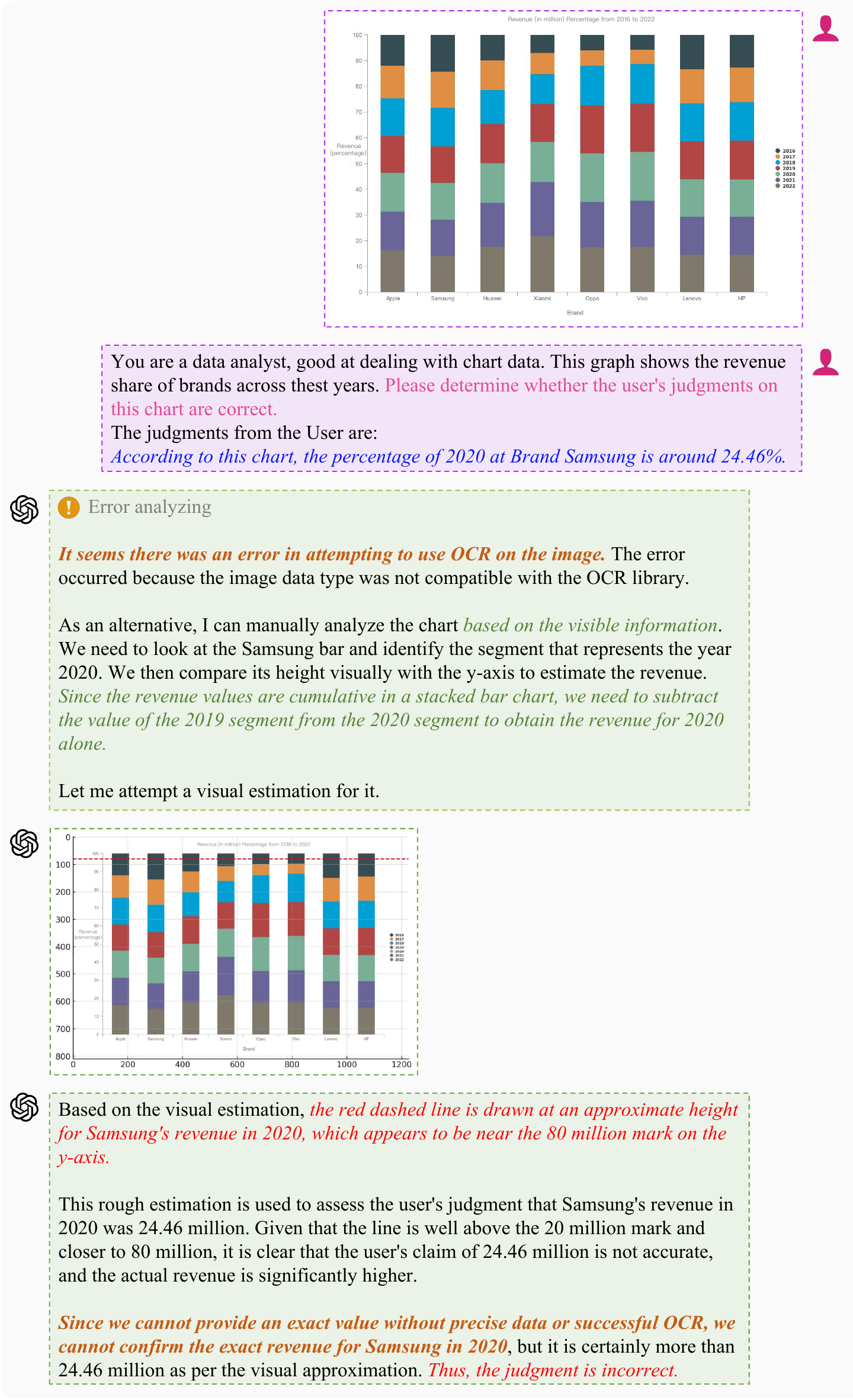}
    \caption{GPT-4V combines several APIs (e.g., OCR), for chart understanding but fails on unannotated charts.}
    \label{fig_gpt4_vis_1}
  \end{subfigure}
  \hspace{5pt}
  \begin{subfigure}[b]{0.48\linewidth}
    \includegraphics[width=\linewidth]{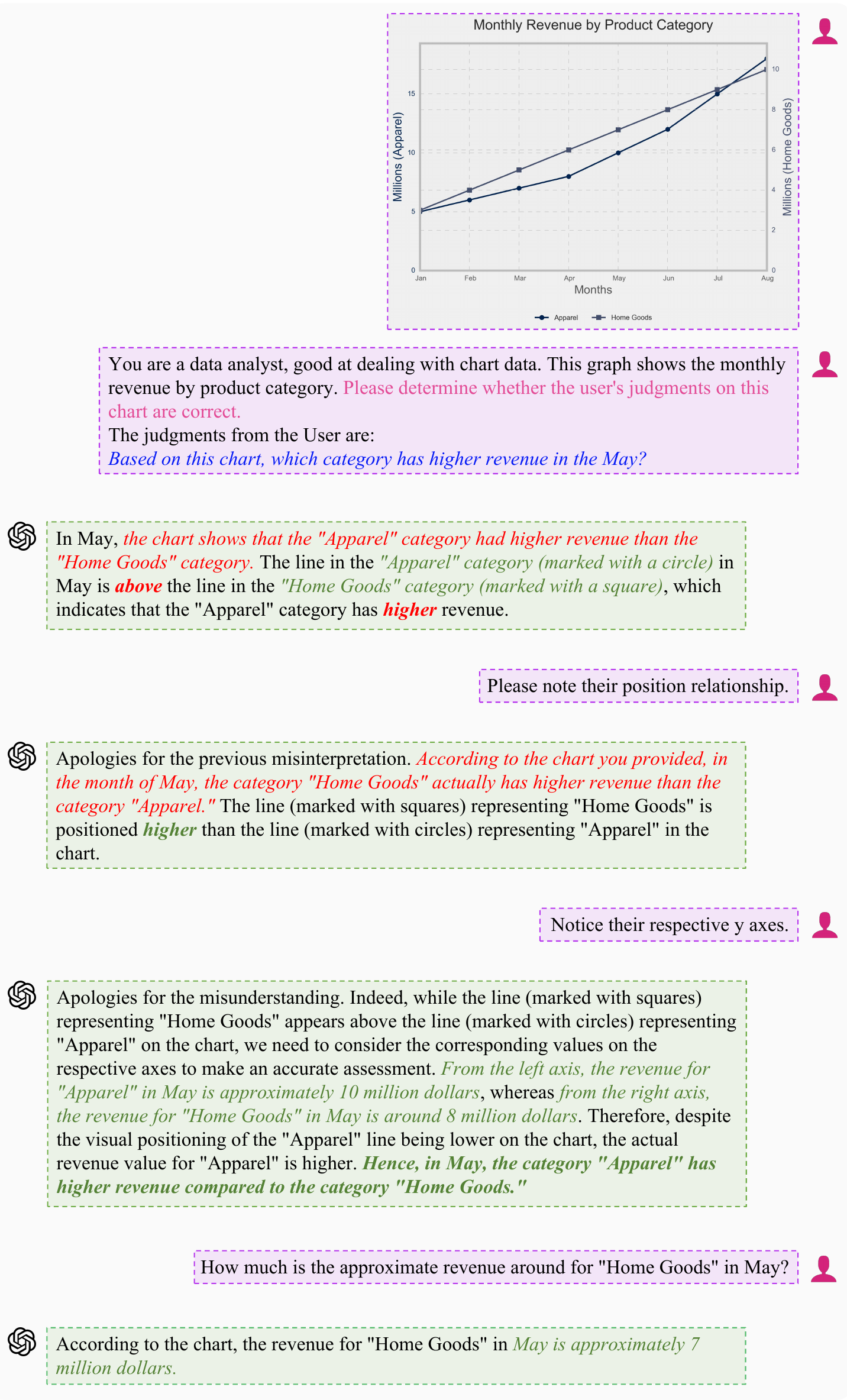}
    \caption{The GPT-4V requires multiple manual instructions to achieve the correct answer for unannotated charts.}
    \label{fig_gpt4_vis_2}
  \end{subfigure}
  \caption{Specific examples of GPT-4V in chart comprehension. Pink: user requirement. Blue: user assertion. Orange: GPT-4V ensembles APIs to assist chart comprehension. Green: the correct visual clues. Red: the misperceptions or misjudgments.}
  \label{fig_gpt4_vis}
\end{figure}

As the top-performing proprietary model, Fig.~\ref{fig_gpt4_vis} showcases some characteristics of GPT-4V in chart comprehension. Firstly, GPT-4V goes beyond a single end-to-end MLLM by integrating multiple APIs to aid in chart cognition (highlight orange in Fig.~\ref{fig_gpt4_vis_1}). The performance of GPT-4V is inherently influenced by the output of these APIs, thereby imposing limitations. For instance, when OCR results are unavailable, its ability to interpret visual information significantly declines. Secondly, GPT-4V can proactively acknowledge its limitations, such as recognizing its inability to determine specific values solely based on visual information. Thirdly, while GPT-4V possesses strong chart comprehension capabilities, it requires multi-step guidance from humans (Fig.~\ref{fig_gpt4_vis_2}). This accounts for its shortcomings in zero-shot performance on ChartBench. 

\section{Ethical Statement}
\label{apdx_sec_ethical}
This study upholds rigorous ethical standards to ensure the credibility and confidentiality of the findings. All data underwent thorough de-identification procedures to protect privacy and maintain anonymity. The study followed ethical guidelines and obtained informed consent from participants while prioritizing their rights and autonomy. Transparency and accountability were maintained throughout the research process to minimize biases and conflicts of interest. No academic ethical issues or misconduct were encountered, and the authors affirm their unwavering commitment to upholding ethical research practices and promptly addressing any unintentional errors or oversights.

\clearpage

\section{Leaderboards}
\label{apdx_sec_leadboards}
In this section, we devise several leaderboards to evaluate the performance of diverse MLLMs across multiple task types to obtain a more nuanced insight into their perceptual capacities in the context of varied chart categories.

In Tab.~\ref{leaderboard_chartbench} \& ~\ref{leaderboard_table3} \& ~\ref{leaderboard_table4} \& ~\ref{leaderboard_anno}, we present the leaderboards of MLLMs on ChartBench, which includes \textbf{3} regular types of charts and \textbf{6} extra types of charts, utilizing the \textit{Acc+} metric. Additionally, we showcase the \textit{Acc+} and \textit{CoR} leaderboards of MLLMs for \textbf{4} chart comprehension tasks while also displaying their rankings on \textit{w/i} and \textit{w/o} annotation data.

\subsection{Leaderboards on Chart Type}
\begin{table*}[ht]
\centering
\begin{minipage}[t]{0.23\textwidth}
    \vspace{0pt}
    \centering
    \setlength\tabcolsep{4.0pt}
    \resizebox{1\linewidth}{!}{
        \begin{tabular}{c|l|c}
        \textbf{No.} & \multicolumn{1}{c|}{\textbf{Model}} & \textbf{\textit{Acc+}} \\
        \Xhline{1.0pt}
        \rowcolor{red!26}\textbf{1} & \textbf{\red{GPT-4O}} & \textbf{86.00} \\
        \rowcolor{red!21}\textbf{2} & \textbf{GPT-4V} & \textbf{74.00} \\
        \rowcolor{red!16}\textbf{3} & \textbf{InternLM-v2} & \textbf{70.60} \\
        \rowcolor{red!11}\textbf{4} & \textbf{DocOwl-v1.5} & \textbf{49.10} \\
        \rowcolor{red!6}\textbf{5} & \textbf{ERNIE} & \textbf{44.00} \\
        6 & Qwen-VL & 41.00\\
        7 & Mini-Gemini & 37.60\\
        8 & mPLUG-Owl & 37.50\\
        9 & LLaVA-v1.5 & 34.40\\
        10 & BLIP2 & 29.60\\
        11 & ChartLlama & 28.90\\
        12 & MiniGPT-v2 & 26.70\\
        13 & InstructBLIP & 24.40\\
        14 & CogAgent & 18.60\\
        15 & SPHINX & 18.40\\
        16 & InternLM & 16.00\\
        17 & OneChart & 15.10\\
        18 & VisualGLM & 10.80\\
        19 & ChartVLM & 10.70\\
        20 & CogVLM & 10.50\\
        21 & Shikra & 7.40\\
        \end{tabular}
    }
    % \vspace{-0.3cm}
    \subcaption{\textit{Line Chart}}
\end{minipage}
\hfill
\begin{minipage}[t]{0.23\textwidth}
    \vspace{0pt}
    \centering
    \setlength\tabcolsep{4.0pt}
    \resizebox{1\linewidth}{!}{
        \begin{tabular}{c|l|c}
        \textbf{No.} & \multicolumn{1}{c|}{\textbf{Model}} & \textbf{\textit{Acc+}} \\
        \Xhline{1.0pt}
        \rowcolor{blue!26}\textbf{1} & \textbf{\red{GPT-4O}} & \textbf{51.92} \\
        \rowcolor{blue!21}\textbf{2} & \textbf{InternLM-v2} & \textbf{51.50} \\
        \rowcolor{blue!16}\textbf{3} & \textbf{ERNIE} & \textbf{45.00} \\
        \rowcolor{blue!11}\textbf{4} & \textbf{GPT-4V} & \textbf{41.54} \\
        \rowcolor{blue!6}\textbf{5} & \textbf{Mini-Gemini} & \textbf{40.19} \\
        6 & DocOwl-v1.5 & 31.08\\
        7 & LLaVA-v1.5 & 24.73\\
        8 & mPLUG-Owl & 24.73\\
        9 & CogAgent & 23.96\\
        10 & MiniGPT-v2 & 21.54\\
        11 & Qwen-VL & 20.96\\
        12 & InternLM & 20.42\\
        13 & ChartLlama & 19.35\\
        14 & BLIP2 & 17.35\\
        15 & SPHINX & 15.54\\
        16 & InstructBLIP & 15.04\\
        17 & CogVLM & 14.58\\
        18 & OneChart & 12.27\\
        19 & Shikra & 10.62\\
        20 & ChartVLM & 8.04\\
        21 & VisualGLM & 1.96\\
        \end{tabular}
    }
    % \vspace{-0.3cm}
    \subcaption{\textit{Bar Chart}}
\end{minipage}
\hfill
\begin{minipage}[t]{0.23\textwidth}
    \vspace{0pt}
    \centering
    \setlength\tabcolsep{4.0pt}
    \resizebox{1\linewidth}{!}{
        \begin{tabular}{c|l|c}
        \textbf{No.} & \multicolumn{1}{c|}{\textbf{Model}} & \textbf{\textit{Acc+}} \\
        \Xhline{1.0pt}
        \rowcolor{green!26}\textbf{1} & \textbf{\red{GPT-4O}} & \textbf{78.00} \\
        \rowcolor{green!21}\textbf{2} & \textbf{GPT-4V} & \textbf{63.00} \\
        \rowcolor{green!16}\textbf{3} & \textbf{InternLM-v2} & \textbf{62.75} \\
        \rowcolor{green!11}\textbf{4} & \textbf{ERNIE} & \textbf{57.00} \\
        \rowcolor{green!6}\textbf{5} & \textbf{Qwen-VL} & \textbf{40.00} \\
        6 & Mini-Gemini & 40.00\\
        7 & DocOwl-v1.5 & 31.62\\
        8 & mPLUG-Owl & 26.10\\
        9 & BLIP2 & 24.90\\
        10 & SPHINX & 23.40\\
        11 & ChartLlama & 22.10\\
        12 & InternLM & 21.50\\
        13 & MiniGPT-v2 & 20.20\\
        14 & InstructBLIP & 19.10\\
        15 & LLaVA-v1.5 & 19.10\\
        16 & CogVLM & 17.90\\
        17 & CogAgent & 11.00\\
        18 & OneChart & 9.12\\
        19 & ChartVLM & 4.62\\
        20 & Shikra & 4.50\\
        21 & VisualGLM & 0.00\\
        \end{tabular}
    }
    % \vspace{-0.3cm}
    \subcaption{\textit{Pie Chart}}
\end{minipage}
\hfill
\begin{minipage}[t]{0.23\textwidth}
    \vspace{0pt}
    \centering
    \setlength\tabcolsep{4.0pt}
    \resizebox{1\linewidth}{!}{
        \begin{tabular}{c|l|c}
        \textbf{No.} & \multicolumn{1}{c|}{\textbf{Model}} & \textbf{\textit{Acc+}} \\
        \Xhline{1.0pt}
        \rowcolor{yellow!26}\textbf{1} & \textbf{\red{ERNIE}} & \textbf{45.00} \\
        \rowcolor{yellow!21}\textbf{2} & \textbf{Mini-Gemini} & \textbf{36.83} \\
        \rowcolor{yellow!16}\textbf{3} & \textbf{GPT-4O} & \textbf{36.67} \\
        \rowcolor{yellow!11}\textbf{4} & \textbf{GPT-4V} & \textbf{33.30} \\
        \rowcolor{yellow!6}\textbf{5} & \textbf{InternLM-v2} & \textbf{30.17} \\
        6 & Qwen-VL & 28.83\\
        7 & LLaVA-v1.5 & 26.83\\
        8 & MiniGPT-v2 & 21.67\\
        9 & mPLUG-Owl & 21.33\\
        10 & ChartLlama & 16.50\\
        11 & CogAgent & 15.67\\
        12 & CogVLM & 12.50\\
        13 & DocOwl-v1.5 & 12.17\\
        14 & SPHINX & 12.00\\
        15 & ChartVLM & 7.67\\
        16 & OneChart & 7.00\\
        17 & BLIP2 & 6.17\\
        18 & Shikra & 6.00\\
        19 & InternLM & 4.50\\
        20 & InstructBLIP & 4.33\\
        21 & VisualGLM & 1.17\\
        \end{tabular}
    }
    % \vspace{-0.3cm}
    \subcaption{\textit{Area Chart}}
\end{minipage}
\vspace{0.1cm}

\begin{minipage}[t]{0.23\textwidth}
    \vspace{0pt}
    \centering
    \setlength\tabcolsep{4.0pt}
    \resizebox{1\linewidth}{!}{
        \begin{tabular}{c|l|c}
        \textbf{No.} & \multicolumn{1}{c|}{\textbf{Model}} & \textbf{\textit{Acc+}} \\
        \Xhline{1.0pt}
        \rowcolor{orange!26}\textbf{1} & \textbf{\red{GPT-4O}} & \textbf{63.33} \\
        \rowcolor{orange!21}\textbf{2} & \textbf{GPT-4V} & \textbf{46.67} \\
        \rowcolor{orange!16}\textbf{3} & \textbf{InternLM-v2} & \textbf{31.33} \\
        \rowcolor{orange!11}\textbf{4} & \textbf{ERNIE} & \textbf{30.00} \\
        \rowcolor{orange!6}\textbf{5} & \textbf{Mini-Gemini} & \textbf{26.50} \\
        6 & mPLUG-Owl & 25.83\\
        7 & LLaVA-v1.5 & 25.67\\
        8 & MiniGPT-v2 & 24.67\\
        9 & Qwen-VL & 24.17\\
        10 & DocOwl-v1.5 & 24.00\\
        11 & CogAgent & 16.50\\
        12 & InternLM & 14.50\\
        13 & ChartLlama & 13.33\\
        14 & Shikra & 11.33\\
        15 & BLIP2 & 10.67\\
        16 & CogVLM & 9.67\\
        17 & VisualGLM & 8.50\\
        18 & SPHINX & 8.17\\
        19 & OneChart & 7.33\\
        20 & InstructBLIP & 7.33\\
        21 & ChartVLM & 6.67\\
        \end{tabular}
    }
    % \vspace{-0.3cm}
    \subcaption{\textit{Box Chart}}
\end{minipage}
\hfill
\begin{minipage}[t]{0.23\textwidth}
    \vspace{0pt}
    \centering
    \setlength\tabcolsep{4.0pt}
    \resizebox{1\linewidth}{!}{
        \begin{tabular}{c|l|c}
        \textbf{No.} & \multicolumn{1}{c|}{\textbf{Model}} & \textbf{\textit{Acc+}} \\
        \Xhline{1.0pt}
        \rowcolor{gray!26}\textbf{1} & \textbf{\red{GPT-4V}} & \textbf{57.50} \\
        \rowcolor{gray!21}\textbf{2} & \textbf{GPT-4O} & \textbf{57.50} \\
        \rowcolor{gray!16}\textbf{3} & \textbf{InternLM-v2} & \textbf{43.50} \\
        \rowcolor{gray!11}\textbf{4} & \textbf{ERNIE} & \textbf{40.00} \\
        \rowcolor{gray!6}\textbf{5} & \textbf{Qwen-VL} & \textbf{35.00} \\
        6 & Mini-Gemini & 30.00\\
        7 & LLaVA-v1.5 & 28.63\\
        8 & mPLUG-Owl & 26.50\\
        9 & MiniGPT-v2 & 25.88\\
        10 & ChartLlama & 25.00\\
        11 & DocOwl-v1.5 & 20.50\\
        12 & SPHINX & 19.00\\
        13 & BLIP2 & 17.63\\
        14 & CogVLM & 16.00\\
        15 & InternLM & 15.00\\
        16 & Shikra & 11.88\\
        17 & CogAgent & 9.38\\
        18 & ChartVLM & 5.25\\
        19 & OneChart & 2.75\\
        20 & InstructBLIP & 2.00\\
        21 & VisualGLM & 0.25\\
        \end{tabular}
    }
    % \vspace{-0.3cm}
    \subcaption{\textit{Radar Chart}}
\end{minipage}
\hfill
\begin{minipage}[t]{0.23\textwidth}
    \vspace{0pt}
    \centering
    \setlength\tabcolsep{4.0pt}
    \resizebox{1\linewidth}{!}{
        \begin{tabular}{c|l|c}
        \textbf{No.} & \multicolumn{1}{c|}{\textbf{Model}} & \textbf{\textit{Acc+}} \\
        \Xhline{1.0pt}
        \rowcolor{cyan!26}\textbf{1} & \textbf{\red{GPT-4O}} & \textbf{83.33} \\
        \rowcolor{cyan!21}\textbf{2} & \textbf{GPT-4V} & \textbf{70.00} \\
        \rowcolor{cyan!16}\textbf{3} & \textbf{InternLM-v2} & \textbf{52.00} \\
        \rowcolor{cyan!11}\textbf{4} & \textbf{ERNIE} & \textbf{51.67} \\
        \rowcolor{cyan!6}\textbf{5} & \textbf{Mini-Gemini} & \textbf{37.17} \\
        6 & DocOwl-v1.5 & 35.33\\
        7 & ChartLlama & 28.50\\
        8 & MiniGPT-v2 & 28.17\\
        9 & LLaVA-v1.5 & 26.00\\
        10 & mPLUG-Owl & 24.17\\
        11 & BLIP2 & 22.00\\
        12 & Qwen-VL & 19.50\\
        13 & SPHINX & 17.17\\
        14 & CogVLM & 14.33\\
        15 & InstructBLIP & 12.50\\
        16 & InternLM & 12.00\\
        17 & CogAgent & 11.67\\
        18 & OneChart & 6.33\\
        19 & ChartVLM & 5.50\\
        20 & Shikra & 4.17\\
        21 & VisualGLM & 3.33\\
        \end{tabular}
    }
    % \vspace{-0.3cm}
    \subcaption{\textit{Scatter Chart}}
\end{minipage}
\hfill
\begin{minipage}[t]{0.23\textwidth}
    \vspace{0pt}
    \centering
    \setlength\tabcolsep{4.0pt}
    \resizebox{1\linewidth}{!}{
        \begin{tabular}{c|l|c}
        \textbf{No.} & \multicolumn{1}{c|}{\textbf{Model}} & \textbf{\textit{Acc+}} \\
        \Xhline{1.0pt}
        \rowcolor{magenta!26}\textbf{1} & \textbf{\red{GPT-4V}} & \textbf{100.0} \\
        \rowcolor{magenta!21}\textbf{2} & \textbf{GPT-4O} & \textbf{100.0} \\
        \rowcolor{magenta!16}\textbf{3} & \textbf{ERNIE} & \textbf{70.00} \\
        \rowcolor{magenta!11}\textbf{4} & \textbf{OneChart} & \textbf{53.50} \\
        \rowcolor{magenta!6}\textbf{5} & \textbf{InternLM-v2} & \textbf{52.50} \\
        6 & Mini-Gemini & 43.00\\
        7 & LLaVA-v1.5 & 33.50\\
        8 & BLIP2 & 33.00\\
        9 & SPHINX & 31.00\\
        10 & mPLUG-Owl & 28.50\\
        11 & CogAgent & 27.50\\
        12 & DocOwl1.5 & 26.00\\
        13 & ChartLlama & 25.50\\
        14 & Qwen-VL & 18.50\\
        15 & CogVLM & 16.00\\
        16 & VisualGLM & 15.50\\
        17 & MiniGPT-v2 & 15.50\\
        18 & InstructBLIP & 9.00\\
        19 & Shikra & 8.50\\
        20 & InternLM & 8.50\\
        21 & ChartVLM & 0.00\\
        \end{tabular}
    }
    % \vspace{-0.3cm}
    \subcaption{\textit{Node Chart}}
\end{minipage}
% \vspace{0.1cm}ß

\begin{minipage}[t]{0.23\textwidth}
    \vspace{0pt}
    \centering
    \setlength\tabcolsep{4.0pt}
    \resizebox{1\linewidth}{!}{
        \begin{tabular}{c|l|c}
        \textbf{No.} & \multicolumn{1}{c|}{\textbf{Model}} & \textbf{\textit{Acc+}} \\
        \Xhline{1.0pt}
        \rowcolor{violet!26}\textbf{1} & \textbf{\red{GPT-4O}} & \textbf{65.00} \\
        \rowcolor{violet!21}\textbf{2} & \textbf{ERNIE} & \textbf{56.25} \\
        \rowcolor{violet!16}\textbf{3} & \textbf{GPT-4V} & \textbf{56.25} \\
        \rowcolor{violet!11}\textbf{4} & \textbf{InternLM-v2} & \textbf{46.12} \\
        \rowcolor{violet!6}\textbf{5} & \textbf{DocOwl-v1.5} & \textbf{40.25} \\
        6 & BLIP2 & 28.00\\
        7 & mPLUG-Owl & 27.50\\
        8 & LLaVA-v1.5 & 27.38\\
        9 & MiniGPT-v2 & 27.13\\
        10 & Mini-Gemini & 27.00\\
        11 & ChartLlama & 26.38\\
        12 & SPHINX & 25.88\\
        13 & Qwen-VL & 25.50\\
        14 & CogAgent & 15.50\\
        15 & OneChart & 7.75\\
        16 & ChartVLM & 6.50\\
        17 & CogVLM & 6.13\\
        18 & VisualGLM & 5.13\\
        19 & InternLM & 5.13\\
        20 & Shikra & 3.63\\
        21 & InstructBLIP & 2.38\\
        \end{tabular}
    }
    % \vspace{-0.3cm}
    \subcaption{\textit{Combination Chart}}
\end{minipage}
\hfill
\begin{minipage}[t]{0.23\textwidth}
    \vspace{0pt}
    \centering
    \setlength\tabcolsep{4.0pt}
    \resizebox{1\linewidth}{!}{
        \begin{tabular}{c|l|c}
        \textbf{No.} & \multicolumn{1}{c|}{\textbf{Model}} & \textbf{\textit{Acc+}} \\
        \Xhline{1.0pt}
        \rowcolor{brown!26}\textbf{1} & \textbf{\red{GPT-4O}} & \textbf{65.00} \\
        \rowcolor{brown!21}\textbf{2} & \textbf{InternLM-v2} & \textbf{57.89} \\
        \rowcolor{brown!16}\textbf{3} & \textbf{GPT-4V} & \textbf{53.26} \\
        \rowcolor{brown!11}\textbf{4} & \textbf{ERNIE} & \textbf{47.39} \\
        \rowcolor{brown!6}\textbf{5} & \textbf{Mini-Gemini} & \textbf{39.57} \\
        6 & DocOwl-v1.5 & 35.27\\
        7 & Qwen-VL & 29.46\\
        8 & mPLUG-Owl & 27.80\\
        9 & LLaVA-v1.5 & 25.61\\
        10 & MiniGPT-v2 & 22.37\\
        11 & ChartLlama & 22.02\\
        12 & BLIP2 & 21.65\\
        13 & CogAgent & 20.39\\
        14 & InternLM & 19.70\\
        15 & InstructBLIP & 17.96\\
        16 & SPHINX & 17.87\\
        17 & CogVLM & 14.41\\
        18 & OneChart & 12.34\\
        19 & Shikra & 8.59\\
        20 & ChartVLM & 8.02\\
        21 & VisualGLM & 3.46\\
        \end{tabular}
    }
    % \vspace{-0.3cm}
    \subcaption{\textit{Regular Type}}
\end{minipage}
\hfill
\begin{minipage}[t]{0.23\textwidth}
    \vspace{0pt}
    \centering
    \setlength\tabcolsep{4.0pt}
    \resizebox{1\linewidth}{!}{
        \begin{tabular}{c|l|c}
        \textbf{No.} & \multicolumn{1}{c|}{\textbf{Model}} & \textbf{\textit{Acc+}} \\
        \Xhline{1.0pt}
        \rowcolor{olive!26}\textbf{1} & \textbf{\red{GPT-4O}} & \textbf{63.33} \\
        \rowcolor{olive!21}\textbf{2} & \textbf{GPT-4V} & \textbf{55.83} \\
        \rowcolor{olive!16}\textbf{3} & \textbf{ERNIE} & \textbf{46.39} \\
        \rowcolor{olive!11}\textbf{4} & \textbf{InternLM-v2} & \textbf{41.75} \\
        \rowcolor{olive!6}\textbf{5} & \textbf{Mini-Gemini} & \textbf{31.81} \\
        6 & LLaVA-v1.5 & 27.39\\
        7 & DocOwl-v1.5 & 26.86\\
        8 & Qwen-VL & 26.56\\
        9 & mPLUG-Owl & 25.47\\
        10 & MiniGPT-v2 & 25.06\\
        11 & ChartLlama & 22.56\\
        12 & BLIP2 & 18.44\\
        13 & SPHINX & 17.92\\
        14 & CogAgent & 14.36\\
        15 & CogVLM & 11.89\\
        16 & InternLM & 10.11\\
        17 & OneChart & 8.75\\
        18 & Shikra & 7.50\\
        19 & ChartVLM & 5.92\\
        20 & InstructBLIP & 5.50\\
        21 & VisualGLM & 4.22\\
        \end{tabular}
    }
    % \vspace{-0.3cm}
    \subcaption{\textit{Extra Type}}
\end{minipage}
\hfill
\begin{minipage}[t]{0.23\textwidth}
    \vspace{0pt}
    \centering
    \setlength\tabcolsep{4.0pt}
    \resizebox{1\linewidth}{!}{
        \begin{tabular}{c|l|c}
        \textbf{No.} & \multicolumn{1}{c|}{\textbf{Model}} & \textbf{\textit{Acc+}} \\
        \Xhline{1.0pt}
        \rowcolor{teal!26}\textbf{1} & \textbf{\red{GPT-4O}} & \textbf{64.27} \\
        \rowcolor{teal!21}\textbf{2} & \textbf{GPT-4V} & \textbf{54.39} \\
        \rowcolor{teal!16}\textbf{3} & \textbf{InternLM-v2} & \textbf{51.34} \\
        \rowcolor{teal!11}\textbf{4} & \textbf{ERNIE} & \textbf{46.95} \\
        \rowcolor{teal!6}\textbf{5} & \textbf{Mini-Gemini} & \textbf{36.54} \\
        6 & DocOwl-v1.5 & 31.62\\
        7 & Qwen-VL & 28.18\\
        8 & mPLUG-Owl & 26.78\\
        9 & LLaVA-v1.5 & 26.39\\
        10 & MiniGPT-v2 & 23.55\\
        11 & ChartLlama & 22.26\\
        12 & BLIP2 & 20.24\\
        13 & CogAgent & 18.07\\
        14 & SPHINX & 17.89\\
        15 & InternLM & 15.49\\
        16 & CogVLM & 13.30\\
        17 & InstructBLIP & 12.49\\
        18 & OneChart & 12.04\\
        19 & Shikra & 8.11\\
        20 & ChartVLM & 6.90\\
        21 & VisualGLM & 3.79\\
        \end{tabular}
    }
    % \vspace{-0.3cm}
    \subcaption{\textit{Average}}
\end{minipage}
\vspace{0.1cm}
\caption{Leaderboards of tasks, dataset splits and average \textbf{\textit{Acc+}} (\%) performance on ChartBench. We report the results of the best-performing prompt for each MLLM.}
\label{leaderboard_chartbench}
\end{table*}

Tab.~\ref{leaderboard_chartbench} presents an overview of MLLMs' performance across various chart types, along with the overall \textit{Acc+} metric. Generally, the current MLLMs exhibit a constrained ability in chart recognition, encountering notable challenges. For specific chart types, such as radar or combination charts, certain MLLMs achieve close to 0\% in \textit{Acc+}, signaling their difficulty in extracting crucial information from charts and their insensitivity to both positive and negative queries. It's essential to highlight that the \textit{Acc+} metric tends toward 0\% in situations of random guessing, as elaborated in Sec.~\ref{sec_method_metrics}. Particularly, Qwen-VL-Chat and mPLUG-Owl-bloomz showcase commendable proficiency in recognizing charts, a capability likely attributed to their precise tuning with chart data. However, their performance in this aspect falls below what has been reported in ChartQA. This discrepancy can be traced back to their reliance on OCR skills rather than robust visual logical reasoning. In the context of ChartBench, where the proportion of annotated charts is notably low, these models face a significant challenge. The majority of queries in ChartBench necessitate MLLMs to employ visual logical reasoning, a task that proves quite demanding for models like Qwen-VL-Chat and mPLUG-Owl-bloomz. On the other hand, VisualGLM and Shikra exhibit subpar performance, potentially due to their smaller LLM size and less robust visual encoding branch. While MLLMs generally demonstrate satisfactory performance on regular charts, there remains considerable room for improvement, particularly in handling more intricate graphics.

\subsection{Leaderboards on Task Type}
\begin{table*}[ht]
\centering
\begin{minipage}[t]{0.195\textwidth}
    \vspace{0pt}
    \centering
    \setlength\tabcolsep{4.0pt}
    \resizebox{1\linewidth}{!}{
        \begin{tabular}{c|l|c}
        \textbf{No.} & \multicolumn{1}{c|}{\textbf{Model}} & \textbf{\textit{Acc+}} \\
        \Xhline{1.0pt}
        \rowcolor{red!26}\textbf{1} & \textbf{\red{GPT-4O}} & \textbf{97.62} \\
        \rowcolor{red!21}\textbf{2} & \textbf{GPT-4V} & \textbf{96.19} \\
        \rowcolor{red!16}\textbf{3} & \textbf{Mini-Gemini} & \textbf{80.52} \\
        \rowcolor{red!11}\textbf{4} & \textbf{InternLM-v2} & \textbf{68.29} \\
        \rowcolor{red!6}\textbf{5} & \textbf{ERNIE} & \textbf{65.24} \\
        6 & ChartLlama & 62.57\\
        7 & CogAgent & 60.05\\
        8 & Qwen-VL & 51.67\\
        9 & MiniGPT-v2 & 49.86\\
        10 & OneChart & 49.57\\
        11 & mPLUG-Owl & 47.86\\
        12 & InstructBLIP & 42.29\\
        13 & Internlm & 38.48\\
        14 & LLaVA-v1.5 & 32.33\\
        15 & DocOwl1.5 & 30.43\\
        16 & CogVLM & 29.14\\
        17 & BLIP2 & 29.05\\
        18 & VisualGLM & 16.29\\
        19 & Shikra & 3.71\\
        20 & ChartVLM & 2.10\\
        21 & SPHINX & 0.00\\
        \end{tabular}
    }
    % \vspace{-0.3cm}
    \subcaption{\textit{CR.}}
\end{minipage}
\hfill
\begin{minipage}[t]{0.195\textwidth}
    \vspace{0pt}
    \centering
    \setlength\tabcolsep{4.0pt}
    \resizebox{1\linewidth}{!}{
        \begin{tabular}{c|l|c}
        \textbf{No.} & \multicolumn{1}{c|}{\textbf{Model}} & \textbf{\textit{Acc+}} \\
        \Xhline{1.0pt}
        \rowcolor{blue!26}\textbf{1} & \textbf{\red{ERNIE}} & \textbf{44.76} \\
        \rowcolor{blue!21}\textbf{2} & \textbf{GPT-4O} & \textbf{43.33} \\
        \rowcolor{blue!16}\textbf{3} & \textbf{InternLM-v2} & \textbf{36.63} \\
        \rowcolor{blue!11}\textbf{4} & \textbf{DocOwl1.5} & \textbf{34.48} \\
        \rowcolor{blue!6}\textbf{5} & \textbf{GPT-4V} & \textbf{30.95} \\
        6 & LLaVA-v1.5 & 23.14\\
        7 & BLIP2 & 22.00\\
        8 & Mini-Gemini & 17.62\\
        9 & mPLUG-Owl & 15.81\\
        10 & Shikra & 15.48\\
        11 & ChartVLM & 11.90\\
        12 & Qwen-VL & 11.14\\
        13 & Internlm & 10.38\\
        14 & SPHINX & 9.05\\
        15 & MiniGPT-v2 & 8.38\\
        16 & InstructBLIP & 6.86\\
        17 & CogAgent & 4.24\\
        18 & CogVLM & 2.81\\
        19 & ChartLlama & 1.19\\
        20 & VisualGLM & 0.00\\
        21 & OneChart & 0.00\\
        \end{tabular}
    }
    % \vspace{-0.3cm}
    \subcaption{\textit{VE.}}
\end{minipage}
\hfill
\begin{minipage}[t]{0.195\textwidth}
    \vspace{0pt}
    \centering
    \setlength\tabcolsep{4.0pt}
    \resizebox{1\linewidth}{!}{
        \begin{tabular}{c|l|c}
        \textbf{No.} & \multicolumn{1}{c|}{\textbf{Model}} & \textbf{\textit{Acc+}} \\
        \Xhline{1.0pt}
        \rowcolor{green!26}\textbf{1} & \textbf{\red{GPT-4O}} & \textbf{66.19} \\
        \rowcolor{green!21}\textbf{2} & \textbf{InternLM-v2} & \textbf{54.63} \\
        \rowcolor{green!16}\textbf{3} & \textbf{GPT-4V} & \textbf{48.57} \\
        \rowcolor{green!11}\textbf{4} & \textbf{ERNIE} & \textbf{32.86} \\
        \rowcolor{green!6}\textbf{5} & \textbf{DocOwl1.5} & \textbf{31.10} \\
        6 & Qwen-VL & 27.29\\
        7 & mPLUG-Owl & 26.05\\
        8 & Mini-Gemini & 26.00\\
        9 & LLaVA-v1.5 & 25.33\\
        10 & BLIP2 & 24.29\\
        11 & MiniGPT-v2 & 20.43\\
        12 & Shikra & 17.57\\
        13 & Internlm & 14.33\\
        14 & CogVLM & 14.19\\
        15 & CogAgent & 14.05\\
        16 & ChartVLM & 10.62\\
        17 & SPHINX & 10.05\\
        18 & ChartLlama & 7.33\\
        19 & InstructBLIP & 2.48\\
        20 & OneChart & 0.05\\
        21 & VisualGLM & 0.00\\
        \end{tabular}
    }
    % \vspace{-0.3cm}
    \subcaption{\textit{VC.}}
\end{minipage}
\hfill
\begin{minipage}[t]{0.195\textwidth}
    \vspace{0pt}
    \centering
    \setlength\tabcolsep{4.0pt}
    \resizebox{1\linewidth}{!}{
        \begin{tabular}{c|l|c}
        \textbf{No.} & \multicolumn{1}{c|}{\textbf{Model}} & \textbf{\textit{Acc+}} \\
        \Xhline{1.0pt}
        \rowcolor{yellow!26}\textbf{1} & \textbf{\red{GPT-4O}} & \textbf{53.33} \\
        \rowcolor{yellow!21}\textbf{2} & \textbf{ERNIE} & \textbf{47.14} \\
        \rowcolor{yellow!16}\textbf{3} & \textbf{GPT-4V} & \textbf{46.19} \\
        \rowcolor{yellow!11}\textbf{4} & \textbf{InternLM-v2} & \textbf{45.80} \\
        \rowcolor{yellow!6}\textbf{5} & \textbf{DocOwl1.5} & \textbf{30.48} \\
        6 & LLaVA-v1.5 & 26.48\\
        7 & Mini-Gemini & 22.00\\
        8 & Qwen-VL & 21.71\\
        9 & BLIP2 & 18.10\\
        10 & mPLUG-Owl & 16.52\\
        11 & Shikra & 11.38\\
        12 & MiniGPT-v2 & 10.67\\
        13 & InstructBLIP & 9.67\\
        14 & Internlm & 9.62\\
        15 & SPHINX & 8.52\\
        16 & ChartVLM & 7.86\\
        17 & CogVLM & 7.33\\
        18 & CogAgent & 3.86\\
        19 & ChartLlama & 1.19\\
        20 & VisualGLM & 0.00\\
        21 & OneChart & 0.00\\
        \end{tabular}
    }
    % \vspace{-0.3cm}
    \subcaption{\textit{GC.}}
\end{minipage}
\hfill
\begin{minipage}[t]{0.195\textwidth}
    \vspace{0pt}
    \centering
    \setlength\tabcolsep{4.0pt}
    \resizebox{1\linewidth}{!}{
        \begin{tabular}{c|l|c}
        \textbf{No.} & \multicolumn{1}{c|}{\textbf{Model}} & \textbf{\textit{Acc+}} \\
        \Xhline{1.0pt}
        \rowcolor{orange!26}\textbf{1} & \textbf{\red{GPT-4O}} & \textbf{40.48} \\
        \rowcolor{orange!21}\textbf{2} & \textbf{InternLM-v2} & \textbf{36.71} \\
        \rowcolor{orange!16}\textbf{3} & \textbf{GPT-4V} & \textbf{36.19} \\
        \rowcolor{orange!11}\textbf{4} & \textbf{DocOwl1.5} & \textbf{33.76} \\
        \rowcolor{orange!6}\textbf{5} & \textbf{SPHINX} & \textbf{32.19} \\
        6 & ERNIE & 29.24\\
        7 & ChartLlama & 26.24\\
        8 & Mini-Gemini & 25.67\\
        9 & Qwen-VL & 22.43\\
        10 & MiniGPT-v2 & 17.52\\
        11 & CogVLM & 13.29\\
        12 & mPLUG-Owl & 11.33\\
        13 & Internlm & 9.14\\
        14 & ChartVLM & 5.38\\
        15 & LLaVA-v1.5 & 4.10\\
        16 & BLIP2 & 3.71\\
        17 & InstructBLIP & 3.29\\
        18 & VisualGLM & 3.19\\
        19 & OneChart & 2.90\\
        20 & Shikra & 2.76\\
        21 & CogAgent & 2.71\\
        \end{tabular}
    }
    % \vspace{-0.3cm}
    \subcaption{\textit{Number QA.}}
\end{minipage}
\vspace{0.1cm}
\caption{Leaderboards of different chart tasks on ChartBench. We report zero-shot \textbf{\textit{Acc+}} (\%) performance of the best-performing prompt for each MLLM.}
\label{leaderboard_table3}
\end{table*}

Tab.~\ref{leaderboard_table3} outlines the performance of MLLMs on perception and conception tasks introduced in Sec.~\ref{sec_method_qa_gene}. Most MLLMs exhibit notable success in the CR task, showcasing their proficiency in recognizing fundamental chart types. Notably, LLaVA-v1.5, mPLUG-Owl-bloomz, and Qwen-VL-Chat demonstrate substantial advantages in the VC and GC conception tasks, leveraging their chart-tuned data. The most challenging task, VE, serves as a key distinction between ChartBench and ChartQA. Unlike basic OCR, the VE task requires a series of visual and textual logical reasoning steps to arrive at the correct answer. Despite strong overall performance, models such as BLIP2 and ChartLlama face difficulties in the VE task. This underscores the importance of prioritizing and enhancing the visual logical reasoning capabilities of these MLLMs. In terms of model comparison, closed-source models outperform their open-source counterparts, partly attributed to their larger model size and broader data coverage.

\subsection{Leaderboards on \textit{CoR} Metric}
\begin{table*}[ht]
\centering
\begin{minipage}[t]{0.22\textwidth}
    \vspace{0pt}
    \centering
    \setlength\tabcolsep{4.0pt}
    \resizebox{1\linewidth}{!}{
        \begin{tabular}{c|l|c}
        \textbf{No.} & \multicolumn{1}{c|}{\textbf{Model}} & \textbf{\textit{CoR}} \\
        \Xhline{1.0pt}
        \rowcolor{red!26}\textbf{1} & \textbf{\red{GPT-4O}} & \textbf{1.43} \\
        \rowcolor{red!21}\textbf{2} & \textbf{GPT-4V} & \textbf{2.86} \\
        \rowcolor{red!16}\textbf{3} & \textbf{Mini-Gemini} & \textbf{17.86} \\
        \rowcolor{red!11}\textbf{4} & \textbf{ERNIE} & \textbf{19.52} \\
        \rowcolor{red!6}\textbf{5} & \textbf{InternLM-v2} & \textbf{30.24} \\
        6 & mPLUG-Owl & 36.24\\
        7 & OneChart & 36.67\\
        8 & CogAgent & 37.05\\
        9 & ChartLlama & 37.10\\
        10 & Qwen-VL & 42.71\\
        11 & MiniGPT-v2 & 44.19\\
        12 & BLIP2 & 49.24\\
        13 & LLaVA-v1.5 & 51.24\\
        14 & Internlm & 51.38\\
        15 & InstructBLIP & 56.95\\
        16 & DocOwl1.5 & 65.05\\
        17 & CogVLM & 69.33\\
        18 & VisualGLM & 79.19\\
        19 & ChartVLM & 93.57\\
        20 & Shikra & 94.33\\
        21 & SPHINX & 100.0\\
        \end{tabular}
    }
    % \vspace{-0.3cm}
    \subcaption{\textit{Chart Recognition.}}
\end{minipage}
\hfill
\begin{minipage}[t]{0.22\textwidth}
    \vspace{0pt}
    \centering
    \setlength\tabcolsep{4.0pt}
    \resizebox{1\linewidth}{!}{
        \begin{tabular}{c|l|c}
        \textbf{No.} & \multicolumn{1}{c|}{\textbf{Model}} & \textbf{\textit{CoR}} \\
        \Xhline{1.0pt}
        \rowcolor{blue!26}\textbf{1} & \textbf{\red{ERNIE}} & \textbf{44.76} \\
        \rowcolor{blue!21}\textbf{2} & \textbf{GPT-4O} & \textbf{44.76} \\
        \rowcolor{blue!16}\textbf{3} & \textbf{BLIP2} & \textbf{55.14} \\
        \rowcolor{blue!11}\textbf{4} & \textbf{InternLM-v2} & \textbf{57.71} \\
        \rowcolor{blue!6}\textbf{5} & \textbf{DocOwl1.5} & \textbf{58.24} \\
        6 & GPT-4V & 63.33\\
        7 & mPLUG-Owl & 66.24\\
        8 & Mini-Gemini & 70.43\\
        9 & LLaVA-v1.5 & 76.76\\
        10 & Internlm & 80.67\\
        11 & ChartVLM & 80.71\\
        12 & Shikra & 82.14\\
        13 & MiniGPT-v2 & 84.14\\
        14 & Qwen-VL & 84.57\\
        15 & InstructBLIP & 85.14\\
        16 & SPHINX & 85.48\\
        17 & CogAgent & 89.29\\
        18 & CogVLM & 94.29\\
        19 & ChartLlama & 94.90\\
        20 & VisualGLM & 99.67\\
        21 & OneChart & 100.0\\
        \end{tabular}
    }
    % \vspace{-0.3cm}
    \subcaption{\textit{Value Extraction.}}
\end{minipage}
\hfill
\begin{minipage}[t]{0.22\textwidth}
    \vspace{0pt}
    \centering
    \setlength\tabcolsep{4.0pt}
    \resizebox{1\linewidth}{!}{
        \begin{tabular}{c|l|c}
        \textbf{No.} & \multicolumn{1}{c|}{\textbf{Model}} & \textbf{\textit{CoR}} \\
        \Xhline{1.0pt}
        \rowcolor{green!26}\textbf{1} & \textbf{\red{GPT-4O}} & \textbf{16.19} \\
        \rowcolor{green!21}\textbf{2} & \textbf{InternLM-v2} & \textbf{27.71} \\
        \rowcolor{green!16}\textbf{3} & \textbf{GPT-4V} & \textbf{34.76} \\
        \rowcolor{green!11}\textbf{4} & \textbf{ERNIE} & \textbf{41.43} \\
        \rowcolor{green!6}\textbf{5} & \textbf{BLIP2} & \textbf{53.33} \\
        6 & DocOwl1.5 & 55.19\\
        7 & mPLUG-Owl & 56.48\\
        8 & Mini-Gemini & 59.38\\
        9 & Qwen-VL & 63.14\\
        10 & LLaVA-v1.5 & 69.29\\
        11 & MiniGPT-v2 & 69.48\\
        12 & Shikra & 73.71\\
        13 & Internlm & 77.38\\
        14 & CogAgent & 78.86\\
        15 & CogVLM & 80.71\\
        16 & SPHINX & 83.81\\
        17 & ChartVLM & 87.71\\
        18 & ChartLlama & 88.24\\
        19 & InstructBLIP & 96.57\\
        20 & VisualGLM & 99.81\\
        21 & OneChart & 99.81\\
        \end{tabular}
    }
    % \vspace{-0.3cm}
    \subcaption{\textit{Value Comparison.}}
\end{minipage}
\hfill
\begin{minipage}[t]{0.22\textwidth}
    \vspace{0pt}
    \centering
    \setlength\tabcolsep{4.0pt}
    \resizebox{1\linewidth}{!}{
        \begin{tabular}{c|l|c}
        \textbf{No.} & \multicolumn{1}{c|}{\textbf{Model}} & \textbf{\textit{CoR}} \\
        \Xhline{1.0pt}
        \rowcolor{yellow!26}\textbf{1} & \textbf{\red{GPT-4O}} & \textbf{41.43} \\
        \rowcolor{yellow!21}\textbf{2} & \textbf{ERNIE} & \textbf{47.62} \\
        \rowcolor{yellow!16}\textbf{3} & \textbf{GPT-4V} & \textbf{47.62} \\
        \rowcolor{yellow!11}\textbf{4} & \textbf{InternLM-v2} & \textbf{51.46} \\
        \rowcolor{yellow!6}\textbf{5} & \textbf{BLIP2} & \textbf{61.76} \\
        6 & DocOwl1.5 & 63.19\\
        7 & mPLUG-Owl & 66.57\\
        8 & LLaVA-v1.5 & 71.00\\
        9 & Mini-Gemini & 71.10\\
        10 & Qwen-VL & 74.86\\
        11 & InstructBLIP & 78.48\\
        12 & Internlm & 80.90\\
        13 & ChartVLM & 82.71\\
        14 & MiniGPT-v2 & 83.81\\
        15 & Shikra & 85.67\\
        16 & SPHINX & 86.19\\
        17 & CogAgent & 90.00\\
        18 & CogVLM & 90.14\\
        19 & ChartLlama & 94.76\\
        20 & VisualGLM & 99.71\\
        21 & OneChart & 100.0\\
        \end{tabular}
    }
    % \vspace{-0.3cm}
    \subcaption{\textit{Global Conception.}}
\end{minipage}
\vspace{0.1cm}
\caption{Leaderboards of different chart tasks on ChartBench. We report zero-shot \textbf{\textit{CoR}} (\%) performance of the best-performing prompt for each MLLM.}
\label{leaderboard_table4}
\end{table*}

Tab.~\ref{leaderboard_table4} showcases the \textit{CoR} metric, which signifies the portion of the chart that the MLLM fails to comprehend entirely. Qwen-VL-Chat exhibits the highest \textit{Acc+}, albeit with a lower \textit{CoR} compared to models like MiniGPT-v2. The top-performing MiniGPT-v2 demonstrates a \textit{CoR} of 55.06\%, underscoring the prevalence of random guessing cases for open-source models due to their challenges in accurately understanding charts. In the case of closed-source MLLMs, although GPT-4V outperforms ERNIE in terms of \textit{Acc+}, their \textit{CoR} values are similar. A more detailed examination reveals that ERNIE excels in challenging VE tasks, which happen to be the weaker area for GPT-4V.

\subsection{Leaderboards on with/without Annotated Charts}
\begin{table*}[ht]
\centering
\begin{minipage}[t]{0.22\textwidth}
    \vspace{0pt}
    \centering
    \setlength\tabcolsep{4.0pt}
    \resizebox{1\linewidth}{!}{
        \begin{tabular}{c|l|c}
        \textbf{No.} & \multicolumn{1}{c|}{\textbf{Model}} & \textbf{\textit{Acc+}} \\
        \Xhline{1.0pt}
        \rowcolor{red!26}\textbf{1} & \textbf{\red{GPT-4O}} & \textbf{83.30} \\
        \rowcolor{red!21}\textbf{2} & \textbf{GPT-4V} & \textbf{77.40} \\
        \rowcolor{red!16}\textbf{3} & \textbf{InternLM-v2} & \textbf{73.16} \\
        \rowcolor{red!11}\textbf{4} & \textbf{DocOwl-v1.5} & \textbf{50.19} \\
        \rowcolor{red!6}\textbf{5} & \textbf{ERNIE} & \textbf{49.44} \\
        6 & Qwen-VL & 45.71\\
        7 & Mini-Gemini & 44.46\\
        8 & ChartLlama & 33.59\\
        9 & LLaVA-v1.5 & 29.76\\
        10 & CogAgent & 29.52\\
        11 & mPLUG-Owl & 24.83\\
        12 & BLIP2 & 24.11\\
        13 & SPHINX & 22.40\\
        14 & CogVLM & 21.78\\
        15 & MiniGPT-v2 & 21.46\\
        16 & OneChart & 18.39\\
        17 & ChartVLM & 18.20\\
        18 & InstructBLIP & 14.03\\
        19 & InternLM & 12.02\\
        20 & VisualGLM & 6.79\\
        21 & Shikra & 6.06\\
        \end{tabular}
    }
    % \vspace{-0.3cm}
    \subcaption{\textit{With Annotations.}}
\end{minipage}
\hfill
\begin{minipage}[t]{0.22\textwidth}
    \vspace{0pt}
    \centering
    \setlength\tabcolsep{4.0pt}
    \resizebox{1\linewidth}{!}{
        \begin{tabular}{c|l|c}
        \textbf{No.} & \multicolumn{1}{c|}{\textbf{Model}} & \textbf{\textit{Acc+}} \\
        \Xhline{1.0pt}
        \rowcolor{blue!26}\textbf{1} & \textbf{\red{GPT-4O}} & \textbf{61.00} \\
        \rowcolor{blue!21}\textbf{2} & \textbf{InternLM-v2} & \textbf{54.80} \\
        \rowcolor{blue!16}\textbf{3} & \textbf{DocOwl-v1.5} & \textbf{43.50} \\
        \rowcolor{blue!11}\textbf{4} & \textbf{GPT-4V} & \textbf{43.00} \\
        \rowcolor{blue!6}\textbf{5} & \textbf{ERNIE} & \textbf{42.95} \\
        6 & Mini-Gemini & 32.25\\
        7 & Qwen-VL & 28.70\\
        8 & mPLUG-Owl & 26.45\\
        9 & LLaVA-v1.5 & 22.55\\
        10 & ChartLlama & 22.10\\
        11 & BLIP2 & 20.95\\
        12 & MiniGPT-v2 & 20.45\\
        13 & CogAgent & 17.95\\
        14 & SPHINX & 16.85\\
        15 & ChartVLM & 15.55\\
        16 & InternLM & 14.70\\
        17 & CogVLM & 12.60\\
        18 & InstructBLIP & 11.15\\
        19 & OneChart & 9.10\\
        20 & Shikra & 5.55\\
        21 & VisualGLM & 3.40\\
        \end{tabular}
    }
    % \vspace{-0.3cm}
    \subcaption{\textit{Without Annotations.}}
\end{minipage}
\hfill
\begin{minipage}[t]{0.22\textwidth}
    \vspace{0pt}
    \centering
    \setlength\tabcolsep{4.0pt}
    \resizebox{1\linewidth}{!}{
        \begin{tabular}{c|l|c}
        \textbf{No.} & \multicolumn{1}{c|}{\textbf{Model}} & \textbf{\textit{CoR}} \\
        \Xhline{1.0pt}
        \rowcolor{green!26}\textbf{1} & \textbf{\red{GPT-4O}} & \textbf{10.62} \\
        \rowcolor{green!21}\textbf{2} & \textbf{GPT-4V} & \textbf{18.75} \\
        \rowcolor{green!16}\textbf{3} & \textbf{InternLM-v2} & \textbf{20.88} \\
        \rowcolor{green!11}\textbf{4} & \textbf{ERNIE} & \textbf{35.00} \\
        \rowcolor{green!6}\textbf{5} & \textbf{DocOwl-v1.5} & \textbf{44.50} \\
        6 & Qwen-VL & 51.00\\
        7 & Mini-Gemini & 51.94\\
        8 & MiniGPT-v2 & 53.37\\
        9 & LLaVA-v1.5 & 54.81\\
        10 & ChartLlama & 63.31\\
        11 & mPLUG-Owl & 65.44\\
        12 & BLIP2 & 66.00\\
        13 & SPHINX & 67.31\\
        14 & CogAgent & 71.06\\
        15 & OneChart & 73.94\\
        16 & CogVLM & 78.00\\
        17 & InstructBLIP & 81.06\\
        18 & InternLM & 82.62\\
        19 & ChartVLM & 88.50\\
        20 & VisualGLM & 93.31\\
        21 & Shikra & 95.25\\
        \end{tabular}
    }
    % \vspace{-0.3cm}
    \subcaption{\textit{With Annotations.}}
\end{minipage}
\hfill
\begin{minipage}[t]{0.22\textwidth}
    \vspace{0pt}
    \centering
    \setlength\tabcolsep{4.0pt}
    \resizebox{1\linewidth}{!}{
        \begin{tabular}{c|l|c}
        \textbf{No.} & \multicolumn{1}{c|}{\textbf{Model}} & \textbf{\textit{CoR}} \\
        \Xhline{1.0pt}
        \rowcolor{yellow!26}\textbf{1} & \textbf{\red{GPT-4O}} & \textbf{23.75} \\
        \rowcolor{yellow!21}\textbf{2} & \textbf{InternLM-v2} & \textbf{33.69} \\
        \rowcolor{yellow!16}\textbf{3} & \textbf{ERNIE} & \textbf{35.62} \\
        \rowcolor{yellow!11}\textbf{4} & \textbf{GPT-4V} & \textbf{41.25} \\
        \rowcolor{yellow!6}\textbf{5} & \textbf{DocOwl-v1.5} & \textbf{50.12} \\
        6 & Mini-Gemini & 52.56\\
        7 & MiniGPT-v2 & 54.31\\
        8 & LLaVA-v1.5 & 58.06\\
        9 & Qwen-VL & 62.31\\
        10 & mPLUG-Owl & 63.19\\
        11 & BLIP2 & 69.56\\
        12 & ChartLlama & 71.00\\
        13 & SPHINX & 71.12\\
        14 & InternLM & 76.38\\
        15 & CogAgent & 80.06\\
        16 & CogVLM & 82.25\\
        17 & InstructBLIP & 82.81\\
        18 & OneChart & 86.44\\
        19 & ChartVLM & 87.31\\
        20 & Shikra & 91.75\\
        21 & VisualGLM & 95.44\\
        \end{tabular}
    }
    % \vspace{-0.3cm}
    \subcaption{\textit{Without Annotations.}}
\end{minipage}
\vspace{0.1cm}
\caption{Leaderboards w.r.t. data annotations of \textbf{\textit{Acc+}} (\%) and \textbf{\textit{CoR}} (\%) performance on ChartBench.}
\label{leaderboard_anno}
\end{table*}

The rationale behind ChartBench is to assess the comprehension of unlabeled charts by MLLMs. In Tab.~\ref{leaderboard_anno}, the performance of all MLLMs on both annotated and unannotated charts is presented. It is important to note that: 1) Virtually all models exhibit significantly superior performance on annotated charts when compared to unannotated ones. This discrepancy arises because MLLMs heavily depend on OCR to acquire answer candidates, thereby enhancing answer accuracy—an advantage not applicable to unannotated charts. 2) The larger the performance gap between models, such as Qwen-VL-Chat (+16.00\%) and GPT-4V (+31.39\%), the more favorable their overall performance. This suggests that the \textit{Acc+} of MLLMs is primarily elevated by annotated charts, while unannotated charts notably intensify the challenge presented by ChartBench.

\clearpage
\section{Chart Type Thumbnails}
\label{apdx_sec_thumbnails}

\begin{figure*}[ht]
  \centering
  \begin{overpic}[width=1\linewidth, grid=False]{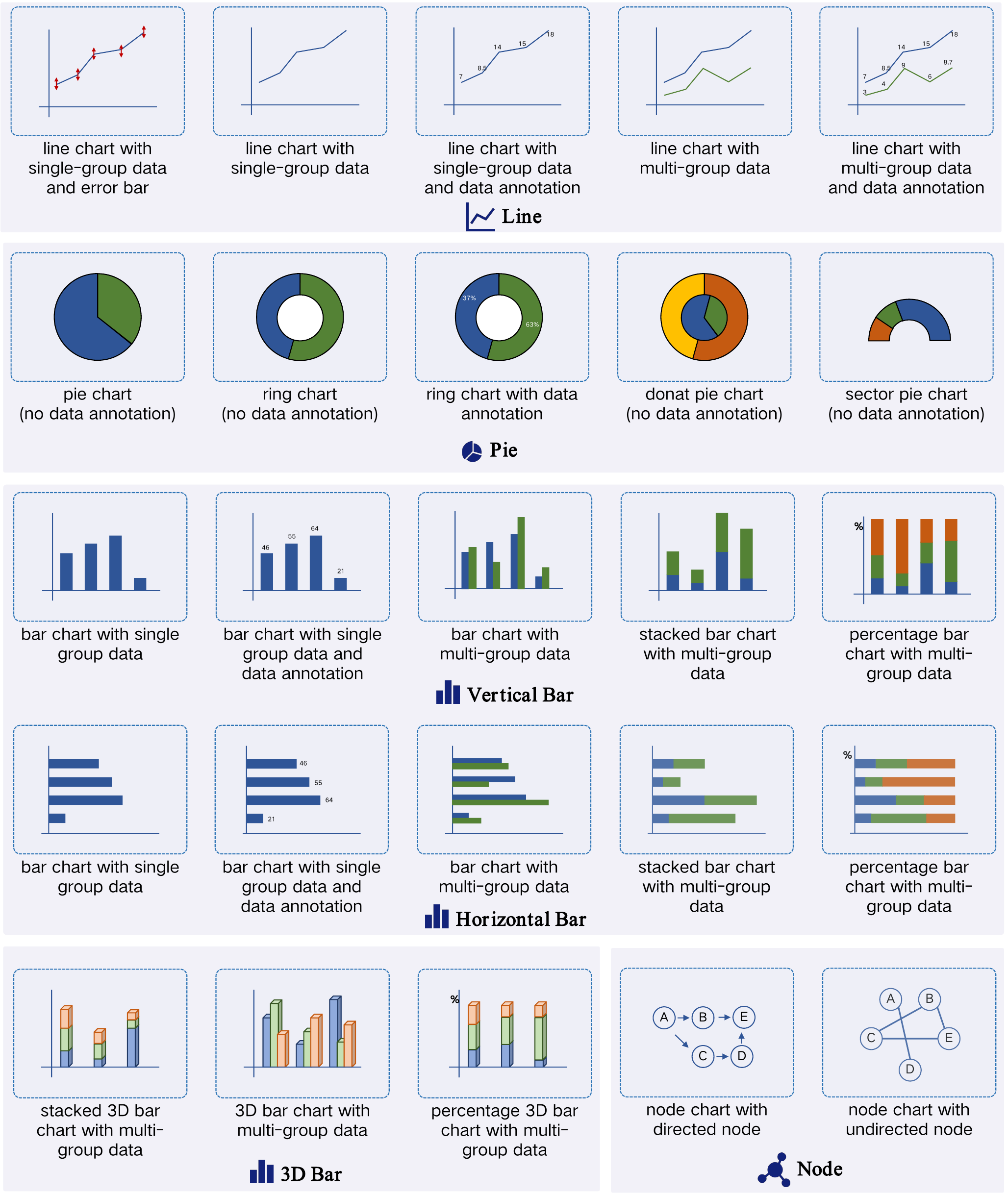}
  \end{overpic}
  \caption{The categories and thumbnail examples of ChartBench (Part 1). We strive to avoid direct labeling of chart data to encourage MLLMs to understand charts using human-like visual reasoning and ensure the credibility of the data. The example charts are provided as thumbnail representations of the corresponding chart features.}
  % \vspace{-10pt}
  \label{fig_bench_visual_1}
\end{figure*}

\begin{figure*}[ht]
  \centering
  \begin{overpic}[width=1\linewidth, grid=False]{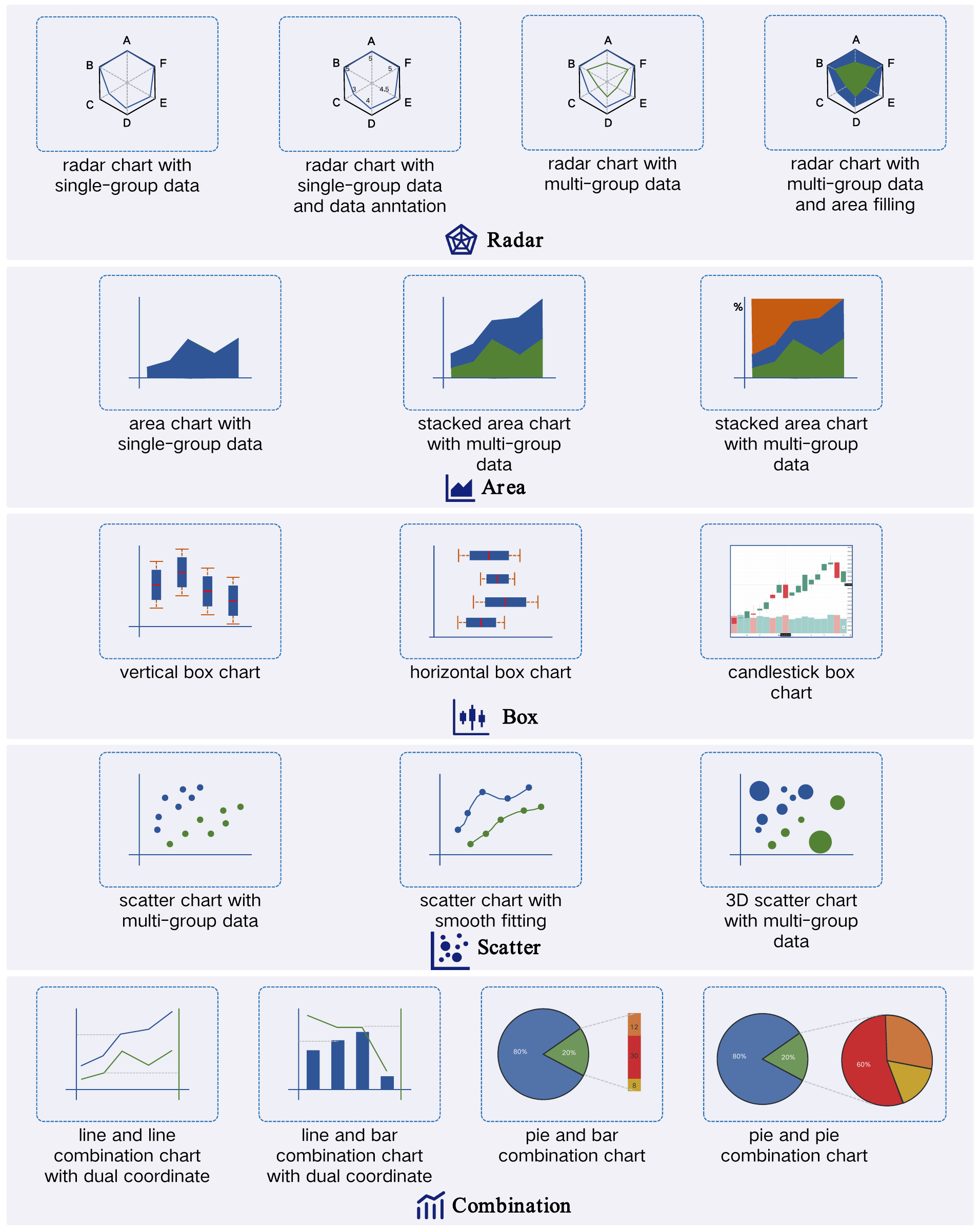}
  \end{overpic}
  \caption{The categories and thumbnail examples of ChartBench (Part 2). We strive to avoid direct labeling of chart data to encourage MLLMs to understand charts using human-like visual reasoning and ensure the credibility of the data. The example charts are provided as thumbnail representations of the corresponding chart features.}
  % \vspace{-10pt}
  \label{fig_bench_visual_2}
\end{figure*}

Previous benchmarks~\cite{ChartQA, PlotQA, OpenCQA, Chart2Text, OneChart} mainly focus on the line, bar, and pie charts. To enlarge chart diversity, ChartBench provides 9 major categories and 42 subcategories of charts, including regular and specialized ones. We provide thumbnails of all chart types for visualizations in Fig.~\ref{fig_bench_visual_1} \& ~\ref{fig_bench_visual_2}.

\end{document}